\documentclass{article}

\usepackage{microtype}
\usepackage{graphicx}
\usepackage{subfigure}
\usepackage{booktabs} 
\usepackage{float}

\usepackage{hyperref}

\usepackage[accepted]{icml2025}

\usepackage{amsmath}
\usepackage{amssymb}
\usepackage{mathtools}
\usepackage{amsthm}

\usepackage[capitalize,noabbrev]{cleveref}

\theoremstyle{plain}

\theoremstyle{definition}

\theoremstyle{remark}

\usepackage[textsize=tiny]{todonotes}

\icmltitlerunning{TabFSBench: Tabular Benchmark for Feature Shifts in Open Environment}

\usepackage{multirow} 
\usepackage{lscape} 
\usepackage{pdfpages} 
\usepackage{longtable} 
\usepackage[T1]{fontenc} 
\usepackage{graphicx} 
\usepackage[most]{tcolorbox} 
\usepackage{flushend}

\definecolor{codegreen}{rgb}{0,0.6,0}
\definecolor{codegray}{rgb}{0.5,0.5,0.5}
\definecolor{codepurple}{rgb}{0.58,0,0.82}
\definecolor{backcolour}{rgb}{0.95,0.95,0.92}

\lstdefinestyle{mystyle}{
    backgroundcolor=\color{backcolour},   
    commentstyle=\color{codegreen},
    keywordstyle=\color{magenta},
    numberstyle=\tiny\color{codegray},
    stringstyle=\color{codepurple},
    basicstyle=\ttfamily\footnotesize,
    breakatwhitespace=false,         
    breaklines=true,                 
    captionpos=b,                    
    keepspaces=true,                 
    numbers=left,                    
    numbersep=5pt,                  
    showspaces=false,                
    showstringspaces=false,
    showtabs=false,                  
    tabsize=4
}
\begin{document}

\twocolumn[
\icmltitle{TabFSBench: Tabular Benchmark for Feature Shifts in Open Environments}




\begin{icmlauthorlist}
\icmlauthor{Zi-Jian Cheng}{sch,comp}
\icmlauthor{Zi-Yi Jia}{sch,comp}
\icmlauthor{Zhi Zhou}{comp,yyy}
\icmlauthor{Yu-Feng Li}{comp,yyy}
\icmlauthor{Lan-Zhe Guo}{sch,comp}
\end{icmlauthorlist}

\icmlaffiliation{comp}{National Key Laboratory for Novel Software Technology, Nanjing University, China}
\icmlaffiliation{yyy}{School of Artificial Intelligence, Nanjing University, China}
\icmlaffiliation{sch}{School of Intelligence Science and Technology, Nanjing University, China}

\icmlcorrespondingauthor{Yu-Feng Li}{liyf@nju.edu.cn}
\icmlcorrespondingauthor{Lan-Zhe Guo}{guolz@nju.edu.cn}

\icmlkeywords{Machine Learning, ICML}

\vskip 0.3in
]



\printAffiliationsAndNotice{}  

\begin{abstract}
Tabular data is widely utilized in various machine learning tasks. Current tabular learning research predominantly focuses on closed environments, while in real-world applications, open environments are often encountered, where distribution and feature shifts occur, leading to significant degradation in model performance. Previous research has primarily concentrated on mitigating distribution shifts, whereas feature shifts, a distinctive and unexplored challenge of tabular data, have garnered limited attention. To this end, this paper conducts the first comprehensive study on feature shifts in tabular data and introduces the first \textbf{tab}ular \textbf{f}eature-\textbf{s}hift \textbf{bench}mark (TabFSBench). TabFSBench evaluates impacts of four distinct feature-shift scenarios on four tabular model categories across various datasets and assesses the performance of large language models (LLMs) and tabular LLMs in the tabular benchmark for the first time. Our study demonstrates three main observations: (1) most tabular models have the limited applicability in feature-shift scenarios; (2) the shifted feature set importance has a linear relationship with model performance degradation; (3) model performance in closed environments correlates with feature-shift performance. Future research direction is also explored for each observation.

Benchmark: \href{https://github.com/LAMDASZ-ML/TabFSBench}{LAMDASZ-ML/TabFSBench}.
\end{abstract}

\section{Introduction}

\begin{figure*}
\begin{center}
\centerline{\includegraphics[width=\textwidth]{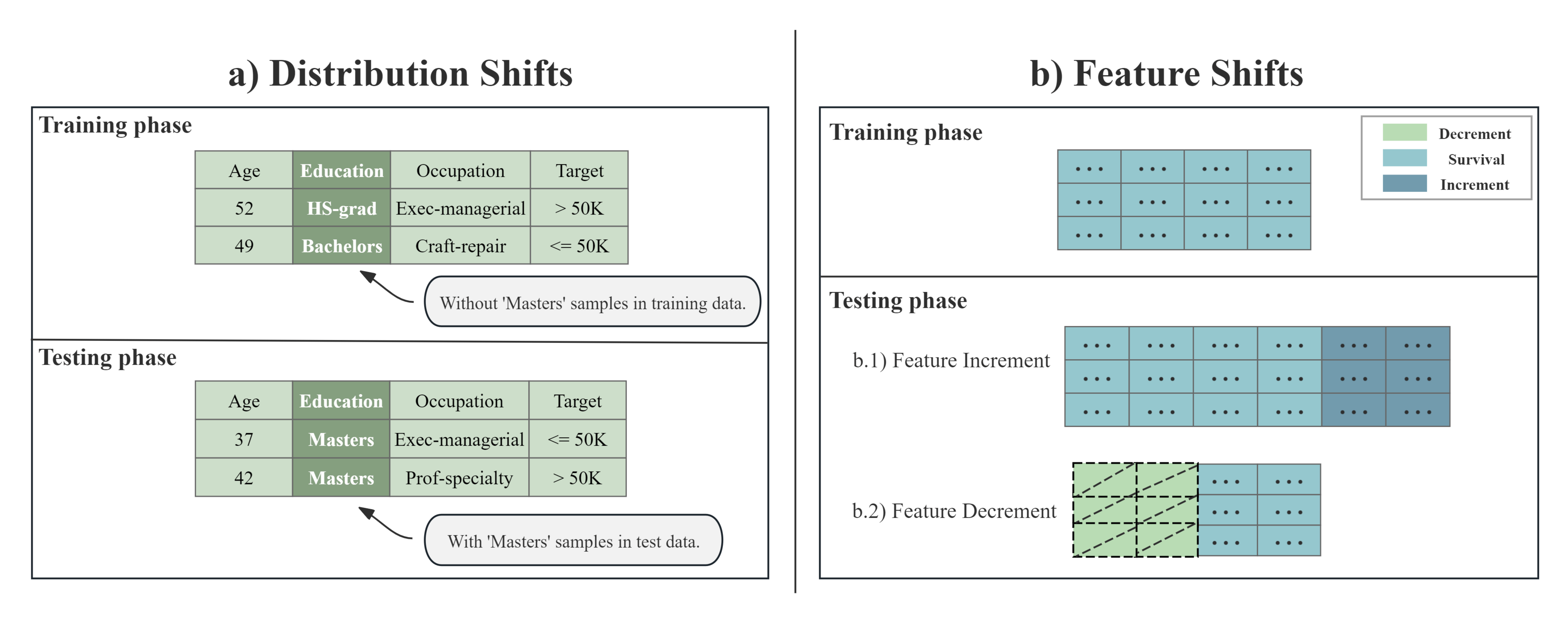}}
\vskip -0.1in
\caption{Open environment challenges: a) Distribution Shifts. Change in data distribution between training and testing while keeping features unchanged. b) Feature Shifts. Change in the feature set between training and testing, either by addition or removal of features.}
\label{figure1}
\end{center}
\vskip -0.2in
\end{figure*}

Tabular data~\cite{altman2017tabular} represents a meticulously structured format of data, systematically arranged in rows and columns, where each row signifies a record and each column corresponds to a feature~\cite{sahakyan2021explainable}. The applications of tabular data in real-world scenarios are extensive. For instance, tabular data is utilized in financial analyses, including credit scoring~\cite{west2000neural} and stock market predictions~\cite{zhu2021tat}. Furthermore, tabular data forms the foundation for various applications in the medical sector, encompassing disease diagnosis~\cite{yildiz2024gradient} and pharmaceutical development~\cite{meijerink2020uncertainty}. Existing machine learning models applicable to tabular data can be divided into four categories, containing tree-based models~\cite{prokhorenkova2019catboostunbiasedboostingcategorical, Ke2017LightGBMAH}, deep-learning models~\cite{Hollmann2022TabPFNAT, ye2024modernneighborhoodcomponentsanalysis}, LLMs~\cite{touvron2023llama, openai2024gpt4technicalreport}, and specialized LLMs designed for tabular data~\cite{hegselmann2023tabllmfewshotclassificationtabular, wang2023unipredict}, namely tabular LLMs. These models have exhibited remarkable effectiveness across diverse tabular tasks.

Most tabular models of the aforementioned four categories are typically trained and tested in closed environments~\cite{zhou2022open}, where both the distribution of data and the space of features are maintained consistently. However, real-world tabular tasks frequently occur in open environments~\cite{parmar2023open}, where significant challenges such as distribution and feature shift occur. For example, in a traffic management system, alterations in traffic flow due to unforeseen incidents (e.g., extreme meteorological events) illustrate modifications in data distribution. Simultaneously, the failure or substitution of specific traffic monitoring equipment can induce feature shifts. These dynamic alterations underscore the complexity and unpredictability associated with machine learning tasks in open environments, necessitating the development of models characterized by enhanced adaptability and robustness. Therefore, recent research has progressively concentrated on the advancement of tabular models designed to acclimate effectively in open environments.

Distribution shift~\cite{zhou24ftta} represents a challenge in open environments that has attracted substantial research attention. It refers to the phenomenon where tabular data processed by a model during the training and testing phases exhibit significant differences in distribution~\cite{wang2021tent}. Numerous research have proposed solutions to address distribution shifts, including domain adaptation~\cite{ajakan2014domain, arjovsky2019invariant} and domain generalization~\cite{zhou2022domain, zhao2024domain}. Various benchmarks on distribution shifts in tabular data have been established. For example, TableShift~\cite{gardner2024benchmarking} focuses on distribution shifts in tabular classification tasks, while Wild-Tab~\cite{kolesnikov2023wild} concentrates on distribution shifts in tabular regression tasks.

Feature shift, another prevalent and substantial challenge in open environments, denotes the phenomenon in which the set of features available to a model for the same task dynamically changes due to temporal evolution or spatial variations. For instance, in the weather forecasting task, critical sensors may cease functioning due to malfunctions or aging, or they may be replaced by new sensors due to technological upgrades, leading to significant alterations in accessible features. This shift not only disrupts the consistency of the model's input features but may also significantly degrade the model performance and robustness. However, research on feature shifts is relatively limited, and there is a lack of high-quality benchmarks for feature-shift scenarios. Therefore, comprehensively investigating the challenge of feature shifts in tabular data is important for enhancing the performance and robustness of models on feature shifts in open environments scenarios.

To this end, this paper presents the first systematic study on the challenge of feature shifts in open environments scenarios and proposes the first feature-shift benchmark for tabular data (TabFSBench). Our contributions are as follows:
\begin{itemize}
    \item \textbf{Tabular Benchmark for Feature Shifts.} We provide publicly available datasets from classification and regression tasks across various domains. We design four feature-shift scenarios: single shift, most/least-relevant shift, and random shift, to evaluate models.
    \item \textbf{Implementations of Various Models and APIs.} We evaluate four categories of tabular models and conduct the first assessment on LLMs and tabular LLMs in the tabular benchmark. We provide callable APIs to facilitate future research on the feature-shift challenge.
    \item \textbf{Empirical Research and Analysis.} We conduct extensive experiments on feature shifts and identify three observations alongside associated future work:
    \begin{itemize}
        \item Most models have the limited applicability in feature shifts, where tabular LLMs demonstrate potential. Future work can design multi-level finetuning frameworks to enhance tabular LLMs' reasoning capabilities.
        \item Shifted features' importance has a linear trend with model performance degradation. Future work can propose feature importance-driven optimization algorithms to emphasize and protect strong-correlated features.
        \item Model closed environment performance correlates with feature-shift performance. Solid theoretical analysis of the relationship between close and open environments performance should be studied and how to further improve the robustness of these models in open environments is also an important research direction.
    \end{itemize}
\end{itemize}

\section{Task and Notation}
\label{2}
\subsection{Task Setting}

\begin{table*}
\caption{Datasets in TabFSBench. \textbf{\#Numerical} means numerical features. \textbf{\#Categorical} means categorical features.}
\label{table1}
\vskip 0.2in
\begin{center}
\begin{small}
\begin{tabular}{cccccc}
\toprule
Tasks                                       & Dataset        & \#Samples & \#Numerical & \#Categorical & \#Labels \\
\midrule
\multirow{4}{*}{Binary Classification}      & credit         & 1,000      & 7          & 13            & 2         \\
                                            & electricity    & 45,312     & 8          & 0             & 2         \\
                                            & heart          & 918       & 6          & 5             & 2         \\
                                            & MiniBooNE      & 72,998     & 50         & 0             & 2         \\ 
\midrule
\multirow{4}{*}{Multi-Class Classification} & Iris           & 150       & 4          & 0             & 3         \\
                                            & penguins       & 345       & 4          & 2             & 3         \\
                                            & eye\_movements & 10,936     & 27         & 0             & 4         \\
                                            & jannis         & 83,733     & 54         & 0             & 4         \\ 
\midrule
\multirow{4}{*}{Regression}                 & abalone        & 4,178      & 7          & 1             & \textbackslash{}         \\
                                            & bike           & 10,886     & 6          & 3             & \textbackslash{}         \\
                                            & concrete       & 1,031      & 8          & 0             & \textbackslash{}         \\
                                            & laptop         & 1,275      & 8          & 14            & \textbackslash{} \\
\bottomrule
\end{tabular}

\end{small}
\end{center}
\vskip -0.1in
\end{table*}

Formally, the goal of tabular prediction tasks is to train a machine-learning model \( f: \mathcal{X} \rightarrow \mathcal{Y} \), where \( \mathcal{X} \) is the input space and \( \mathcal{Y} \) is the output space. We define the set of features in \( \mathcal{X} \) as $C$.

\subsection{Feature Shifts Notation}
We partition the feature set $C$ from training and testing phases into $C^{train}$ and $C^{test}$. In closed environments, the feature set from the training phase is identical to that from the testing phase, i.e., $C^{train} = C^{test}$. While in open environments scenarios, although $C^{train}$ remains invariant, $C^{test}$ may encounter two challenges, namely, distribution shift and feature shift. We illustrate these challenges by using forest disease monitoring as an example, where a sensor can be regarded as a feature. Figure~\ref{figure1} depicts these challenges of open environments.

\paragraph{Distribution Shift.} When certain raw sensors fail, new sensors are installed and commence monitoring. The number of features remains unchanged, i.e., $\mathcal{C}_{\text{train}} = \mathcal{C}_{\text{test}}$, but the monitoring precision (data distribution) shifts. Therefore, distribution shift refers to the phenomenon where data distribution varies between training and testing phases (see Figure~\ref{figure1}(a)). Predictions can be generated without any data processing, although model performance may deteriorate due to covariate shifts or concept shifts~\cite{shao2024open}. 

\paragraph{Feature Shift.} Feature shift is another open environments challenge, wherein the feature set previously utilized as inputs is either removed partially or added by new features. It contains two scenarios:
\begin{itemize}
        \item \textbf{Feature Increment.} Additional sensors are deployed and no existing sensor fails, resulting in an expansion of the feature set (see Figure~\ref{figure1}(b.1)), i.e., $C^{train} \subsetneqq C^{test}$. In this case, to maintain consistency in input dimensions between training and testing phases, the model typically truncates the newly added features in $C^{test}$ and retains features from $C^{test}$ corresponding to the features in $C^{train}$.
        \item \textbf{Feature Decrement.} Certain existing sensors cease to function and no new sensors are added, leading to a reduction in the feature set (see Figure~\ref{figure1}(b.2)), i.e., $C^{test} \subsetneqq C^{train}$. In this case, to maintain consistency in input dimensions between training and testing phases and allow the model to predict properly, shifted features in $C^{test}$ need to be imputed.
\end{itemize}

Given the limited research addressing feature shift and the established observation that feature increment generally does not result in model performance degradation, we conduct an empirical experimental analysis focusing on feature-decrement scenarios.

\section{TabFSBench: A Feature-Shift Benchmark for Tabular Data}
\label{3}
TabFSBench is a benchmark for evaluating feature shifts in tabular data, comprising twelve tabular tasks and assessing four categories of tabular models. We compare model performance across four feature-shift scenarios and closed environments. Additionally, we provide callable Python APIs\footnote{https://github.com/LAMDASZ-ML/TabFSBench} to facilitate future research on the feature-shift challenge in open environments.

\subsection{Datasets}
\label{3.1}
To effectively reproduce feature-shift scenarios, we select open-source and reliable datasets from OpenML and Kaggle's extensive dataset library, including three curated tasks of binary classification, multi-class classification, and regression, covering various domains such as finance, healthcare and geology. The primary attributes of the datasets used in TabFSBench are presented in Table~\ref{table1}. Detailed information on the datasets can be found in Appendix~\ref{appendix:B}.

\begin{table}[t]
\caption{Kendall's $\tau$ coefficients for four feature importance metrics consistency.}
\label{kendall}
\begin{center}
\begin{tabular}{cc}
\toprule
Metric             & $\tau$    \\
\midrule
Pearson            & 0.60 \\
Spearman           & 0.61 \\
SHAP               & 0.49 \\
Mutual Information & 0.53 \\
\bottomrule
\end{tabular}
\end{center}
\vskip -0.2in
\end{table}

\subsection{Feature-shift Scenarios}
\label{3.2}
To evaluate the consistency of feature importance rankings, we compute Kendall's $\tau$ correlation coefficients among four metrics: Pearson correlation coefficient, Spearman's rank correlation, SHAP values, and mutual information. Table~\ref{kendall} reveals high concordance across these measures, with particularly strong agreement between PCC and Spearman ($\tau$ = 0.61). While both demonstrate comparable performance, we ultimately select PCC for its widespread adoption and intuitive interpretability. The marginal differences between these metrics' rankings are found to be statistically insignificant and not affect our analytical conclusions. 

Hence, to effectively assess the impact of feature shifts on model performance and analyze whether there is a relationship between the importance of shifted features and model performance degradation, we employ pearson correlation $\rho$ to rank features of the given dataset, thereby indicating the importance of each feature for the task. $|\rho|$ greater than 0.7 can be considered as a moderate linear correlation\cite{iversen2012statistics}. Details regarding experiments on four feature importance metrics and pearson correlation are provided in Appendix~\ref{appendix:D}.

\paragraph{Single Shift.}
To evaluate the impact of the absence of features with different importance levels on model performance, we design the single shift experiment. For a given dataset, we first compute correlations of features. Then, we sequentially remove one feature by employing a sampling-with-replacement approach in the ascending order. Finally, we compare the gap in model performance among shifted features with different correlations.

\paragraph{Most/Least-Relevant Shift.}
\label{3.2.2} 
To evaluate how the model performance changes when the feature set with different importance is shifted, we design the most/least-relevant shift experiment. For a given dataset, we first compute correlations of features. Then, we remove features in the ascending (least relevant) or descending (most relevant) order. Finally, we compare the gap in model performance among shifted feature sets with different importance.

\paragraph{Random Shift.} 
To systematically evaluate model robustness under feature shifts, we design a controlled experiment where we randomly sample feature subsets from the training feature space $\mathcal{C}^{train}$ to create shifted test scenarios. By progressively removing features during testing (where removing one feature represents a shift ratio of $1/n$, with $n$ being the total feature count), we quantify performance degradation across different shift magnitudes. For statistical reliability, we evaluate up to $\min(10,000, \binom{N}{n})$ distinct feature combinations per shift ratio $n/N$ and report the mean performance across all combinations.

\subsection{Impute Strategy.} We compare the performance of various models by using their respective imputation methods, random imputation, and mean imputation. The results (see Appendix~\ref{appendix:impute}) demonstrate that mean imputation better simulates scenarios with shifted features. Notably, benchmarks such as LAMDA-Talent~\cite{liu2024talenttabularanalyticslearning} also adopt mean imputation for handling missing values, underscoring its widespread applicability. Hence, we opt for uniform mean-value imputation of shifted features to ensure that predictions intuitively reflect model robustness in feature-shift scenarios. Specifically, for numerical features, we employ the mean value of the feature within the training set as the imputed value. For categorical features, we utilize the value that occurs most frequently as the imputed value. We do not select zero or other arbitrary values for imputation, as this would introduce artificial shifts in the data distribution, which contradicts the objective of evaluating model robustness in the context of feature shifts.

\subsection{TabFSBench API}
To facilitate the use of TabFSBench for experimental setups in feature-shift scenarios, we have designed the TabFSBench APIs. More details are available at \href{https://github.com/LAMDASZ-ML/TabFSBench}{https://github.com/LAMDASZ-ML/TabFSBench}. The APIs are divided into five parameters: dataset, model, task, degree, and export\_dataset.

The \texttt{dataset} parameter specifies the dataset to be used and requires the full name of the dataset. TabFSBench supports datasets from OpenML, Kaggle, and local directories.

The \texttt{model} parameter defines the model to be evaluated and can be selected from tree-based models, deep-learning models, LLMs, and tabular LLMs. New models can be added by following the instructions in the "How to Add New Models" section.

The \texttt{task} parameter determines the type of feature-shift experiment to be conducted. The available options include single, least, most, and random.

The \texttt{degree} parameter indicates the proportion of features to be shifted. The valid range for this parameter is from 0 to 1, where 0 signifies no features are removed, and 1 signifies all features are removed.

The \texttt{export\_dataset} parameter controls whether the modified dataset (after removing shifted features) is exported as a CSV file for further use.

\begin{figure}
\begin{center}
\centerline{\includegraphics[width=\linewidth, trim=1cm 0cm 1cm 0cm, clip]{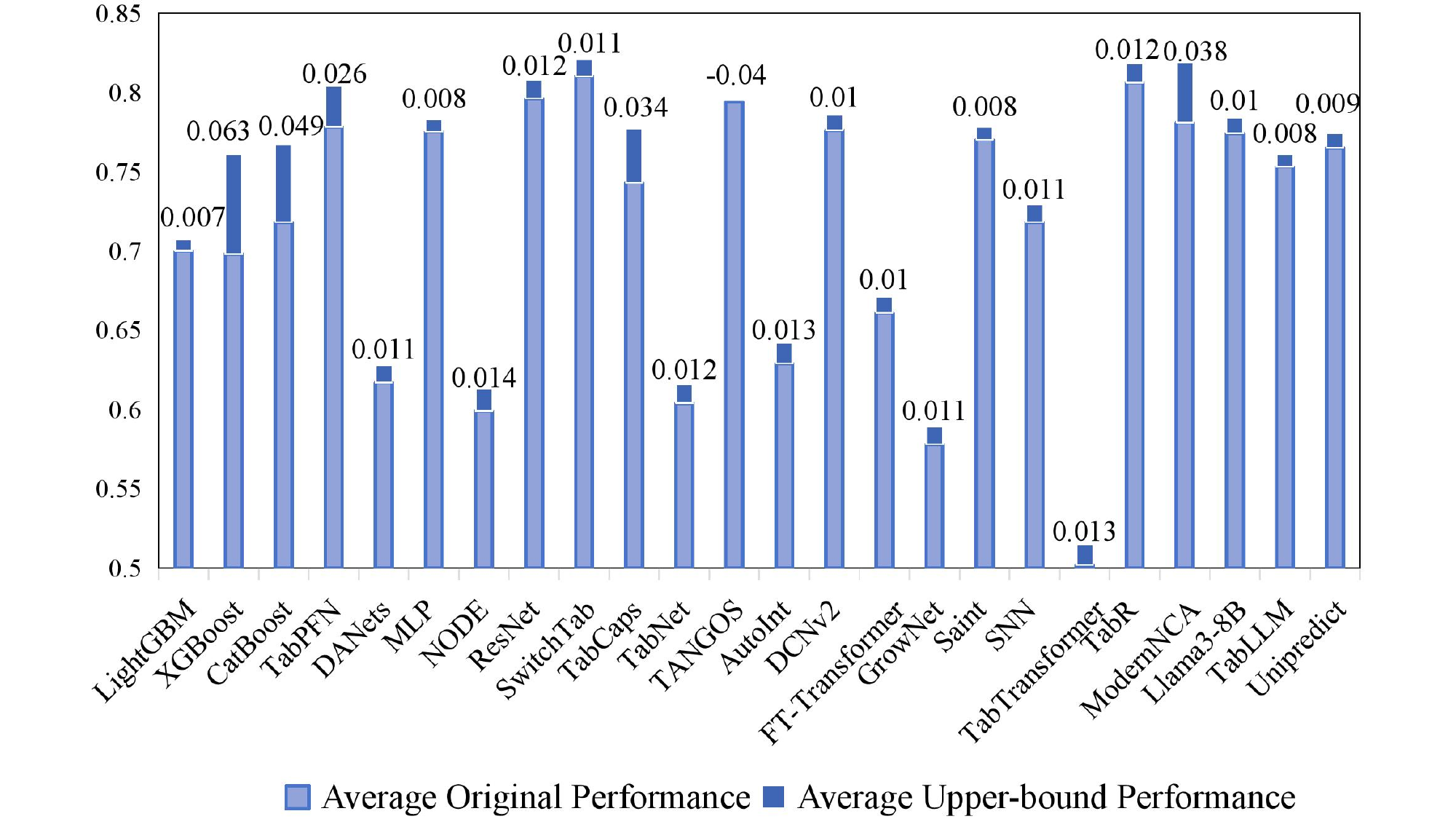}}
\caption{Upper bound and original model performance.}
\label{upper}
\end{center}
\vskip -0.3in
\end{figure}

An example command for running a feature-shift experiment in TabFSBench is as follows:

\vspace{10pt}
\begin{minipage}{\columnwidth}
\begin{tcolorbox}[colback=white, colframe=black, title=Example Command, fonttitle=\bfseries, rounded corners]
\texttt{python run\_experiment.py ---dataset Adult ---model LightGBM ---task random ---degree 0.3 ---export\_dataset True}
\end{tcolorbox}
\end{minipage}

\section{Experiment Setup}

\begin{table*}
\caption{Average performance gap in different tasks. We attribute the different feature-shift degree for each dataset to 20\%, 40\%, 60\%, 80\% and 100\%. Then compute the model's performance gap for each degree of feature shifts, task by task. '$\backslash$' means this model can't handle the regression task. For classification tasks, we choose accuracy. For regression tasks, we choose RMSE. The best is in \textbf{bold} and second best is \underline{underlined}. Model abbreviations are in Appendix~\ref{appendix:C}.}
\label{table2}
\begin{center}
\begin{small}

\resizebox{\textwidth}{16mm}{
\begin{tabular}{cccccccccccccccccccccccccc}
\toprule
Task                                            & Shift & LGB  & XGB  & CATB         & PFN              & DAN  & MLP & NODE & RES & SWI  & CAP              & NET  & GOS       & INT  & DCN & FTT  & GRO  & SAT & SNN  & TTF  & TABR & NCA  & LMA  & TLLM               & UNI            \\
\midrule
\multirow{5}{*}{\textbf{Binary Classification}} & 20\%  & -0.065 & -0.076 & -0.035 & -0.048           & -0.022      & -0.024 & \textbf{-0.009} & -0.031 & -0.010          & -0.020           & -0.015 & -0.026 & -0.028 & -0.026 & -0.028 & \underline{ -0.010}    & -0.016 & -0.018 & -0.016         & -0.062 & -0.161 & -0.045          & -0.022           & -0.019\\
 & 40\%  & -0.192 & -0.218 & -0.105 & -0.118           & -0.056      & -0.062 & \underline{ -0.031}    & -0.085 & \textbf{-0.031} & -0.051           & -0.055 & -0.069 & -0.065 & -0.070 & -0.077 & -0.041          & -0.042 & -0.048 & -0.044         & -0.158 & -0.271 & -0.118          & -0.042           & -0.038 \\
 & 60\%  & -0.249 & -0.261 & -0.155 & -0.165           & -0.090      & -0.107 & -0.061          & -0.150 & -0.063          & -0.090           & -0.099 & -0.121 & -0.115 & -0.120 & -0.138 & -0.077          & -0.072 & -0.086 & -0.078         & -0.214 & -0.301 & -0.176          & \textbf{-0.056}  & \underline{ -0.058} \\
 & 80\%  & -0.310 & -0.328 & -0.238 & -0.235           & -0.154      & -0.185 & -0.109          & -0.233 & -0.123          & -0.157           & -0.161 & -0.203 & -0.188 & -0.201 & -0.232 & -0.138          & -0.142 & -0.143 & -0.133         & -0.263 & -0.329 & -0.244          & \textbf{-0.066}  & \underline{ -0.074}  \\
 & 100\% & -0.353 & -0.375 & -0.332 & -0.309           & -0.226      & -0.271 & -0.179          & -0.315 & -0.235          & -0.263           & -0.210 & -0.286 & -0.269 & -0.285 & -0.305 & -0.198          & -0.247 & -0.226 & -0.203         & -0.318 & -0.357 & -0.351          & \textbf{-0.080}  & \underline{ -0.094}  \\
\midrule
\multirow{5}{*}{\textbf{Multi Classification}}  & 20\%  & -0.047 & -0.043 & -0.043 & -0.020           & -0.015      & -0.023 & \underline{ -0.002}    & -0.034 & -0.019          & -0.012           & -0.025 & -0.030 & -0.015 & -0.025 & -0.017 & -0.008          & -0.031 & -0.017 & -0.009         & -0.046 & -0.087 & \textbf{0.056}  & -0.007           & -0.135 \\
 & 40\%  & -0.144 & -0.125 & -0.123 & -0.069           & -0.052      & -0.065 & \underline{ -0.023}    & -0.090 & -0.049          & -0.044           & -0.070 & -0.082 & -0.071 & -0.067 & -0.067 & -0.026          & -0.095 & -0.055 & -0.032         & -0.126 & -0.206 & -0.101          & \textbf{-0.017}  & -0.137 \\
 & 60\%  & -0.274 & -0.228 & -0.232 & -0.132           & -0.097      & -0.123 & \textbf{-0.045} & -0.171 & -0.096          & -0.084           & -0.108 & -0.150 & -0.145 & -0.135 & -0.145 & \underline{ -0.045}    & -0.192 & -0.102 & -0.056         & -0.221 & -0.344 & -0.217          & -0.103           & -0.123 \\
 & 80\%  & -0.398 & -0.342 & -0.374 & -0.228           & -0.178      & -0.203 & \underline{ -0.084}    & -0.279 & -0.164          & -0.130           & -0.165 & -0.236 & -0.262 & -0.216 & -0.272 & \textbf{-0.077} & -0.320 & -0.164 & -0.086         & -0.355 & -0.462 & -0.291          & -0.314           & -0.139 \\
 & 100\% & -0.552 & -0.496 & -0.516 & -0.388           & -0.287      & -0.360 & \underline{ -0.143}    & -0.488 & -0.347          & -0.232           & -0.270 & -0.423 & -0.383 & -0.362 & -0.464 & \textbf{-0.105} & -0.440 & -0.275 & -0.150         & -0.525 & -0.620 & -0.429          & -0.245           & -0.176 \\
\midrule
\multirow{5}{*}{\textbf{Regression}}            & 20\%  & 0.237  & 0.233  & 0.250  & \textbackslash{} & \underline{ 0.001} & 0.028  & 0.001           & 0.054  & 0.001           & \textbackslash{} & 0.004  & 0.038  & 0.012  & 0.039  & 0.007  & 0.003           & 0.017  & 0.013  & 0.001          & 0.022  & 0.163  & \textbf{-0.233} & \textbackslash{} & \textbackslash{} \\
 & 40\%  & 0.599  & 0.592  & 0.642  & \textbackslash{} & \underline{ 0.003} & 0.076  & 0.003           & 0.133  & 0.003           & \textbackslash{} & 0.018  & 0.109  & 0.034  & 0.102  & 0.025  & 0.005           & 0.051  & 0.038  & \textbf{0.002} & 0.064  & 0.369  & 0.444           & \textbackslash{} & \textbackslash{} \\
 & 60\%  & 0.793  & 0.840  & 0.916  & \textbackslash{} & \underline{ 0.004} & 0.128  & 0.005           & 0.208  & 0.005           & \textbackslash{} & 0.140  & 0.196  & 0.063  & 0.180  & 0.049  & 0.009           & 0.087  & 0.050  & \textbf{0.002} & 0.119  & 0.559  & 0.595           & \textbackslash{} & \textbackslash{}  \\
 & 80\%  & 1.159  & 1.197  & 1.345  & \textbackslash{} & 0.007       & 0.184  & 0.007           & 0.293  & \underline{ 0.006}     & \textbackslash{} & 0.027  & 0.294  & 0.095  & 0.244  & 0.078  & 0.016           & 0.131  & 0.066  & \textbf{0.003} & 0.244  & 0.795  & 0.359           & \textbackslash{} & \textbackslash{}   \\
 & 100\% & 1.405  & 1.490  & 1.669  & \textbackslash{} & 0.011       & 0.250  & \underline{ 0.009}     & 0.380  & 0.013           & \textbackslash{} & 0.029  & 0.377  & 0.163  & 0.317  & 0.112  & 0.018           & 0.167  & 0.059  & \textbf{0.006} & 0.392  & 1.000  & 0.669           & \textbackslash{} & \textbackslash{} \\
\bottomrule
\end{tabular}}

\end{small}
\end{center}
\end{table*}

\label{4}
\subsection{Models Benchmarked}
\label{4.1}
We evaluate a suite of models designed for tabular data, drawing from four categories: tree-based models, deep-learning models, LLMs and tabular LLMs. Appendix~\ref{appendix:C} provides details of all models and their hyperparameters. 

\paragraph{Tree-Based Models.}
Gradient Boosting Decision Tree (GBDT) is a traditional type of tree-based models that incrementally improve the model by sequentially adding decision trees, optimizes the loss function using gradient descent, and prevents overfitting through regularization and tuning parameters. GBDTs are considered to be state-of-the-art models on tabular tasks~\cite{grinsztajn2022tree}. Hence, we evaluate LightGBM~\cite{Ke2017LightGBMAH}, XGBoost~\cite{10.11452939672.2939785} and CatBoost~\cite{prokhorenkova2019catboostunbiasedboostingcategorical} of GBDTs. 

\paragraph{Deep-Learning Models.}
Deep-learning models we evaluate consist of MLP, FT-Transformer~\cite{gorishniy2021revisiting}, TabPFN~\cite{Hollmann2022TabPFNAT}, and other tabular deep-learning models provided by LAMDA-TALENT$ \footnote{https://github.com/qile2000/LAMDA-TALENT} $, including AutoInt~\cite{song2019autoint}, TabNet~\cite{arik2020tabnetattentiveinterpretabletabular}, Tabular ResNet~\cite{gorishniy2021revisiting}, DCN2~\cite{wang2021dcn}, NODE~\cite{popov2019neuralobliviousdecisionensembles}, GrowNet~\cite{badirli2020gradientboostingneuralnetworks}, DANets~\cite{2021DANets}, Saint~\cite{somepalli2021saintimprovedneuralnetworks}, Snn~\cite{Klambauer2017SelfNormalizingNN}, Switchtab~\cite{wu2024switchtabswitchedautoencoderseffective}, Tabcaps~\cite{Chen2023TabCapsAC}, Tabr~\cite{Gorishniy2023TabRTD}, TabTransformer~\cite{huang2020tabtransformertabulardatamodeling}, Tangos~\cite{jeffares2023tangosregularizingtabularneural} and modernNCA~\cite{ye2024modernneighborhoodcomponentsanalysis}.
\paragraph{LLMs.}
We select Llama3-8B, a LLM released by Meta AI in April 2024, as the representative LLM for our evaluation. To construct the input text for Llama3-8B, we employ the \textbf{List Template} format, leveraging the demonstrated proficiency of LLMs in reading and parsing structured list-based inputs, as evidenced by prior research~\cite{hegselmann2023tabllmfewshotclassificationtabular}. A comprehensive explanation of the List Template is provided in Appendix~\ref{appendix:C.3}.
\paragraph{Tabular LLMs.}
LLMs have demonstrated remarkable performance in zero-shot and few-shot tabular tasks leading to the development of tabular LLMs which are specifically designed based on LLMs for tabular tasks. We evaluate TabLLM~\cite{hegselmann2023tabllmfewshotclassificationtabular} and Unipredict~\cite{wang2023unipredict}. About Unipredict, we choose the light version instead of the heavy version, because we observe that Unipredict-light can achieve better performance in the original paper and our own evaluations. The rest of model settings follow the original paper.

\subsection{Pipelines}
\paragraph{Data Set Segmentation.}
We begin by partitioning the dataset into a train\&validation set and a set of test sets. Appendix~\ref{appendix:B} shows the segmentation details of each dataset, including Pearson correlation heat maps of datasets. We uniformly use the same feature segmentation and data preprocessing methods.
\paragraph{Hyperparameter Optimization.}
We use hyperparameter optimization to help models achieve optimal performance in different datasets. In Appendix~\ref{appendix:C}, we provide full hyperparameter grids for each model.
\paragraph{Evaluation Metrics.}
For classification tasks, we utilize accuracy and ROC-AUC as model performance, where higher values denote superior model performance. For regression tasks, we utilize Root Mean Square Error (RMSE) as model performance, where lower values denote superior model performance. We also consider the percentage of model performance gap \textbf{$\Delta$} as model robustness in feature-shift scenarios, 
\begin{equation}
\Delta  = \frac{(metric_i- metric_0 )}{metric_0}
\label{delta_equation}
\end{equation}
$metric_i$ denotes the model performance where $i$ features shift. In subsequent sections, we use $metric$ to refer to \textbf{performance}, and $\Delta$ to refer to \textbf{robustness}.

\begin{table*}
\caption{Average rank in different tasks. We attribute the different feature-shift degree for each dataset to 20\%, 40\%, 60\%, and 80\%. Then compute the model's performance rank for each degree of feature shifts, task by task. 0\% means the model ID performance. '$\backslash$' means this model can't handle the regression task. For classification tasks, we choose accuracy. For regression tasks, we choose RMSE. The best is in \textbf{bold} and second best is \underline{underlined}. Superior accuracy in classification tasks is in \textit{italics}. Model abbreviations are in Appendix~\ref{appendix:C}.}
\label{table3}
\begin{center}
\begin{small}

\resizebox{\textwidth}{24mm}{
\begin{tabular}{cccccccccccccccccccccccccc}
\toprule
Task                                            & Shift & LGB  & XGB  & CATB         & PFN              & DAN  & MLP & NODE & RES & SWI  & CAP              & NET  & GOS       & INT  & DCN & FTT  & GRO  & SAT & SNN  & TTF  & TABR & NCA  & LMA  & TLLM               & UNI            \\
\midrule
\multirow{6}{*}{\textbf{Binary Classification}} & 0\%   & 5        & 3       & 2            & 7                & 19     & 13  & 22   & 11     & 14        & 17               & 21     & 10        & 18      & 12    & 16             & 23      & 8     & 20   & 24             & 4    & 1         & 6         & 9                  & 15                    \\
 & 20\%  & 10       & 13      & 1            & 5                & 18     & 9   & 23   & 8      & 12        & 17               & 19     & 6         & 16      & 11    & 15             & 20      & 2     & 21   & 24             & 14   & 22        & 4         & 7                  & 3                     \\
 & 40\%  & 21       & 23      & 4            & 13               & 15     & 7   & 20   & 9      & 5         & 11               & 17     & 6         & 10      & 8     & 14             & 19      & 1     & 18   & 22             & 16   & 24        & 12        & 3                  & 2                     \\
 & 60\%  & 22       & 23      & 5            & 12               & 10     & 6   & 16   & 14     & 4         & 8                & 19     & 7         & 11      & 9     & 15             & 18      & 2     & 17   & 20             & 21   & 24        & 13        & 3                  & 1                     \\
 & 80\%  & 22       & 24      & 5            & 12               & 7      & 8   & 14   & 16     & 4         & 6                & 21     & 9         & 10      & 11    & 18             & 17      & 3     & 15   & 20             & 19   & 23        & 13        & 1                  & 2                     \\
 & 100\% & 23       & 24      & 8            & 9                & 5      & 6   & 7    & 18     & 4         & 11               & 17     & 10        & 13      & 12    & 19             & 14      & 3     & 16   & 22             & 15   & 21        & 20        & 1                  & 2                     \\
\midrule
\multirow{6}{*}{\textbf{Multi Classification}}  & 0\%   & 6        & 5       & 2            & 7                & 20     & 15  & 22   & 8      & 12        & 18               & 19     & 14        & 21      & 10    & 16             & 24      & 13    & 17   & 23             & 9    & 1         & 11        & 3                  & 4                     \\
 & 20\%  & 10       & 8       & 4            & 16               & 20     & 14  & 23   & 7      & 13        & 19               & 21     & 11        & 18      & 12    & 15             & 24      & 6     & 17   & 22             & 5    & 1         & 9         & 3                  & 2                     \\
 & 40\%  & 15       & 13      & 3            & 14               & 20     & 11  & 23   & 9      & 5         & 17               & 21     & 10        & 19      & 6     & 12             & 24      & 7     & 16   & 22             & 8    & 4         & 18        & 2                  & 1                     \\
 & 60\%  & 20       & 15      & 6            & 10               & 17     & 7   & 22   & 8      & 3         & 14               & 18     & 5         & 19      & 4     & 13             & 24      & 11    & 9    & 21             & 12   & 16        & 23        & 2                  & 1                     \\
 & 80\%  & 24       & 19      & 17           & 9                & 15     & 7   & 12   & 10     & 2         & 6                & 13     & 8         & 21      & 3     & 14             & 23      & 16    & 5    & 11             & 18   & 22        & 20        & 4                  & 1                     \\
 & 100\% & 24       & 18      & 17           & 11               & 13     & 9   & 5    & 19     & 6         & 3                & 10     & 14        & 20      & 8     & 22             & 12      & 15    & 4    & 7              & 21   & 23        & 16        & 2                  & 1                     \\
\midrule
\multirow{6}{*}{\textbf{Regression}}            & 0\%   & 2        & 3       & 1            & \textbackslash{} & 18     & 9   & 17   & 6      & 20        & \textbackslash{} & 16     & 5         & 12      & 8     & 13             & 15      & 10    & 14   & 19             & 7    & 4         & 11        & \textbackslash{}   & \textbackslash{}      \\
 & 20\%  & 3        & 4       & 1            & \textbackslash{} & 18     & 10  & 16   & 8      & 19        & \textbackslash{} & 20     & 7         & 11      & 9     & 13             & 15      & 12    & 14   & 17             & 6    & 5         & 2         & \textbackslash{}   & \textbackslash{}      \\
 & 40\%  & 3        & 4       & 1            & \textbackslash{} & 17     & 10  & 16   & 7      & 20        & \textbackslash{} & 19     & 6         & 11      & 8     & 12             & 15      & 13    & 14   & 18             & 5    & 2         & 9         & \textbackslash{}   & \textbackslash{}      \\
 & 60\%  & 3        & 6       & 1            & \textbackslash{} & 19     & 10  & 17   & 8      & 20        & \textbackslash{} & 15     & 7         & 11      & 9     & 12             & 16      & 13    & 14   & 18             & 5    & 2         & 4         & \textbackslash{}   & \textbackslash{}      \\
 & 80\%  & 8        & 20      & 2            & \textbackslash{} & 17     & 5   & 14   & 4      & 19        & \textbackslash{} & 12     & 3         & 11      & 7     & 10             & 15      & 13    & 16   & 18             & 1    & 6         & 9         & \textbackslash{}   & \textbackslash{}      \\
 & 100\% & 15       & 19      & 6            & \textbackslash{} & 11     & 3   & 10   & 2      & 14        & \textbackslash{} & 8      & 1         & 18      & 4     & 7              & 12      & 17    & 9    & 13             & 5    & 16        & 20        & \textbackslash{}   & \textbackslash{}      \\
\midrule
\textbf{Top 3}                                  &       & 4        & 2       & 9            & 0                & 0      & 1   & 0    & 1      & 2         & 1                & 0      & 2         & 0       & 1     & 0              & 0       & 5     & 0    & 0              & 1    & 5         & 1         & \underline{ 9}            & \textbf{10}           \\
\midrule
\textbf{Average Rank}                           &       & 13.1     & 13.6    & \textbf{4.8} & 10.4             & 15.5   & 8.8 & 16.6 & 9.6    & 10.9      & 12.3             & 17.0   & \underline{ 7.7} & 15.0    & 8.4   & 14.2           & 18.3    & 9.2   & 14.2 & 18.9           & 10.6 & 12.1      & 12.2      & \underline{ \textit{3.3}} & \textit{\textbf{2.9}}\\
\bottomrule
\end{tabular}}

\end{small}
\end{center}
\vskip -0.1in
\end{table*}

\section{Experiment Results}
\label{5}
By conducting four different types of experiments to compare the performance and robustness of models in feature-shift scenarios, we demonstrate three main observations and corresponding future work below, which are detailed below. Comprehensive experimental details are provided in Appendix~\ref{appendix:E}. 

\vspace{5pt}
\begin{minipage}{\columnwidth}
\begin{tcolorbox}[colback=white, colframe=blue!45!black, title=Observation 1, fonttitle=\bfseries, rounded corners]
\textbf{Most models have the limited applicability in feature-shift scenarios.}
\end{tcolorbox}
\end{minipage}
\vspace{5pt}

To systematically investigate model robustness under varying degrees of distributional shift, we design controlled experiments by simulating four types of feature-shift scenarios. We employ $\Delta$ as a quantitative metric to assess model performance degradation across different shift magnitudes. Our empirical analysis yields three key observations.

\textbf{Feature shift impairs model robustness.} 
To rigorously assess the theoretical performance limits of the models under feature shift conditions, we conduct an empirical evaluation by training the models directly on the shifted data distribution and subsequently evaluating their generalization capability on a held-out test set. As illustrated in Figure~\ref{upper}, our experimental results reveal a consistent performance gap between tabular models trained on the original dataset versus those trained on the feature-shift dataset. The comparative analysis demonstrates that while training on the original dataset offers the advantage of broader feature representation, the inherent distributional shift substantially diminishes model robustness. 

\textbf{Most models can't handle feature shifts well.} Table~\ref{table2} indicates that most models struggle to effectively handle the challenge of feature shifts. As the degree of feature shift increases, the performance gap becomes wider. Tree-based models are particularly susceptible to negative impacts from feature shifts. In the classification task, NODE and tabular LLMs have less performance degradation, while in the regression task, DANets and TabTransformer exhibit greater resilience.  Although tabular LLMs are currently limited to classification tasks, they show strong adaptability to feature shifts.

\textbf{No Model Consistently Outperforms.} Table~\ref{table3} reveals that no model can consistently outperform in feature-shift scenarios. Although tabular LLMs display notable robustness, their performance in closed environments and low-degree feature-shift scenarios is not superior. CatBoost continues to highlight its effectiveness in solving tabular tasks. Moreover, the limited performance of NODE, DANets, and TabTransformer, which demonstrate outstanding robustness in Table~\ref{table2}, confirm that some models' robustness is achieved at the expense of closed environment performance.

The potential of tabular LLMs suggests that future work should focus on developing multi-layered fine-tuning frameworks to enhance semantic parsing capabilities of tabular LLMs for features and improve their reasoning abilities in various feature-shift scenarios. Additionally, integrating adaptive prompting mechanisms that dynamically adjust to the presence of feature shifts could further strengthen model robustness. By embedding domain knowledge and causal reasoning into the prompting strategy, LLMs are able to achieve a deeper understanding of the implications of feature shifts, thereby generating more accurate and reliable predictions.

\begin{figure*}
\begin{center}
\centerline{\includegraphics[width=\textwidth]{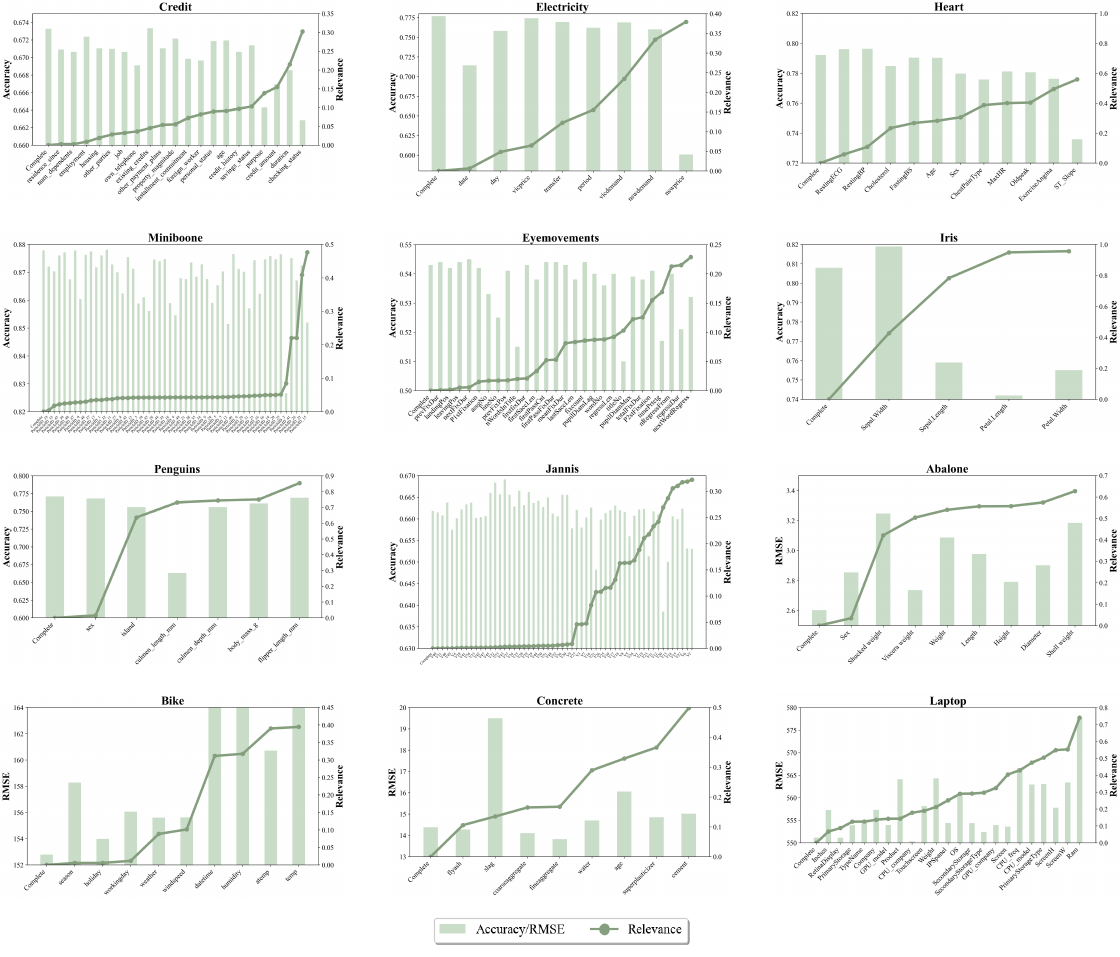}}
\caption{Results for single shift experiments with correlation analysis experiments across 12 datasets. The $x$-axis represents features ordered in ascending correlation with the target, the left $y$-axis represents the average performance (accuracy or RMSE) of all models and the right $y$-axis shows the absolute value of correlation. Features are removed in ascending order of their correlation values to observe the impact on model performance.}
\label{figure2}
\end{center}
\end{figure*}

\begin{minipage}{\columnwidth}
\begin{tcolorbox}[colback=white, colframe=blue!45!black, title=Observation 2, fonttitle=\bfseries, rounded corners]
\textbf{Shifted features' importance has a linear trend with model performance degradation.}
\end{tcolorbox}
\end{minipage}
\vspace{8pt}

Considering that feature shifts exert negative impacts on models from Observation 1, it is necessary to determine whether there is a relationship between the importance of shifted features and model performance degradation. To this end, we conduct single shift experiments by analyzing the average performance of all models for each dataset, and most/least-relevant shift experiments by comparing the correlation sum of shifted feature sets with model performance degradation in feature-shift scenarios. We obtain two observations from these experiments.

\textbf{Single strong correlated feature shifted causes greater model performance degradation.} Figure~\ref{figure2} illustrates that model performance decreases more significantly as the shifted single feature becomes more relevant to the target. The extent of model degradation caused by a strongly correlated feature shift is larger than that caused by a weakly correlated feature shift. We also observe that model performance may potentially improve if features that are less relevant to the target are removed, whose detailed analysis can be found in Appendix~\ref{appendix:E.2}.

\begin{figure*}
\begin{center}
 \begin{minipage}{0.48\linewidth}
 \includegraphics[width=1\linewidth]{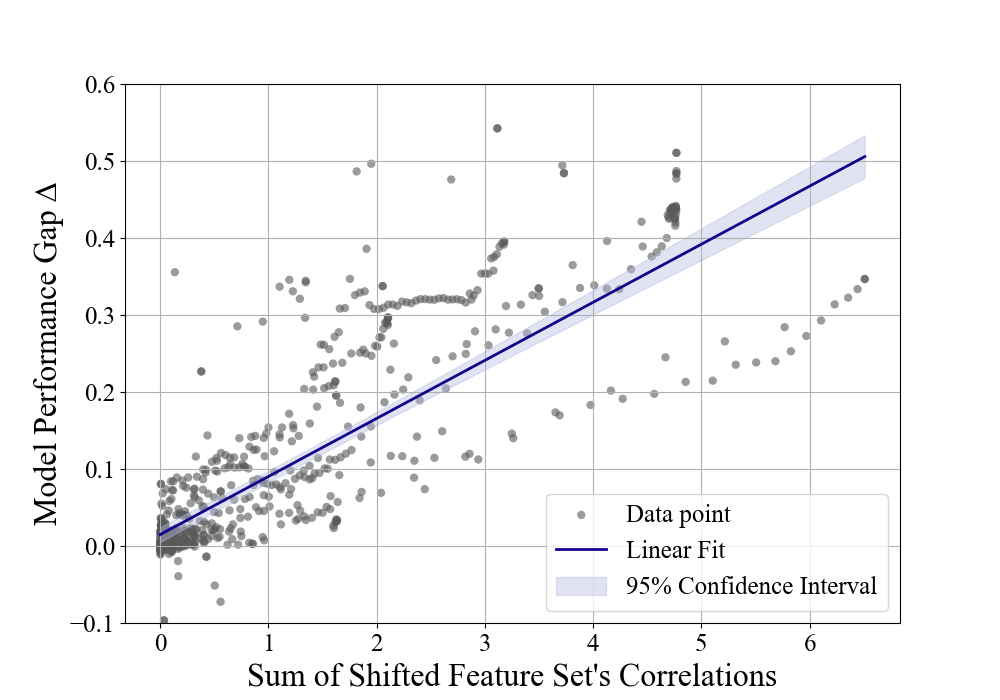} 
\caption{We use $\Delta$ (described in equation~\ref{delta_equation}) to measure the performance decrease. Sum of shifted feature set's correlations refers to the sum of Pearson correlation coefficients of shifted features. Notably, performance decrease and sum of shifted feature set's correlations demonstrate a strong correlation, with a Pearson correlation coefficient of $\rho$ = 0.7405.} 
\label{figure3}
 \end{minipage}
 \hfill
\begin{minipage}{0.48\linewidth}
\includegraphics[width=1\linewidth, trim=6cm 0cm 6cm 0cm, clip]{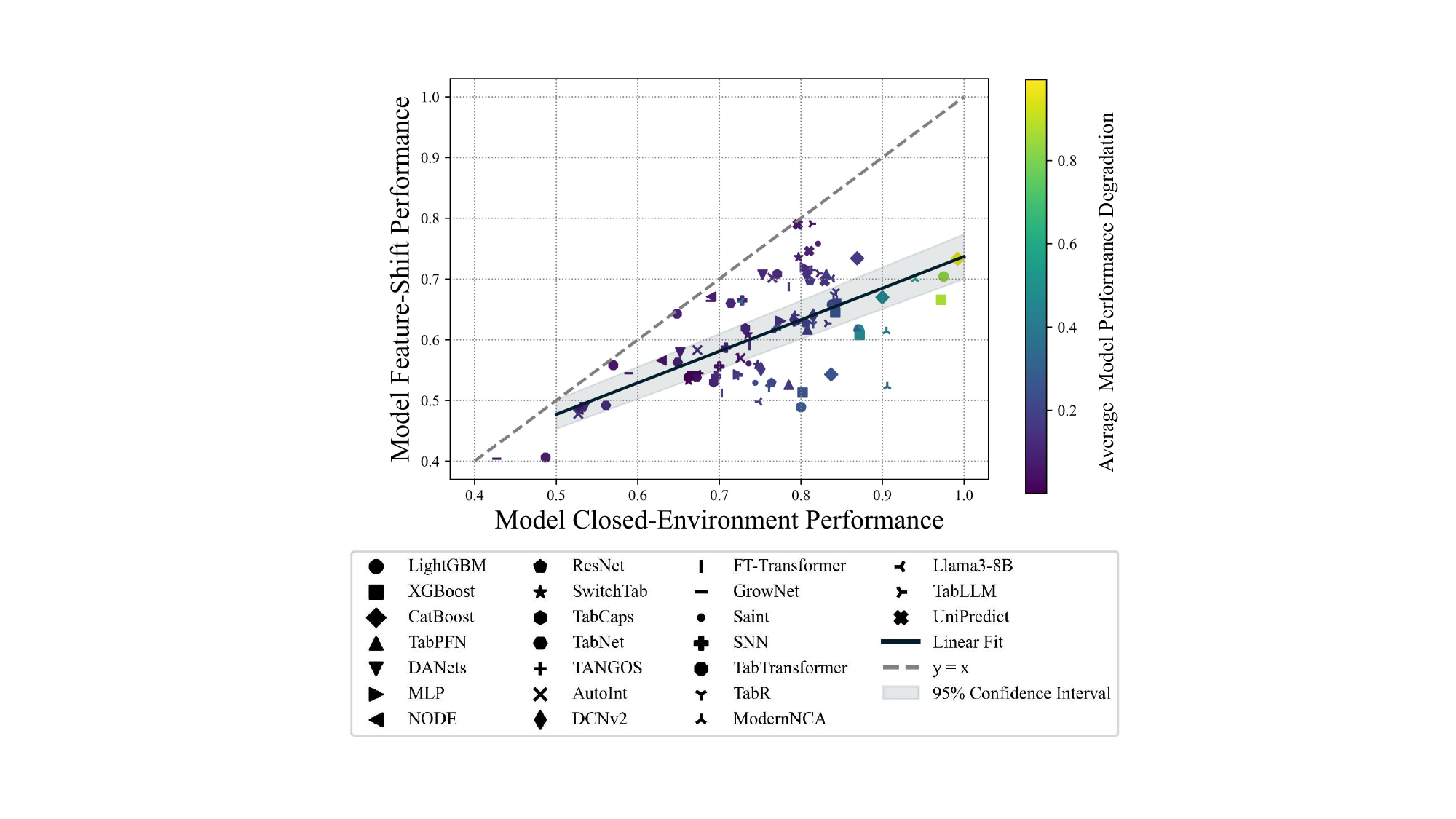} 
\vskip -0.3in
\caption{Model performance in closed environments vs. model feature-shift performance. \textbf{Closed environment} means that the dataset does not have any degree of feature shift. \textbf{Feature-shift} means average model performance in all degrees of feature shifts.} 
\label{figure4}
 \end{minipage}
\end{center}
\vskip -0.1in
\end{figure*}

\textbf{Shifted feature set's correlations have a relationship with model performance degradation linearly.} Figure~\ref{figure3} shows that as the correlation sum of the feature set gets larger, the model performance degradation gets larger. We observe a linear trend between the correlations of the shifted feature set and model performance degradation (see the blue line in Figure~\ref{figure3}; $\rho$ = 0.74).

The linear trend between the correlation sum of shifted features and model performance degradation underscores the importance of strongly correlated features. Therefore, it is essential to develop feature importance-driven optimization algorithms that incorporate adaptive mechanisms, such as dynamic feature weighting or hierarchical feature selection, to emphasize strong-correlated features. Additionally, this highlights the necessity of mitigating shifts in strong-correlated features in open environments.

\vspace{3pt}
\begin{minipage}{\columnwidth}
\begin{tcolorbox}[colback=white, colframe=blue!45!black, title=Observation 3, fonttitle=\bfseries, rounded corners]
\textbf{Model closed environment performance correlates with feature-shift performance.}
\end{tcolorbox}
\end{minipage}
\vspace{3pt}

In Figure~\ref{figure4}, a comparative analysis of model performance is presented, contrasting the outcomes in closed environments with those in feature-shift scenarios within the context of random shift experiments. The analysis reveals a notable trend: models that exhibit superior performance in closed environments tend to demonstrate relatively enhanced performance when subjected to feature-shift scenarios. This observation suggests the existence of a positive correlation between model performance in closed environments and their robustness in feature-shift scenarios. Specifically, the enhanced performance in closed environments appears to confer a degree of resilience to the models when confronted with the challenges posed by feature-shift scenarios, thereby highlighting the potential interdependence between these two distinct operational contexts.

This observation suggests that improving model performance in closed environments may serve as a foundational step to enhance their adaptability and robustness in feature-shift scenarios. A rigorous theoretical investigation into the relationship between closed environment performance and open environments adaptability is warranted, with particular emphasis on elucidating the underlying mechanisms that govern model robustness. Further exploration into methodologies for enhancing model resilience in open environments, such as through domain adaptation, transfer learning, or robustness-aware training paradigms.


\section{Conclusion}
We introduce TabFSBench, a comprehensive and rigorously designed benchmark specifically tailored to systematically investigate feature shifts in tabular data. TabFSBench encompasses diverse tasks, enabling the evaluation of model performance and robustness, and benchmarking of tabular models in feature-shift scenarios. To enhance accessibility and ensure reproducibility, we provide intuitive and user-friendly Python APIs, facilitating seamless dataset retrieval and integration into experimental workflows. Additionally, we conduct extensive empirical evaluations across four distinct feature-shift scenarios. Our three observations not only underscore the significant challenges posed by feature shifts but also offer insights for the future development of feature-shift research.

While this paper provides comprehensive evaluation of feature shift impacts on tabular models, several limitations warrant discussion. First, the current framework focuses primarily on feature decrement scenarios, as conventional tabular models inherently require fixed input dimensions and thus automatically ignore newly added features in increment scenarios. Second, our evaluation excludes specialized architectures designed for feature-increment issues. Third, the analysis does not examine how models respond to shifted features with identical correlation structures. Additionally, while we have evaluated diverse tabular tasks, the scope remains limited by the current absence of large-scale tabular LLM evaluations and comparative studies with other shift types. These aspects represent important directions for future empirical validation and benchmark expansion on feature-shift challenges in open environments.

\section*{Benchmark Availability Statement}
The benchmark code for this paper is available at \href{https://github.com/LAMDASZ-ML/TabFSBench}{https://github.com/LAMDASZ-ML/TabFSBench}. The project page of TabFSBench which contains the leaderboard is available at \href{https://clementcheng0217.github.io/Tab-index}{Home-TabFSBench}.

\section*{Acknowledgements}
This research is supported by the National Science Foundation of China (62306133, 624B2068), and the Key Program of Jiangsu Science Foundation (BK20243012). We would like to thank reviewers for their constructive suggestions.

\section*{Impact Statement}
The paper introduces a novel contribution aimed at advancing the burgeoning field of open environments. The observations elucidated within this work possess broad and multifaceted societal implications. However, due to the limitations of this paper, a detailed elaboration of these implications is reserved for future discourse.These discussions are expected to center on best practices and the development of regulatory frameworks that can effectively harness the benefits of open environments machine learning. 

\bibliography{main}
\bibliographystyle{icml2025}

\newpage
\appendix
\onecolumn
\section{Related work}
\subsection{Open Environments Challenges}
Traditional machine learning research predominantly assumes closed environment scenarios, where key factors in the learning process remain stable and predictable~\cite{guo2025robust, jia2024realistic}. Although machine learning has achieved remarkable success across various applications, an increasing number of real-world tasks—particularly those situated in open environments—face dynamic changes in critical factors. These challenges have given rise to the field of open environments Machine Learning (Open ML). \citet{zhou2022open} identified four core challenges in Open ML: the emergence of new classes, data distribution shifts, evolving learning objectives, and feature space shifts.

The emergence of new classes refers to the occurrence of previously unseen categories during the testing phase that were absent during training. Data distribution shift pertains to changes in the distribution of test data, violating the traditional assumption that data are independent and identically distributed (i.i.d.). Evolving learning objectives arise as data volume increases and model accuracy improves, prompting a shift in focus from maximizing accuracy to addressing additional priorities, such as minimizing energy consumption. Feature space shift is characterized by the introduction of new features or the removal of existing ones. Conventional machine learning models, which rely on the assumption that training and testing data share the same feature space, often struggle to adapt to such changes, leading to substantial performance degradation in open environments scenarios. To address these challenges, recent research has primarily centered on domain adaptation and domain generalization. Domain adaptation methods utilize a limited amount of labeled or unlabeled target-domain data to transfer knowledge effectively \citep{shen2023balancing}. In contrast, domain generalization focuses on training models with robust generalization capabilities to maintain high performance across diverse target domain distributions. For instance, the MixStyle method \citep{zhou2021domaingeneralizationmixstyle} improves generalization by blending statistical properties, such as mean and standard deviation, from different domains. More recently, Test-Time Adaptation (TTA) has been proposed as a promising approach to address data distribution shifts. TTA enables models to adapt dynamically during inference by leveraging test data batches for continual learning \citep{liang2024comprehensive}. However, much of the research on these approaches has primarily targeted non-tabular domains, such as computer vision and natural language processing \citep{miller2020effect}. Moreover, these methods often fail to surpass the performance of traditional optimization algorithms, such as Stochastic Gradient Descent (SGD) \citep{ruder2017overviewgradientdescentoptimization}, highlighting the need for further advancements tailored to open environments challenges in tabular data.

Although existing Heterogeneous Domain Adaptation (HeDA) methods have achieved significant progress on feature shift of images, tabular data presents a fundamentally different pattern. The inherent structure of it presents challenges when attempting to directly implement HeDA on such datasets. Moreover, our review of the literature reveals that what is often referred to as "feature shift" in many papers is essentially a form of distribution shift. For example,~\citet{he2023domain} regards covariate shift as feature shift.

\subsection{Tabular Data in Machine Learning}
Tabular data, characterized by its structured and heterogeneous features, is extensively used across various fields, including medical diagnostics, financial analysis, recommendation systems, and social sciences \citep{Borisov_Leemann_Sessler_Haug_Pawelczyk_Kasneci_2022,kadra2021well}. Unlike domains such as computer vision or natural language processing, tabular data poses unique challenges to machine learning models due to its high dimensionality, heterogeneity, and the complex interdependencies among features \citep{fang2024large}. Current approaches for modeling tabular data can be broadly classified into two main categories: tree-based ensemble models and deep learning models.

Tree-based ensemble models, such as XGBoost \citep{chizat2020faster}, LightGBM \citep{badirli2020gradientboostingneuralnetworks}, and CatBoost \citep{prokhorenkova2019catboostunbiasedboostingcategorical}, have long been regarded as the state-of-the-art for tabular data modeling. These models excel in handling irregular patterns and non-informative features within the objective function and are well-suited for addressing the non-rotation invariance of tabular data \citep{grinsztajn2022tree}. Their robustness and interpretability further contribute to their widespread adoption. In contrast, the rise of deep learning has spurred the development of numerous deep learning-based models tailored for tabular data. Notable examples include DCN V2 \citep{wang2021dcn}, which integrates multi-layer perceptron (MLP) modules and feature crossover modules; AutoInt \citep{song2019autoint} and FT-Transformer \citep{gorishniy2021revisiting}, both of which leverage Transformer architectures; ResNet-based tabular variants \citep{gorishniy2021revisiting}; and differentiable tree-based heuristics, such as Neural Oblivious Decision Ensembles (NODE) \citep{popov2019neuralobliviousdecisionensembles}. These models aim to exploit the complex inter-feature dependencies inherent in tabular data to improve predictive performance. In close-environment scenarios, the performance of deep learning models on tabular data has not yet surpassed that of tree-based models. Furthermore, the generalization capabilities of both types of models in open environments, particularly when facing feature distribution shifts, have not been assessed.

\subsection{Benchmark in Tabular Data}
In machine learning, establishing effective benchmarks is essential for evaluating and comparing the performance of various algorithms. An ideal benchmark provides standardized datasets and evaluation criteria to ensure the effectiveness, reliability, and robustness of algorithms in practical applications~\cite{wang2022usb, jia2024lamda}. While benchmarks in domains such as computer vision (e.g., ImageNet \citep{fei2009imagenet}), natural language processing (e.g., GLUE \citep{wang-etal-2018-glue}), and audio classification (e.g., AudioSet \citep{gemmeke2017audio}) are relatively mature, benchmarks for tabular data remain underdeveloped. Many existing tabular data benchmarks suffer from significant quality issues. For instance, the German Credit dataset is small in scale, and the Adult dataset contains inherent biases and data quality problems \citep{Bao_Zhou_Zottola_Brubach_Desmarais_Horowitz_Lum_Venkatasubramanian_2021}. These limitations constrain their utility for conducting in-depth research and hinder the development of robust machine learning methods for tabular data.

Furthermore, benchmarks capable of systematically evaluating distribution shifts are crucial for assessing the robustness and adaptability of models to real-world data variations. Distribution shifts refer to the discrepancy between the data distribution during inference and that during training, which can severely impact model performance~\cite{zhou2023ods}. To address this, several benchmarks have been proposed to evaluate the handling of distribution shifts in tabular data. For example, Shifts and Shifts 2.0 \citep{malinin2021shifts,malinin2022shifts} focus on uncertainty estimation and include tasks with temporal and spatio-temporal variations in tabular data. WhyShift \citep{liu2024need} specializes in evaluating spatio-temporal shifts using five real-world tabular datasets. TableShift \citep{gardner2024benchmarking} provides a comprehensive evaluation of 19 model types across 15 binary classification tasks, with 10 tasks explicitly related to domain generalization. Similarly, Wild-Tab \citep{kolesnikov2023wild} emphasizes domain generalization in tabular regression tasks, comparing 10 domain generalization methods against standard Empirical Risk Minimization (ERM) applied to Multilayer Perceptrons (MLPs). However, despite these benchmarks offering comprehensive scenarios for addressing distribution shifts, their focus remains primarily on changes in data distributions, with limited attention to feature space shifts, such as the addition or removal of entire features in a dataset. Although \citet{gardner2024large} explored feature drop experiments, their evaluation was confined to XGBoost and their proposed model, lacking comprehensiveness. The absence of benchmarks that adequately account for feature space shifts hinders researchers from effectively evaluating and optimizing algorithms to address such challenges, which remains a critical gap in the field of tabular data processing.

\section{TabFSBench Components}

To promote systematic evaluation and community collaboration in the context of feature-shifted tabular data learning, we present TabFSBench, a modular benchmark framework. It features reproducible evaluation protocols, an extensible API interface, and a continuously updated public leaderboard.

\subsection{Benchmark Composition and Datasets}

TabFSBench currently comprises 12 datasets selected from Grinsztajn et al.~\cite{grinsztajn2022tree} and TabZilla~\cite{mcelfresh2023neural}. These datasets exhibit significant heterogeneity in terms of scale, domain, and task structure, covering a wide range of potential feature shift scenarios such as covariate shift and missing values. To comprehensively assess model robustness under real-world distributional changes, we design four distinct experimental configurations that systematically evaluate model generalization across various perturbations. All experiments are repeated with multiple random seeds to mitigate stochastic variance. We further aim to construct practically meaningful feature-shift datasets in future versions, thereby encouraging deeper investigation into robustness under feature shift.

\subsection{Public Leaderboard and Community Contribution}

We have established and actively maintain a project homepage and a complete ranking system. Evaluation results of newly released models or datasets are updated regularly to ensure that the research community can access the latest benchmark performance. For instance, TabFSBench already includes evaluation results for representative models such as TabPFN v2. We encourage researchers to submit evaluation results for their custom models or datasets. In future updates, we plan to support both public and private leaderboard modes to meet a broader range of user requirements.

\subsection{API Design}

To enable reproducibility and flexible evaluation, TabFSBench offers a command-line option \texttt{--export\_dataset}, which allows users to export dataset variants under different feature-shift settings (e.g., single-column missingness, controlled levels of missingness, and complete enumeration of missing scenarios) by setting the flag to \texttt{True}.

We also provide a \texttt{README.md} file, which describes how users can add new datasets and models. Further details on code functionality will be elaborated in the final version.

\section{Real-World Feature-Shift Challenges}
\subsection{Real-World Feature-Shift Datasets}
There currently exists no dataset specifically designed for feature shift, unlike Tableshift~\cite{gardner2024benchmarking} which was developed for distribution shifts. However, we have preliminarily constructed a feature-shifted dataset based on the Heart dataset. Given that different features in the original dataset require distinct measurement instruments, we categorized the features into three groups: basic features, electrocardiogram (ECG) features, and exercise stress test features.

In the constructed feature-shifted Heart dataset, both the training set and the test set step 0 contain all features. However, patients in the test set step 1 lack ECG measurements, resulting in the absence of RestingECG and ST\_Slope features. Similarly, patients in the test set step 2 did not undergo an exercise stress test, leading to the absence of ExerciseAngina and Oldpeak. A subset of examples is provided at \href{https://github.com/LAMDASZ-ML/TabFSBench}{https://github.com/LAMDASZ-ML/TabFSBench}.

Note that for meaningful evaluation of feature-shifted datasets, models must be assessed under specific partitioning schemes. Applying the four experimental settings proposed in our paper would undermine the unique characteristics and practical relevance of such datasets.

\subsection{Real-World Feature-Shift Issues}
Section 2.2 of this paper employs forest disease monitoring as a case study to demonstrate how sensor degradation leads to a reduction in available features. As further evidenced by the designed heart dataset in above, incomplete medical examinations may result in missing diagnostic indicators (features) due to the absence of specific equipment.

The feature shift phenomenon also manifests prominently in financial and transportation domains:
\begin{itemize}
    \item Finance: Stock prediction models trained on comprehensive features (e.g., financial ratios, macroeconomic indicators) may encounter missing features (e.g., market sentiment indices) during real-world deployment due to unforeseen events.
    \item Transportation: Accident prediction models relying on features like road conditions and weather data may experience partial feature absence caused by sensor failures or insufficient data collection.
\end{itemize}

\section{Benchmark Datasets}
\begin{figure*}[htbp]
\vskip 0.2in
\begin{center}
\centerline{\includegraphics[width=0.8\textwidth]{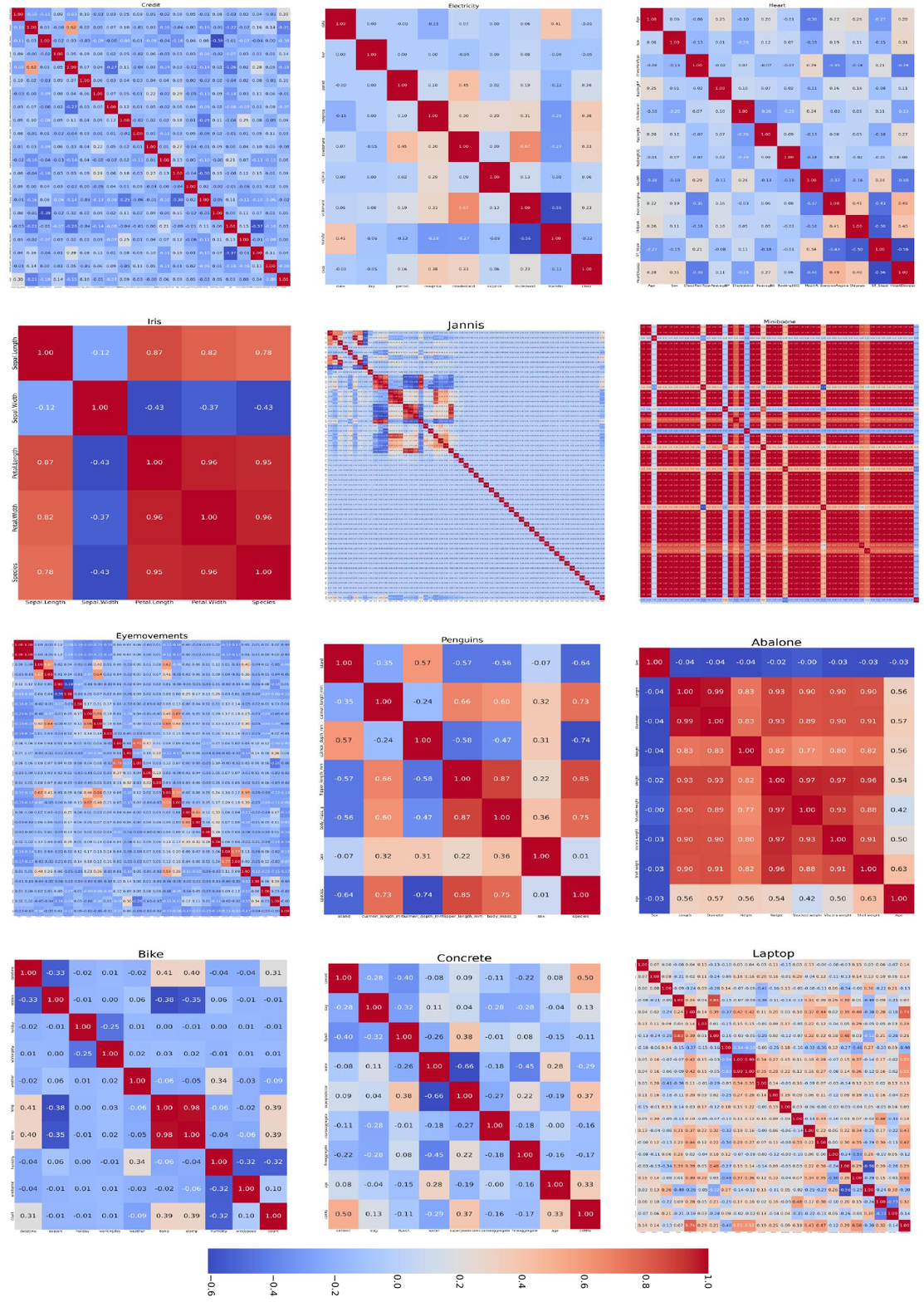}}
\caption{Pearson Correlation Analysis of the Datasets. Images are sorted in binary classification, multi-class classification, and regression order. The last column of each image is chosen as the input column with the target Pearson correlation coefficient.}
\label{pearson}
\end{center}
\vskip -0.2in
\end{figure*}

This section provides background information and the sources of each dataset in TabFSBench. Pearson correlation analyses of the datasets are also provided in Figure~\ref{pearson}.
\label{appendix:B}
\subsection{Binary Classification}
\paragraph{Credit}
The original dataset contains 1,000 entries with 20 categorical/symbolic attributes prepared by Prof. Hofmann. In this dataset, each entry represents a person who takes a credit from a bank. Each person is classified as having good or bad credit risk according to the set of attributes. The target is to determine whether the customer's credit is good or bad. This dataset is available at \href{https://www.openml.org/search?type=data\&sort=runs\&id=31\&status=active}{https://www.openml.org/search?type=data\&sort=runs\&id=31\&status=active}.
\paragraph{Electricity}
The Electricity dataset, collected from the Australian New South Wales Electricity Market, contains 45,312 instances from May 1996 to December 1998. Each instance represents a 30-minute period and includes fields for the day, timestamp, electricity demand in New South Wales and Victoria, scheduled electricity transfer, and a class label. The target is to predict whether the price in New South Wales is up or down relative to a 24-hour moving average, based on market demand and supply fluctuations. This dataset is available on \href{https://www.kaggle.com/datasets/vstacknocopyright/electricity}{https://www.kaggle.com/datasets/vstacknocopyright/electricity}.
\paragraph{Heart}
Cardiovascular diseases (CVDs) are the leading cause of death globally, responsible for 17.9 million deaths annually.  Heart failure is a common event caused by CVDs and this dataset contains 11 features that can be used to predict a possible heart disease. The target is to determine whether the patient's heart disease is present or absent. This dataset is available on \href{https://www.kaggle.com/datasets/fedesoriano/heart-failure-prediction
}{https://www.kaggle.com/datasets/fedesoriano/heart-failure-prediction
}.
\paragraph{Miniboone}  This dataset aims to construct a predictive model using various machine learning algorithms and document the end-to-end steps using a template. The MiniBooNE Particle Identification dataset is a binary classification task where we attempt to predict one of two possible outcomes. The target is to determine whether the neutrino is an electron or a muon. This dataset is available at \href{https://www.kaggle.com/datasets/alexanderliapatis/miniboone}{https://www.kaggle.com/datasets/alexanderliapatis/miniboone}.
\subsection{Multi-class Classification}
\paragraph{Iris}
The Iris flower dataset, introduced by Ronald Fisher in 1936, contains 150 samples from three Iris species: Iris setosa, Iris virginica, and Iris versicolor. Each sample has four features: sepal length, sepal width, petal length, and petal width, measured in centimeters. The target is to classify the Iris species as setosa, versicolor, or virginica. This dataset is available on \href{https://www.kaggle.com/datasets/uciml/iris}{https://www.kaggle.com/datasets/uciml/iris}.
\paragraph{Jannis} This dataset is used in the tabular benchmark from \cite{grinsztajn2022tree}. It belongs to the 'classification on numerical features' benchmark. The dataset is designed to test classification performance using numerical features, and it presents challenges such as varying data distributions, class imbalances, and potential missing values. It serves as a critical evaluation tool for machine learning models in real-world scenarios, including medical diagnosis, credit rating, and object recognition tasks. This dataset is available on \href{https://www.openml.org/search?type=data\&status=active\&id=45021}{https://www.openml.org/search?type=data\&status=active\&id=45021}.
\paragraph{Penguins}
Data were collected and made available by Dr. Kristen Gorman and the Palmer Station, Antarctica LTER, a member of the Long Term Ecological Research Network. The goal of the Palmer Penguins dataset is to offer a comprehensive resource for data exploration and visualization, serving as an alternative to the Iris dataset. The target is to classify the penguin species as Adelie, Chinstrap, or Gentoo. This dataset is available at \href{https://www.kaggle.com/datasets/youssefaboelwafa/clustering-penguins-species}{https://www.kaggle.com/datasets/youssefaboelwafa/clustering-penguins-species}.

\paragraph{Eye Movements}
This dataset is designed to predict the relevance of sentences in relation to a given question based on eye movement data. The target is to classify sentences as irrelevant, relevant, or correct, using 27 features, including landing position (landingPos), first fixation duration (P1stFixation), next fixation duration (nextFixDur), time spent on the predicted region (timePrtctg), and other relevant eye movement metrics. This dataset is available at \href{https://www.kaggle.com/datasets/vinnyr12/eye-movements}{https://www.kaggle.com/datasets/vinnyr12/eye-movements}.
\subsection{Regression}
\paragraph{Abalone}
The age of abalone is traditionally determined by cutting the shell, staining it, and counting the rings under a microscope, a process that is both tedious and time-consuming. This dataset uses easier-to-obtain physical measurements, such as length, diameter, and weight, to predict the abalone's age. The target is to predict the age, providing a more efficient approach.  This dataset is available on \href{https://www.kaggle.com/datasets/rodolfomendes/abalone-dataset}{https://www.kaggle.com/datasets/rodolfomendes/abalone-dataset}.
\paragraph{Bike}
The dataset records the rental of shared bikes in the Washington area from 2011-01-01 to 2012-12-31, containing 11 features such as season, holiday, working day, and weather conditions. The goal is to predict the total count of bikes rented each hour, with the target being to forecast the number of bicycles available for rent today based on historical rental patterns and external factors like temperature, humidity, and seasonal trends. This dataset is available on \href{https://www.kaggle.com/datasets/abdullapathan/bikesharingdemand}{https://www.kaggle.com/datasets/abdullapathan/bikesharingdemand}.
\paragraph{Concrete}
Concrete is the most important material in civil engineering, and its compressive strength is influenced by a highly nonlinear relationship with its ingredients and age. The dataset contains 9 attributes, including variables such as cement, water, and age. The target is to predict the concrete compressive strength (measured in MPa) using these input variables. This dataset is available on \href{https://www.kaggle.com/datasets/maajdl/yeh-concret-data}{https://www.kaggle.com/datasets/maajdl/yeh-concret-data}.
\paragraph{Laptop}
The original dataset was pretty compact with a lot of details in each column. The columns mostly consisted of long strings of data, which was pretty human-readable and concise but for Machine Learning algorithms to work more efficiently it's better to separate the different details into their own columns. After doing so, 28 duplicate rows were exposed and removed with this dataset being the final result. The target is to predict the price of this laptop. This dataset is available on \href{https://www.kaggle.com/datasets/owm4096/laptop-prices}{https://www.kaggle.com/datasets/owm4096/laptop-prices}.

\section{Benchmark Models}
In this section, we provide introductions to tree-based and deep learning models and hyperparameter grids in Table~\ref{Hyperparameter_tree_model},~\ref{Hyperparameter_dl_model}. Among them, TabPFN and tabcaps are only applicable to the classification task, not to the regression task. We implement adaptive hyperparameter optimization based on the Optuna framework and following previous studies~\cite{liu2024talenttabularanalyticslearning}, fixing the batch size at 1024 and conducting 100 independent trials through train-validation splits to prevent test set leakage, with the best-performed hyperparameters fixed during the final 15 seeds. For LLMs, we provide their prompts in Figure~\ref{prompt}.
\label{appendix:C}
\subsection{Tree-Based Models}
\paragraph{LightGBM} LightGBM\citep{badirli2020gradientboostingneuralnetworks} is a machine learning model based on the Boosting algorithm. Its core algorithms are Gradient-based One-Side Sampling (GOSS) and Exclusive Feature Bundling (EFB). GOSS calculates the importance of each sample through gradients, discards unimportant samples, and selects a subset of important samples for training. EFB is used to reduce the dimensionality of the feature space."
\paragraph{XGBoost} XGBoost\citep{chizat2020faster} is an efficient and flexible machine learning model that incrementally builds multiple decision trees by optimizing the loss function, with each tree correcting the errors of the previous one to continuously improve the model's predictive performance. XGBoost also incorporates the gradient boosting algorithm, iteratively training decision tree-based models with the goal of minimizing residuals and enhancing predictive accuracy.
\paragraph{CatBoost}
CatBoost \citep{prokhorenkova2019catboostunbiasedboostingcategorical} is a powerful boosting-based model designed for efficient handling of categorical features. It uses the "Ordered Boosting" technique, which calculates gradients sequentially to prevent target leakage and maintain the independence of each training instance. At the same time, CatBoost employs "Target-based Categorical Encoding," converting categorical variables into numerical representations based on target statistics, thereby reducing the need for extensive preprocessing and improving model performance.

\begin{table*}[t]
\begin{center}
\begin{small}

\caption{Hyperparameter Grids of Tree-based Models. }
\label{Hyperparameter_tree_model}
\vskip 0.1in
\begin{tabular}{ccc}
\toprule
\textbf{Model} & \textbf{Hyperparameter} & \textbf{Values} \\ 
\midrule
\multirow{4}{*}{\textbf{LightGBM}} 
& Num Leaves & $\{31, 127\}$ \\
& Learning Rate & $\{0.01, 0.1\}$ \\
& Min Data In Leaf & $\{20, 50, 100\}$ \\
& Min Sum Hessian In Leaf & $\{1e-3, 1e-2, 1e-1\}$ \\
\midrule
\multirow{6}{*}{\textbf{XGBoost}} 
& Learning Rate & $\{0.01, 0.1\}$ \\
& Max. Depth & $\{1, 5, 9\}$ \\
& N Estimators & $\{10000, 20000, 30000\}$ \\
& Subsample & $\{0.5, 0.8, 1.0\}$ \\
& Colsample Bytree & $\{0.5, 0.8, 1.0\}$ \\
& Min Child Weight & $\{1, 3, 5\}$ \\
\midrule
\multirow{3}{*}{\textbf{CatBoost}} 
& Learning Rate & $\{0.01, 0.05, 0.1\}$ \\
& Depth & $\{4, 6, 8\}$ \\
& Iterations & $\{500, 1000, 2000\}$ \\
\bottomrule
\end{tabular}

\end{small}
\end{center}
\vskip -0.1in
\end{table*}

\subsection{Deep Learning Models}
\paragraph{AutoInt} AutoInt\citep{song2019autoint} efficiently handles large-scale data by mapping numerical and categorical features into the same low-dimensional space and leveraging multi-head self-attentive neural networks to model feature interactions.
\paragraph{DANets} DANets\citep{2021DANets} enhance the feature representation capacity of tabular data by introducing Abstract Layers and shortcut paths, and employ structure re-parameterization to reduce computational complexity, demonstrating effectiveness and extendibility in tabular data tasks.
\paragraph{DCN2}
Deformable ConvNets v2 (DCN2) \citep{zhu2019deformable} enhances object detection and instance segmentation by adapting to the geometric variations of objects. Its reformulation improves the network's focus on relevant image regions through advanced modeling and modulation techniques. Additionally, a feature mimicking scheme is introduced to guide network training, leading to significant performance gains on the COCO benchmark.
\paragraph{FT-Transformer}
FT-Transformer \citep{gorishniy2021revisiting} is a Transformer-based model specifically designed to handle tabular data. It employs separate feature tokenizers for numerical and categorical data, enabling the Transformer to effectively capture complex relationships between features. This adaptation improves performance in structured data tasks.
\paragraph{GrowNet}
GrowNet \citep{badirli2020gradientboostingneuralnetworks} is a gradient boosting model that uses shallow neural networks as weak learners. It incorporates a fully corrective step to address the greedy function approximation issue, enhancing performance in classification, regression, and ranking tasks.
\paragraph{MLP} A Multi-Layer Perceptron (MLP) comprises multiple layers of fully connected neurons, typically including an input layer, one or more hidden layers, and an output layer. During training, the MLP iteratively updates the connection weights between neurons using optimization techniques such as backpropagation and gradient descent, aiming to minimize the prediction error and improve model generalization.
\paragraph{ModernNCA} ModernNCA \citep{ye2024modernneighborhoodcomponentsanalysis} is an enhanced Neighborhood Component Analysis (NCA) model that improves tabular data processing by adjusting learning objectives, integrating deep learning architectures, and using stochastic neighbor sampling for better efficiency and accuracy.
\paragraph{NODE} Neural Oblivious Decision Ensembles (NODE) \citep{popov2019neuralobliviousdecisionensembles} is a deep learning method that combines the Oblivious Ensembles algorithm with neural networks, enabling end-to-end gradient-based optimization and multi-layer hierarchical representation learning for tabular data tasks.
\paragraph{Saint} Based on the Transformer architecture, SAINT \citep{somepalli2021saintimprovedneuralnetworks} employs an enhanced embedding method to classify features better while performing attention over both rows and columns to improve the model's focus on relevant data for tabular problems.
\paragraph{SNN} SNN \citep{Klambauer2017SelfNormalizingNN} is a neural network designed to improve tabular data processing by using self-normalizing properties and the Scaled Exponential Linear Units (SELUs) activation function, enabling robust training of deep networks with layers and enhancing performance across various tasks.
\paragraph{SwitchTab} SwitchTab\citep{wu2024switchtabswitchedautoencoderseffective} is a self-supervised method that uses an asymmetric encoder-decoder framework to separate mutual and salient features in tabular data, generating more representative embeddings for improved prediction and classification performance.
\paragraph{TabCaps} TabCaps~\citep{Chen2023TabCapsAC} is a capsule-based neural network architecture that encapsulates all feature values of a record into structured vectorial representations, thereby enabling collective feature processing and simplifying the treatment of heterogeneous and high-dimensional tabular data for enhanced classification performance.
\paragraph{Tabnet} TabNet\citep{arik2020tabnetattentiveinterpretabletabular} enhances tabular data modeling by using sequential attention to select salient features at each decision step, enabling efficient learning and interpretability. It utilizes sparse attention mechanisms within decision steps to prioritize the most relevant features.
\paragraph{TabPFN} TabPFN\citep{Hollmann2022TabPFNAT} is a Transformer-based model that approximates the posterior predictive distribution for tabular data, enabling fast supervised classification with no hyperparameter tuning. It performs in-context learning, making predictions with labeled sequences without further parameter updates, and can be reused for downstream tasks without retraining.
\paragraph{TabR} TabR\citep{Gorishniy2023TabRTD} is a neural network that enhances tabular data processing by integrating a k-Nearest-Neighbors-like component. It uses an attention mechanism to efficiently retrieve neighbors and extract valuable information, boosting predictive performance.
\paragraph{TabTransformer} TabTransformer\citep{huang2020tabtransformertabulardatamodeling} is a deep learning model that leverages Transformer layers to learn contextual embeddings for categorical features and normalize continuous features, achieving higher prediction accuracy and robust performance against noisy or missing data.
\paragraph{tabular ResNet} 
Tabular ResNet\citep{gorishniy2021revisiting} introduces residual connections by combining the original input \(x\) with its transformed version \(f(x)\), meaning the output of each block is \(x + f(x)\), which improves gradient flow and captures complex feature interactions in tabular data.
\paragraph{TANGOS} A regularization framework that enhances tabular data modeling by encouraging orthogonalization and specialization of neuron attributions. TANGOS\citep{jeffares2023tangosregularizingtabularneural} leverages gradient attributions of neurons to input features to promote focus on sparse and non-overlapping features, thereby achieving diverse and specialized latent unit allocation and improving generalization performance.

\begin{longtable}{@{}c@{\hspace{22pt}}c@{\hspace{22pt}}c@{}}
\caption{Hyperparameter Grids of Deep Learning Models.}\\
\vspace{-20pt}\\
\label{Hyperparameter_dl_model}\\
\toprule
\textbf{Model} & \textbf{Hyperparameter} & \textbf{Values} \\
\midrule
\endhead
\midrule
\multicolumn{3}{c}{\textbf{Continued on next page}} \\
\endfoot
\bottomrule
\endlastfoot

\multirow{6}{*}{\textbf{AutoInt}} 
& N\_Layers & Int$\{1, 6\}$ \\
& D\_Token & $\{8, 16, 32, 64, 128\}$ \\
& Residual Dropout & Uniform $\{0.0, 0.2\}$ \\
& Attention Dropout & Uniform $\{0.0, 0.5\}$ \\
& Learning Rate & Loguniform$\{e^{-5}, 0.001\}$ \\
& Weight Decay & Loguniform$\{e^{-6}, 0.001\}$ \\
\midrule
\multirow{5}{*}{\textbf{DANets}} 
& N\_Layers & Int$\{6, 32\}$ \\
& Dropout & Uniform $\{0.0, 0.2\}$ \\
& Base Outdim & Int$\{64, 128\}$ \\
& Learning Rate & Loguniform$\{e^{-5}, 0.1\}$ \\
& Weight Decay & Loguniform$\{e^{-6}, 0.001\}$ \\
\midrule
\multirow{5}{*}{\textbf{FT-transformer}} 
& Num. Blocks & $\{1,2,3,4\}$ \\
& Residual Dropout & Uniform $\{0.0, 0.2\}$ \\
& Attention Dropout & Uniform $\{0.0, 0.5\}$ \\
& FFN Dropout & Uniform $\{0.0, 0.5\}$ \\
& FFN Factor & $\{64,128,256,512\}$ \\
\midrule
\multirow{6}{*}{\textbf{GrowNet}} 
& D\_Embedding & Int$\{32, 512\}$ \\
& Hidden\_D & Int$\{32, 512\}$ \\
& Learning Rate & Loguniform$\{e^{-5}, 0.1\}$ \\
& Weight Decay & Loguniform$\{e^{-6}, 0.001\}$ \\
& Epochs Per Stage & Int$\{1, 2\}$ \\
& Correct Epoch & Int$\{1, 2\}$ \\
\midrule
\multirow{4}{*}{\textbf{MLP}} 
& D\_layers & $\{1, 8, 64, 512\}$ \\
& Dropout & Uniform $\{0.0, 0.5\}$ \\
& Learning Rate & Loguniform$\{e^{-5}, 0.01\}$ \\
& Weight Decay & Loguniform$\{e^{-6}, 0.001\}$ \\
\midrule
\multirow{4}{*}{\textbf{ModernNCA}} 
& Dropout & Uniform $\{0.0, 0.5\}$ \\
& D\_block & Int$\{64, 1024\}$ \\
& N\_blocks & Int$\{0, 2\}$ \\
& N\_frequencies & Int$\{16, 96\}$ \\
& Frequency Scale & Loguniform$\{0.005, 10\}$ \\
& D\_embedding & Int$\{16, 64\}$ \\
& Sample Rate  & Uniform$\{0.05, 0.6\}$ \\
& Learning Rate & Loguniform$\{e^{-5}, 0.1\}$ \\
& Weight Decay & Loguniform$\{e^{-6}, 0.001\}$ \\
\midrule
\multirow{6}{*}{\textbf{NODE}} 
& Num Layers & Int$\{1, 4\}$ \\
& Depth & Int$\{4, 6\}$ \\
& Tree Dim & Int$\{2, 3\}$ \\
& Layer Dim & $\{512, 1024\}$ \\
& Learning Rate & Loguniform$\{e^{-5}, 0.1\}$ \\
& Weight Decay & Loguniform$\{e^{-6}, 0.001\}$ \\
\midrule
\multirow{8}{*}{\textbf{Saint}} 
& Depth & $\{4, 6\}$ \\
& Heads & $\{4, 8\}$ \\
& Dim & $\{16, 32, 64\}$ \\
& Attn\_Dropout  & Uniform $\{0.0, 0.5\}$ \\
& FF\_Dropout  & Uniform $\{0.0, 0.5\}$ \\
& Attentiontype & $\{"colrow", "row", "col"\}$ \\
& Learning Rate & Loguniform$\{3e^{-5}, 0.001\}$ \\
& Weight Decay & Loguniform$\{e^{-6}, 0.0001\}$ \\
\midrule
\multirow{5}{*}{\textbf{SNN}} 
& D\_layers & $\{2, 16, 1, 512\}$ \\
& Dropout & Uniform $\{0.0, 0.1\}$ \\
& D\_embedding & Int$\{64, 512\}$ \\
& Learning Rate & Loguniform$\{e^{-5}, 0.01\}$ \\
& Weight Decay & Loguniform$\{e^{-6}, 0.001\}$ \\
\midrule
\multirow{3}{*}{\textbf{SwitchTab}} 
& \(\alpha\) & Loguniform$\{0.01, 100\}$ \\
& Learning Rate & Loguniform$\{e^{-6}, 0.001\}$ \\
& Weight Decay & Loguniform$\{e^{-6}, 0.001\}$ \\
\midrule
\multirow{7}{*}{\textbf{TabCaps}} 
& Learning Rate & Loguniform$\{e^{-5}, 0.1\}$ \\
& Weight Decay & Loguniform$\{e^{-6}, 0.001\}$ \\
& Sub Class & Int$\{1, 5\}$ \\
& Init Dim & Int$\{32, 128\}$ \\
& Primary Capsule Size & Int$\{4, 32\}$ \\
& Digit Capsule Size & Int$\{4, 32\}$ \\
& Leaves & Int$\{16, 64\}$ \\
\midrule
\multirow{6}{*}{\textbf{Tabnet}} 
& Learning Rate & uniform$\{0.001, 0.01\}$ \\
& $\gamma$ & uniform$\{1, 2\}$ \\
& N Steps & Int$\{3, 10\}$ \\
& N Independent & Int$\{1, 5\}$ \\
& N Shared & Int$\{1, 5\}$ \\
& Momentum & Uniform$\{0.01, 4\}$ \\
\midrule
\multirow{5}{*}{\textbf{TabR}} 
& D\_main & Int\{96, 384\} \\
& Context Dropout & uniform$\{0.0, 0.6\}$ \\
& Encoder N Blocks & Int\{0, 1\} \\
& Predictor N Blocks & Int\{1, 2\} \\
& Dropout0 & uniform$\{0.0, 0.6\}$ \\
& N Frequencies & Int\{16, 96\} \\
& Frequency Scale & Loguniform$\{0.01, 100\}$ \\
& D Embedding & Int\{16, 64\} \\
& Learning Rate & Loguniform$\{3e^{-5}, 0.001\}$ \\
& Weight Decay & Loguniform$\{e^{-6}, 0.001\}$ \\
\midrule
\multirow{7}{*}{\textbf{TabTransformer}} 
& Depth & $\{1, 2, 3, 6, 12\}$ \\
& Heads & $\{2, 4, 8\}$ \\
& Dim & $\{32, 64, 128, 256\}$ \\
& Attn\_Dropout  & Uniform $\{0.0, 0.5\}$ \\
& FF\_Dropout  & Uniform $\{0.0, 0.5\}$ \\
& Learning Rate & Loguniform$\{e^{-5}, 0.1\}$ \\
& Weight Decay & Loguniform$\{e^{-6}, 0.001\}$ \\
\midrule
\multirow{7}{*}{\textbf{tabular ResNet}} 
& N Layers & Int$\{1, 8\}$ \\
& D & Int$\{64, 512\}$ \\
& D Hidden Factor  & Uniform $\{1.0, 4.0\}$ \\
& Hidden Dropout  & Uniform $\{0.0, 0.5\}$ \\
& Residual Dropout  & Uniform $\{0.0, 0.5\}$ \\
& Learning Rate & Loguniform$\{e^{-5}, 0.01\}$ \\
& Weight Decay & Loguniform$\{e^{-6}, 0.001\}$ \\
\midrule
\multirow{7}{*}{\textbf{TANGOS}} 
& D\_layers & $\{1, 8, 64, 512\}$ \\
& Dropout & uniform$\{0.0, 0.5\}$ \\
& \(\lambda_1\) & Loguniform$\{0.001, 1\}$ \\
& \(\lambda_2\) & Loguniform$\{0.0001, 1\}$ \\
& Subsample & Int$\{30, 100\}$ \\
& Learning Rate & Loguniform$\{0.0001, 0.001\}$ \\
& Weight Decay & Loguniform$\{e^{-6}, 0.001\}$ \\
%
\end{longtable}

\subsection{Large Language Models}
\label{appendix:C.3}
\paragraph{Llama3-8B}  
Llama3-8B, released by Meta AI in April 2024, is a high-performance Transformer-based model. It utilizes open source datasets and multiple optimization techniques for efficient training. The model integrates reinforcement learning from human feedback to address multi-turn consistency, while dynamic position encoding overcomes the limitations of traditional positional encoding in processing long sequences. The development process ensures data legality and fairness.

\begin{figure*}[htbp]
\vskip 0.2in
\begin{center}
\centerline{\includegraphics[width=0.9\textwidth, trim=0cm 4cm 0cm 0cm, clip]{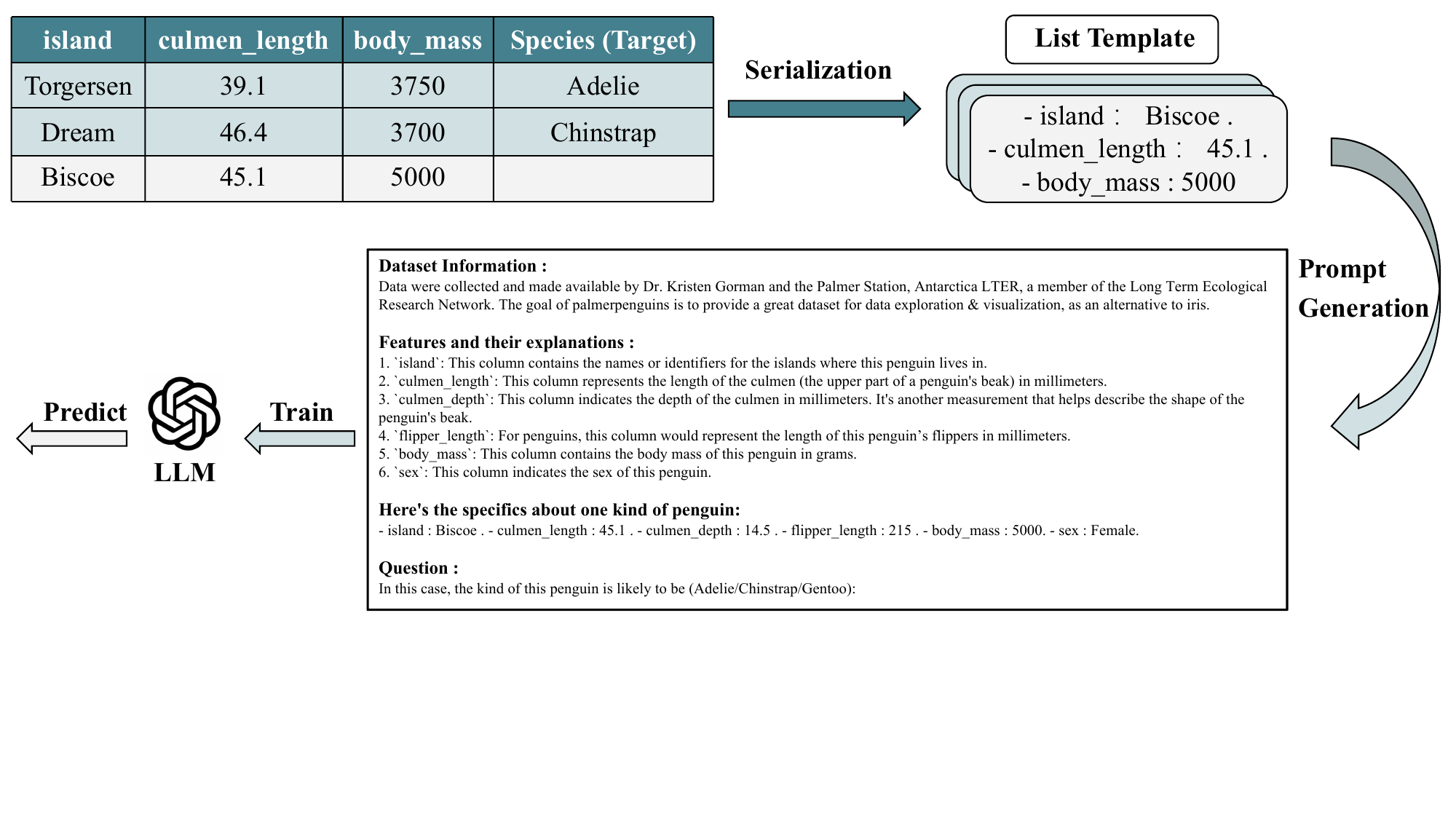}}
\caption{LLM Prompt for the experiments. Take the Penguins dataset as an example.}
\label{prompt}
\end{center}
\vskip -0.2in
\end{figure*}

We convert every row from tabular data into a List Template for input into the LLM. The List Template is a list of column names and feature values, achieving the same outstanding performance as the Text Template according to \cite{hegselmann2023tabllmfewshotclassificationtabular}. We fix an arbitrary ordering of the columns. Figure \ref{prompt} uses the Iris Dataset as an example to show how we use the LLM to handle tabular tasks.

\subsection{Tabular LLMs}
\label{appendix:C.4}
\paragraph{TabLLM}  
TabLLM \cite{hegselmann2023tabllmfewshotclassificationtabular} is a framework that leverages LLMs for efficient tabular data classification. It converts tabular data into natural language strings and uses a few labeled examples for fine-tuning. This approach enables high performance in both zero-shot and few-shot settings, demonstrating its ability to exploit the prior knowledge encoded in LLMs on tabular data.

\paragraph{UniPredict}  
UniPredict \cite{wang2023unipredict} is a framework that leverages LLMs for data-efficient tabular classification. Unlike TabLLM, which uses a generic LLM directly for the task, UniPredict is the first model which is trained on multiple datasets to acquire a rich repository of prior knowledge. This approach allows UniPredict to efficiently handle diverse tabular prediction tasks, achieving strong performance in both few-shot and full-data scenarios while offering scalability across a wide range of datasets.

\section{Feature Imputation Methods}  
\label{appendix:impute}  
We compared the performance of various models using their own imputation methods, random imputation, and mean imputation. Table~\ref{imputation methods} provides a comparison of representative models employing these three missing-feature imputation methods. We present the average performance of several representative models under different imputation methods. Table~\ref{imputation result} shows that the model performs best under mean imputation, and its performance declines under all three types of imputation methods, indicating that the model still faces challenges from feature shift.

\section{Feature Importance Analysis}  
\label{appendix:D}  
\subsection{Importance Metrics}
\begin{figure*}
\vskip -0.2in
\label{E.3}
\begin{center}
\centerline{\includegraphics[width=\textwidth]{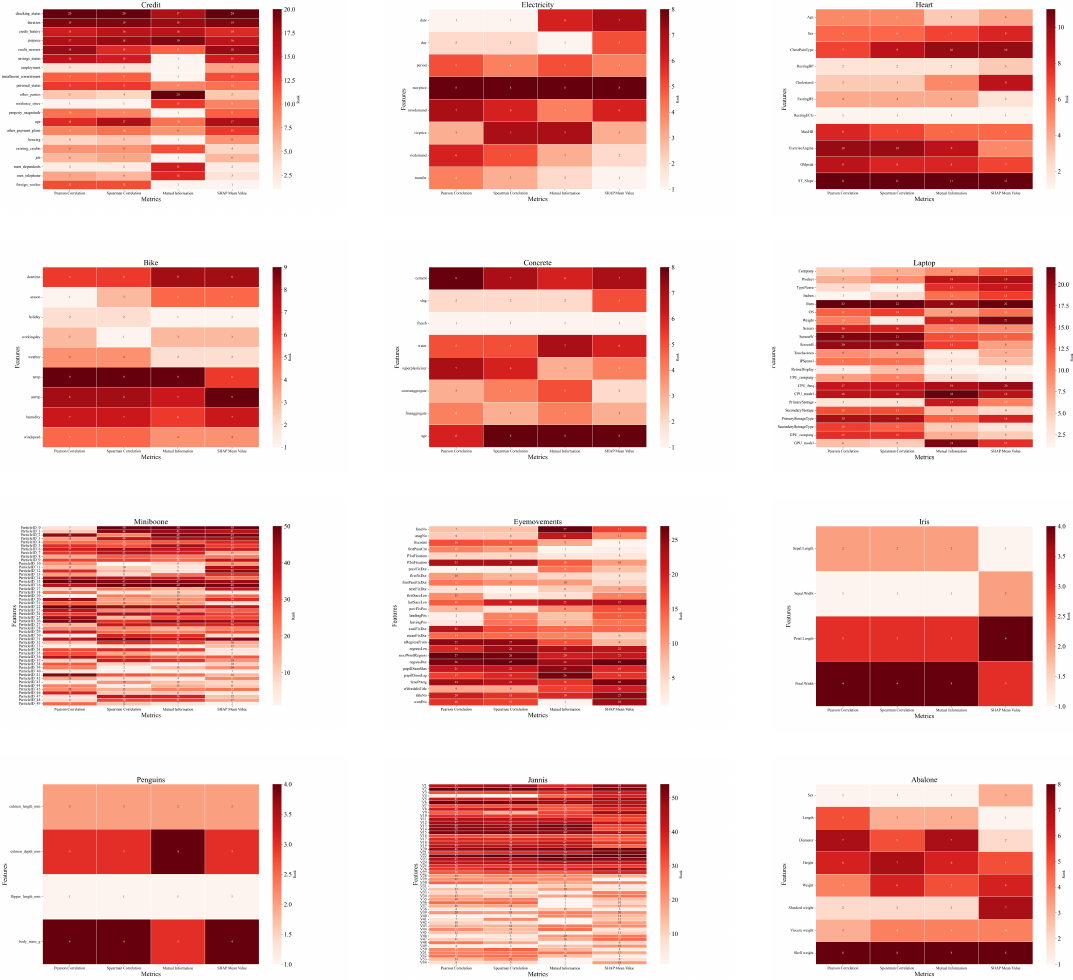}}
\caption{Visual comparison of feature importance rankings across Pearson correlation, Spearman correlation, mutual information, and SHAP values on twelve datasets.}
\label{featureimportance}
\end{center}
\end{figure*}

To evaluate the consistency of feature importance rankings, we compute Kendall's $\tau$ correlation coefficients among four metrics: Pearson correlation coefficient (PCC), Spearman's rank correlation, SHAP values, and mutual information. Figure~\ref{featureimportance} reveal high concordance across these measures, with particularly strong agreement between PCC and Spearman ($\tau$ = 0.61). While both demonstrate comparable performance, we ultimately select PCC for its widespread adoption and intuitive interpretability in the research community. The marginal differences between these metrics' rankings were found to be statistically insignificant and did not affect our analytical conclusions.

Notably, the feature importance derived from PCC exhibits a strong negative correlation with model performance degradation upon feature removal ($\rho$ = -0.78, $p$ < 0.01). This inverse relationship indicates that features assigned higher importance scores by PCC corresponded to greater performance declines when omitted, suggesting consistent feature dependencies across both closed-environment training and feature shift scenarios. This observed stability likely stems from our model's robust feature selection and dynamic weighting mechanisms.

\begin{table*}[t]
\begin{center}
\begin{small}

\caption{Representative self-imputation methods. }
\label{imputation methods}
\vskip 0.1in
\begin{tabular}{ccc}
\toprule
Method   Type                        & Method                                           & Representative   Models   \\
\midrule
\multirow{4}{*}{Has Internal Module} & Treat missing-feature   values as feature minima & CatBoost                  \\
                                     & Left subtree split                               & XGBoost, LightGBM         \\
                                     & Missing-feature values as a separate category    & TabTransformer            \\
                                     & No dimension consistency requirement             & Llam3-8B                  \\
                                     \midrule
No Internal   Module                 & Imput missing   features by 0                    & Most deep-learning models \\
\bottomrule
\end{tabular}

\end{small}
\end{center}
\vskip -0.1in
\end{table*}

\begin{table*}[t]
\begin{center}
\begin{small}

\caption{Methods performance comparison: model imputation methods vs. mean-value imputation vs. random-value imputation.}
\label{imputation result}
\resizebox{\textwidth}{18mm}{
\begin{tabular}{ccccccccccccccccccc}
\toprule
\multirow{2}{*}{Shift}  & \multicolumn{3}{c}{CatBoost} & \multicolumn{3}{c}{XGBoost} & \multicolumn{3}{c}{MLP} & \multicolumn{3}{c}{TabPFN} & \multicolumn{3}{c}{TabTransformer} & \multicolumn{3}{c}{Llam3-8B} \\
\cmidrule(lr){2-4}\cmidrule(lr){5-7}\cmidrule(lr){8-10}\cmidrule(lr){11-13}\cmidrule(lr){14-16}\cmidrule(lr){17-19}
& NAN     & Mean    & Random   & NAN     & Mean    & Random  & NAN    & Mean  & Random & NAN     & Mean   & Random  & NAN       & Mean      & Random     & NAN     & Mean    & Random   \\
\midrule
10\%   & 0.845   & 0.851   & 0.818    & 0.823   & 0.825   & 0.800   & 0.747  & 0.843 & 0.724  & 0.858   & 0.859  & 0.829   & 0.497     & 0.513     & 0.487      & 0.765   & 0.772   & 0.517    \\
20\%   & 0.820   & 0.826   & 0.807    & 0.784   & 0.798   & 0.796   & 0.704  & 0.833 & 0.665  & 0.828   & 0.848  & 0.797   & 0.480     & 0.516     & 0.470      & 0.750   & 0.752   & 0.526    \\
30\%   & 0.792   & 0.801   & 0.767    & 0.746   & 0.773   & 0.756   & 0.686  & 0.821 & 0.624  & 0.798   & 0.835  & 0.762   & 0.486     & 0.502     & 0.476      & 0.736   & 0.721   & 0.520    \\
40\%   & 0.763   & 0.775   & 0.739    & 0.709   & 0.747   & 0.720   & 0.656  & 0.821 & 0.598  & 0.769   & 0.820  & 0.724   & 0.480     & 0.511     & 0.470      & 0.724   & 0.708   & 0.532    \\
50\%   & 0.733   & 0.747   & 0.702    & 0.673   & 0.720   & 0.681   & 0.639  & 0.803 & 0.585  & 0.741   & 0.803  & 0.698   & 0.469     & 0.513     & 0.459      & 0.720   & 0.677   & 0.502    \\
60\%   & 0.702   & 0.716   & 0.665    & 0.639   & 0.692   & 0.655   & 0.616  & 0.792 & 0.580  & 0.694   & 0.784  & 0.656   & 0.482     & 0.510     & 0.472      & 0.704   & 0.683   & 0.546    \\
70\%   & 0.670   & 0.682   & 0.630    & 0.606   & 0.662   & 0.625   & 0.584  & 0.779 & 0.580  & 0.657   & 0.762  & 0.632   & 0.477     & 0.492     & 0.467      & 0.685   & 0.672   & 0.539    \\
80\%   & 0.637   & 0.643   & 0.608    & 0.577   & 0.628   & 0.596   & 0.572  & 0.723 & 0.582  & 0.640   & 0.734  & 0.600   & 0.467     & 0.483     & 0.457      & 0.659   & 0.655   & 0.555    \\
90\%   & 0.600   & 0.596   & 0.574    & 0.559   & 0.589   & 0.571   & 0.546  & 0.718 & 0.578  & 0.593   & 0.695  & 0.575   & 0.467     & 0.494     & 0.455      & 0.625   & 0.651   & 0.529    \\
100\%  & 0.538   & 0.541   & 0.502    & 0.561   & 0.545   & 0.548   & 0.559  & 0.621 & 0.551  & 0.557   & 0.638  & 0.556   & 0.467     & 0.482     & 0.454      & 0.575   & 0.622   & 0.517 \\
\bottomrule
\end{tabular}}

\end{small}
\end{center}
\vskip -0.1in
\end{table*}

\subsection{Pearson Correlation}
To effectively evaluate the impact of feature shifts on model performance, we employ Pearson correlation to rank the features within a given dataset.  
The Pearson correlation explicitly provides the correlation coefficient between an input feature and the target variable. By applying Pearson correlation analysis to datasets with feature shifts, we can observe how the absence of features with varying degrees of correlation affects model performance. This analysis facilitates the identification of patterns and provides insights into feature shifts from multiple perspectives. Pearson correlation serves as an enhancement over both Euclidean distance and cosine similarity when handling datasets with missing dimensions.  
The Pearson correlation coefficient quantifies the linear relationship between two variables by computing the ratio of their covariance to the product of their standard deviations. It is defined within the range $[-1, 1]$ and is calculated as follows:  
\begin{equation}  
\rho _{X,Y} = \frac{cov(X,Y)}{\sigma _X \sigma _Y} = \frac{E[(X - \mu _X)(Y - \mu _Y)]}{\sigma _X \sigma _Y}  
\end{equation}  
Here, $X$ and $Y$ represent two variables, \(cov(X,Y)\) denotes their covariance, and $\sigma_X$ and $\sigma_Y$ are their respective standard deviations.   If the Pearson correlation coefficient ($PCC$) $\rho _{X, Y}$ equals 0, it indicates no linear relationship between $X$ and $Y$. A positive correlation ($0 < \rho _{X,Y} \leq 1$) implies that as $X$ increases, $Y$ also increases. Conversely, a negative correlation ($-1 \leq \rho _{X,Y} < 0$) signifies that as $X$ increases, $Y$ decreases. The closer $|\rho _{X, Y}|$ is to 1, the stronger the linear relationship between the two variables. 

\subsection{Automated Feature Engineering Analysis}
We posit that Automated Feature Engineering (AutoFE) can address feature shift scenarios through two approaches.

\begin{itemize}
    \item \textbf{Imputation:} CAAFE~\cite{hollmann2024large} and OcTree~\cite{nam2024optimized} generate rules for imputing specific features, while OpenFE's generated features do not match the originals~\cite{zhang2023openfe}. We test single shifts using CAAFE and OctTree for imputation. Table~\ref{autofe1} shows that LLM-based AutoFE can effectively generate matching features, enhancing model robustness compared to mean imputation.

    \item \textbf{Generation:} We use AutoFE to generate new features to offset the impact of missing original features on model performance. Table~\ref{autofe2} shows that LLM-based AutoFE has significant potential in feature shift scenarios.

\end{itemize}

As the importance of features increases, both model performance declines, further corroborating Observation 2 of TabFSBench.

\begin{table*}[t]
\caption{Results from AutoFE as an imputation strategy. We evaluate CAAFE and OcTree on the downstream model of CatBoost and TabPFN under the Single Shift experiment.}
\vskip 0.2in
\label{autofe1}
    \centering
    \begin{tabular}{ccccccc}
    \toprule
\multirow{2}{*}{Shift} & \multicolumn{3}{c}{CatBoost} & \multicolumn{3}{c}{TabPFN} \\
\cmidrule(lr){2-4}\cmidrule(lr){5-7}
                       & Baseline  & CAAFE  & OcTree  & Baseline     & CAAFE    & OcTree    \\
                    \midrule
1                      & 0.870     & 0.869  & 0.870   & 0.866        & 0.875    & 0.873     \\
2                      & 0.871     & 0.880  & 0.872   & 0.862        & 0.877    & 0.872     \\
3                      & 0.864     & 0.874  & 0.872   & 0.859        & 0.876    & 0.877     \\
4                      & 0.861     & 0.865  & 0.864   & 0.861        & 0.868    & 0.874     \\
5                      & 0.875     & 0.882  & 0.885   & 0.866        & 0.861    & 0.868     \\
6                      & 0.859     & 0.859  & 0.863   & 0.857        & 0.883    & 0.875     \\
7                      & 0.844     & 0.854  & 0.853   & 0.853        & 0.865    & 0.868     \\
8                      & 0.866     & 0.871  & 0.868   & 0.873        & 0.845    & 0.852     \\
9                      & 0.871     & 0.875  & 0.871   & 0.868        & 0.871    & 0.866     \\
10                     & 0.864     & 0.874  & 0.864   & 0.871        & 0.875    & 0.874     \\
11                     & 0.663     & 0.664  & 0.666   & 0.817        & 0.868    & 0.872     \\
100\%                  & 0.464     & 0.832  & 0.830   & 0.518        & 0.865    & 0.857     \\
\bottomrule
\end{tabular}
\end{table*}

\begin{table*}[t]
\caption{Results from AutoFE as feature generator. We evaluate OpenFE, CAAFE and OcTree on the downstream model of CatBoost and TabPFN under the Random Shift experiment.}
\vskip 0.2in
\label{autofe2}
    \centering
    \begin{tabular}{ccccccccc}
    \toprule
\multirow{2}{*}{Shift} & \multicolumn{4}{c}{CatBoost}       & \multicolumn{4}{c}{TabPFN Baseline} \\
                       \cmidrule(lr){2-5}\cmidrule(lr){6-9}& Baseline & OpenFE & CAAFE & OcTree & Baseline  & OpenFE & CAAFE & OcTree \\
                       \midrule
0                      & 0.879    & 0.886  & 0.887 & 0.888  & 0.862     & 0.853  & 0.863 & 0.860  \\
9\%                    & 0.851    & 0.859  & 0.861 & 0.859  & 0.859     & 0.871  & 0.874 & 0.876  \\
18\%                   & 0.826    & 0.857  & 0.859 & 0.857  & 0.848     & 0.866  & 0.876 & 0.872  \\
27\%                   & 0.801    & 0.855  & 0.862 & 0.855  & 0.835     & 0.866  & 0.872 & 0.870  \\
36\%                   & 0.775    & 0.853  & 0.860 & 0.853  & 0.820     & 0.864  & 0.874 & 0.870  \\
45\%                   & 0.747    & 0.841  & 0.849 & 0.841  & 0.803     & 0.868  & 0.878 & 0.862  \\
54\%                   & 0.716    & 0.842  & 0.845 & 0.842  & 0.784     & 0.851  & 0.861 & 0.861  \\
63\%                   & 0.682    & 0.835  & 0.842 & 0.835  & 0.762     & 0.853  & 0.860 & 0.869  \\
72\%                   & 0.643    & 0.842  & 0.844 & 0.842  & 0.734     & 0.846  & 0.857 & 0.857  \\
81\%                   & 0.596    & 0.850  & 0.853 & 0.850  & 0.695     & 0.857  & 0.860 & 0.867  \\
90\%                   & 0.541    & 0.841  & 0.850 & 0.841  & 0.638     & 0.856  & 0.860 & 0.863  \\
100\%                  & 0.464    & 0.830  & 0.832 & 0.830  & 0.518     & 0.853  & 0.865 & 0.857 \\
\bottomrule
\end{tabular}
\end{table*}

\section{Experiment Details of Observations}
\subsection{Training Details}
The deep learning models, LLMs, and Tabular LLMs were trained on an NVIDIA A800 GPU. Gradient-boosted tree models, where applicable, were trained on a CPU rather than a GPU, using an AMD Ryzen 5 7500F 6-Core Processor. All experimental results are reported as the average of three different random seeds to ensure robustness.

\subsection{Different Types of Feature Shift}
We address the analysis that compare model performance on different types of feature shifts by clarifying two distinct analytical perspectives on feature shifts. First, regarding different types of feature shifts, we direct attention to our comprehensive robustness evaluation across multiple shift categories presented in the leaderboard analysis. Second, examining shifts by feature type, Table~\ref{feature type} reveals differential sensitivity patterns: model performance demonstrates greatest vulnerability to categorical feature perturbations, followed by boolean features, with numerical features exhibiting the most stable behavior. This hierarchy of susceptibility persists across all tested datasets, suggesting inherent algorithmic dependencies on feature types.

\begin{table*}[t]
\begin{center}
\caption{Results on different types of features shifted. We divide feature types into Categorical, Numerical and Boolean.}
\vskip 0.1in
\label{feature type}
\begin{tabular}{ccccc}
\toprule
Model          & Baseline & Categorical & Numerical & Boolean \\
\midrule
LightGBM       & 0.833    & 0.786       & 0.83      & 0.821   \\
XGBoost        & 0.855    & 0.804       & 0.84      & 0.81    \\
CatBoost       & 0.879    & 0.792       & 0.868     & 0.861   \\
TabPFN         & 0.862    & 0.845       & 0.865     & 0.864   \\
DANets         & 0.674    & 0.648       & 0.667     & 0.661   \\
MLP            & 0.848    & 0.837       & 0.848     & 0.838   \\
NODE           & 0.663    & 0.651       & 0.641     & 0.637   \\
ResNet         & 0.857    & 0.845       & 0.86      & 0.851   \\
SwitchTab      & 0.871    & 0.858       & 0.867     & 0.859   \\
TabCaps        & 0.775    & 0.76        & 0.772     & 0.763   \\
TabNet         & 0.681    & 0.659       & 0.678     & 0.654   \\
TANGOS         & 0.859    & 0.853       & 0.863     & 0.86    \\
AutoInt        & 0.683    & 0.678       & 0.673     & 0.696   \\
DCNv2          & 0.868    & 0.854       & 0.861     & 0.841   \\
FT-Transformer & 0.774    & 0.734       & 0.773     & 0.768   \\
GrowNet        & 0.589    & 0.586       & 0.598     & 0.591   \\
Saint          & 0.875    & 0.861       & 0.866     & 0.861   \\
SNN            & 0.788    & 0.775       & 0.78      & 0.776   \\
TabTransformer & 0.522    & 0.504       & 0.523     & 0.508   \\
TabR           & 0.893    & 0.877       & 0.893     & 0.896   \\
ModernNCA      & 0.88     & 0.849       & 0.881     & 0.889   \\
Llama3-8B      & 0.848    & 0.76        & 0.805     & 0.765   \\
TabLLM         & 0.783    & 0.786       & 0.795     & 0.766   \\
UniPredict     & 0.853    & 0.861       & 0.849     & 0.84   \\
\bottomrule
\end{tabular}
\end{center}
\end{table*}

\subsection{Removing Most/Least Relevant Features}
\label{appendix:E.2}
\begin{figure*}[t]
\label{E.2}
\begin{center}
\centerline{\includegraphics[width=\textwidth]{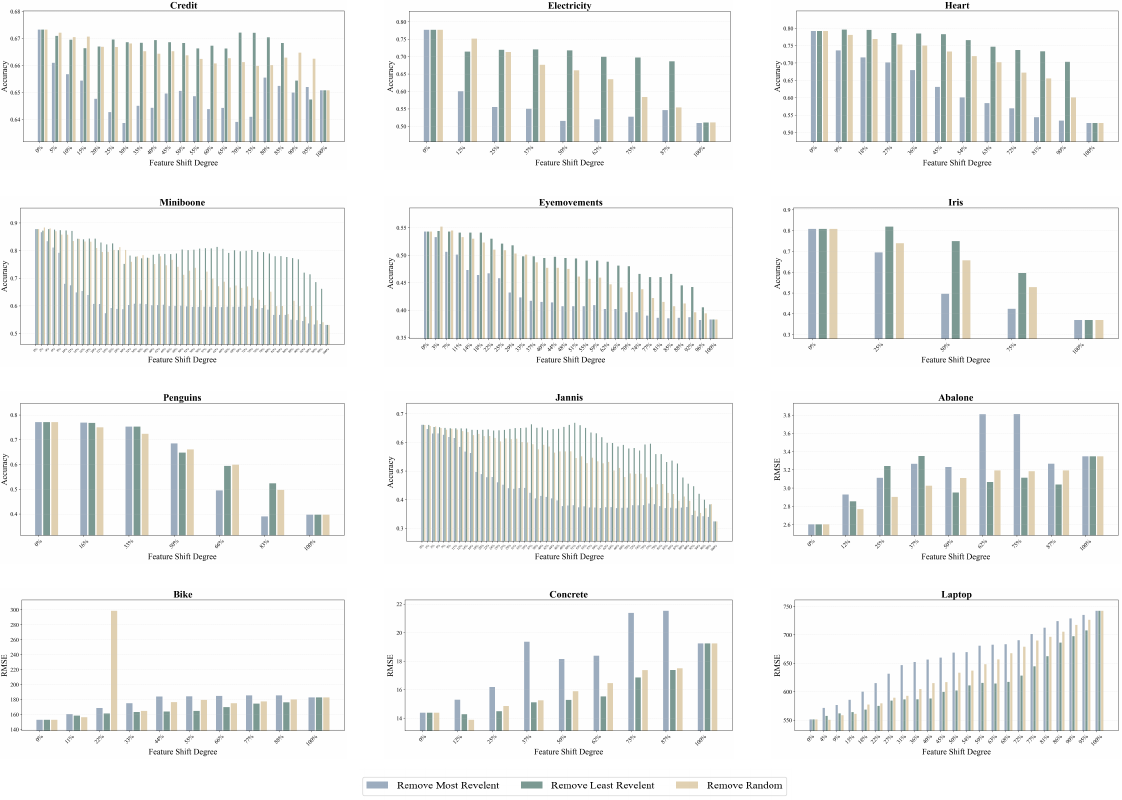}}
\caption{Results for most/least shift experiments with correlation analysis across datasets. The $x$-axis represents the percentage of features removed, while the $y$-axis indicates the average performance (accuracy or RMSE) of all models. Three feature removal strategies are compared: \textbf{Remove Most Relevant} (blue), which removes $k$ most correlated columns; \textbf{Remove Least Relevant} (green), which removes $k$ least correlated columns; and \textbf{Remove Random} (yellow), which removes $k$ randomly selected features. }
\label{experiment 2}
\end{center}
\vskip -0.2in
\end{figure*}

As shown in the experimental results in Figure~\ref{E.2}, under the least relevant feature removal setting, the model's performance remains largely unchanged when removing $0 \sim t$\% of the least relevant columns. Interestingly, we observe that removing a certain number of the most irrelevant columns improves model performance. This observation aligns with the insight from \cite{grinsztajn2022tree} that \textbf{uninformative features can negatively impact model performance}.

\subsection{Detailed Explanations of Figure~\ref{figure3}}
The reason for the presence of multiple trajectories in Figure~\ref{figure3} is that each trajectory comprises points derived from a single dataset. During the fitting process, the results from all datasets were aggregated, leading to the emergence of distinct trajectories corresponding to individual datasets. Notably, these trajectories capture the relationship between feature importance and model performance across different datasets. Specifically, each dataset's trajectory illustrates how its particular pattern of feature importance influences the resulting model performance.

\begin{figure*}
\label{traject1}
\begin{center}
\centerline{\includegraphics[width=\textwidth]{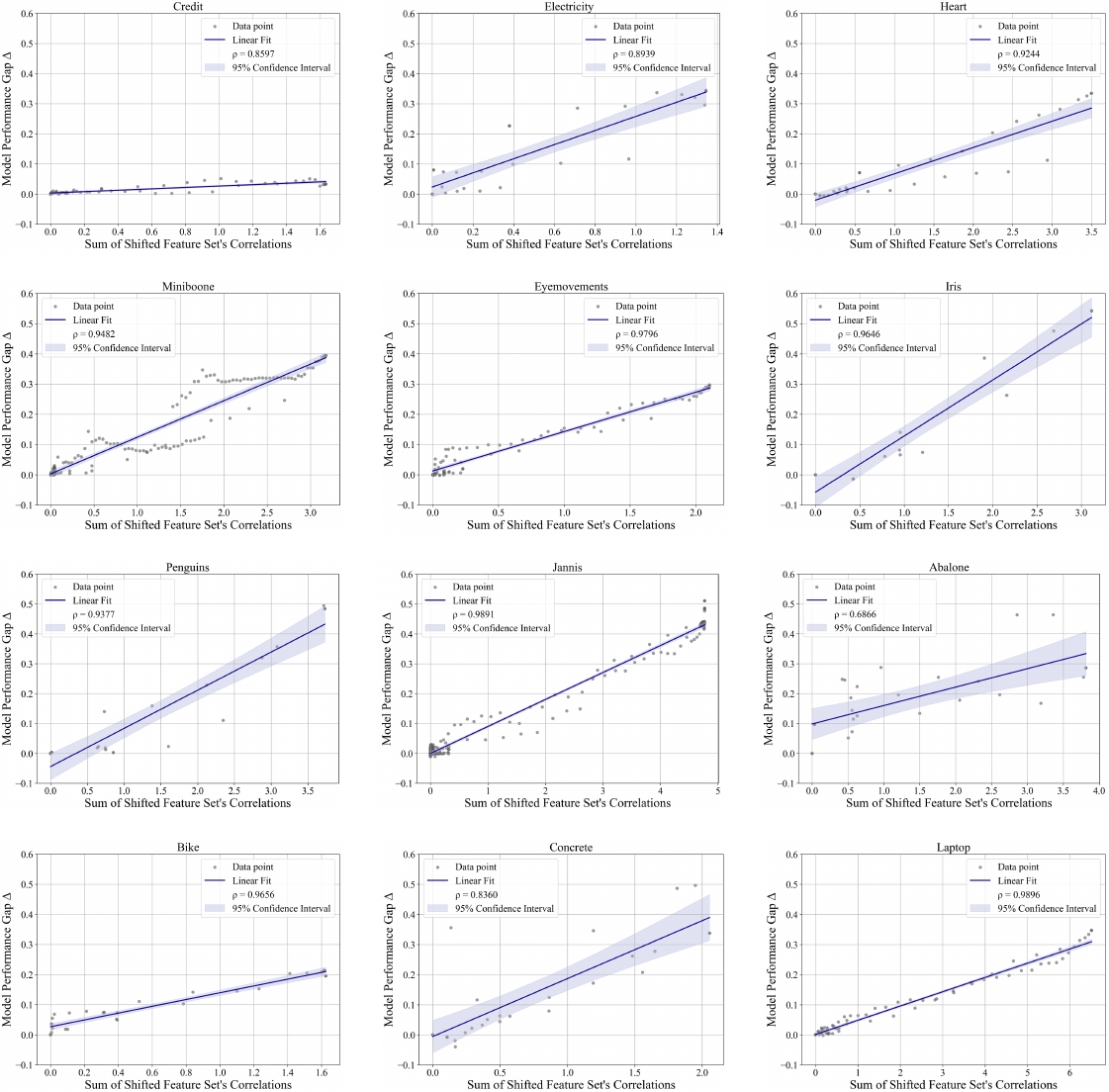}}
\caption{Trajectories between correlation and accuracy for each of twelve datasets.}
\end{center}
\end{figure*}

To better support our conclusions, Figure~\ref{traject1} provides correlation and accuracy plots for twelve different datasets. It indicates that although levels of feature importance vary across different datasets, they all support our research conclusion, namely, that there is a significant linear correlation between feature importance and model performance. These trajectories further confirm the generality and reliability of Observation 2.

\subsection{Results on High-Dimensional Datasets}

Following OcTree~\cite{nam2024optimized}, we conducted random shift experiments on two high-dimensional datasets, \textit{madelon} and \textit{nomao}, and present model performance under 10\%, 20\%, ..., and 100\% feature shift degrees. Table~\ref{high1} and~\ref{high2} demonstrate that as the degree of feature shift increases, the model performance decreases significantly.

\paragraph{Madelon}
The \textit{madelon} dataset was originally designed for the NIPS 2003 feature selection challenge. It is an artificial binary classification task in which instances are generated from clusters positioned at the vertices of a five-dimensional hypercube. The dataset contains 500 features, including 20 informative (or redundant) features and 480 noise features, making it a prototypical benchmark for evaluating robustness and feature selection under high-dimensional settings. The dataset is available from the UCI Machine Learning Repository: \href{https://archive.ics.uci.edu/dataset/171/madelon}{https://archive.ics.uci.edu/dataset/171/madelon}.

\paragraph{Nomao}
The \textit{nomao} dataset originates from a real-world deduplication task involving geolocated business entries. Each sample represents a comparison between two records, described by both numerical and categorical similarity features such as name, address, and geocoordinates. The dataset includes 34,465 samples and 118 features (89 numeric and 29 categorical), with a notable proportion of missing values. Its mixed-type, noisy, and partially incomplete nature makes it a valuable resource for studying model robustness in practical high-dimensional scenarios. The dataset is publicly accessible at \href{https://archive.ics.uci.edu/dataset/227/nomao}{https://archive.ics.uci.edu/dataset/227/nomao}.

\begin{table*}[t]
\begin{center}
\begin{small}
\caption{Results on nomao dataset. We don’t evaluate TabPFN(due to the limitation of TabPFN on high-dimension datasets which have more than 100 features), Llama3-8B and Unipredict (due to the limitations of input).}
\vskip 0.1in
\label{high1}
\begin{tabular}{cccccccccccc}
\toprule
ACC            & ID               & 10\%             & 20\%             & 30\%             & 40\%             & 50\%             & 60\%             & 70\%             & 80\%             & 90\%             & 100\%            \\
\midrule
LightGBM       & 0.970            & 0.969            & 0.968            & 0.969            & 0.967            & 0.964            & 0.958            & 0.961            & 0.954            & 0.926            & 0.890            \\
XGBoost        & 0.974            & 0.974            & 0.973            & 0.973            & 0.972            & 0.967            & 0.971            & 0.964            & 0.933            & 0.939            & 0.874            \\
CatBoost       & 0.970            & 0.969            & 0.969            & 0.969            & 0.969            & 0.960            & 0.960            & 0.960            & 0.955            & 0.945            & 0.892            \\
TabPFN         & \textbackslash{} & \textbackslash{} & \textbackslash{} & \textbackslash{} & \textbackslash{} & \textbackslash{} & \textbackslash{} & \textbackslash{} & \textbackslash{} & \textbackslash{} & \textbackslash{} \\
DANets         & 0.933            & 0.930            & 0.921            & 0.908            & 0.888            & 0.872            & 0.845            & 0.816            & 0.808            & 0.719            & 0.713            \\
MLP            & 0.949            & 0.946            & 0.938            & 0.931            & 0.914            & 0.894            & 0.884            & 0.847            & 0.874            & 0.762            & 0.756            \\
NODE           & 0.778            & 0.768            & 0.760            & 0.750            & 0.723            & 0.716            & 0.713            & 0.713            & 0.713            & 0.713            & 0.713            \\
ResNet         & 0.953            & 0.946            & 0.936            & 0.929            & 0.882            & 0.852            & 0.815            & 0.777            & 0.824            & 0.719            & 0.714            \\
SwitchTab      & 0.936            & 0.941            & 0.935            & 0.931            & 0.911            & 0.886            & 0.874            & 0.847            & 0.867            & 0.767            & 0.753            \\
TabCaps        & 0.920            & 0.916            & 0.905            & 0.897            & 0.899            & 0.882            & 0.876            & 0.852            & 0.864            & 0.832            & 0.804            \\
TabNet         & 0.921            & 0.899            & 0.907            & 0.900            & 0.822            & 0.878            & 0.875            & 0.806            & 0.876            & 0.715            & 0.756            \\
TANGOS         & 0.955            & 0.950            & 0.937            & 0.930            & 0.904            & 0.874            & 0.869            & 0.816            & 0.881            & 0.760            & 0.736            \\
AutoInt        & 0.939            & 0.934            & 0.912            & 0.901            & 0.862            & 0.823            & 0.759            & 0.757            & 0.771            & 0.718            & 0.714            \\
DCNv2          & 0.949            & 0.946            & 0.938            & 0.930            & 0.916            & 0.903            & 0.899            & 0.859            & 0.881            & 0.772            & 0.806            \\
FT-Transformer & 0.934            & 0.933            & 0.925            & 0.909            & 0.860            & 0.859            & 0.801            & 0.772            & 0.790            & 0.714            & 0.713            \\
GrowNet        & 0.832            & 0.818            & 0.812            & 0.797            & 0.779            & 0.785            & 0.759            & 0.741            & 0.759            & 0.715            & 0.729            \\
Saint          & 0.952            & 0.943            & 0.928            & 0.915            & 0.852            & 0.842            & 0.801            & 0.762            & 0.792            & 0.718            & 0.713            \\
SNN            & 0.910            & 0.909            & 0.904            & 0.890            & 0.889            & 0.880            & 0.885            & 0.859            & 0.857            & 0.837            & 0.798            \\
TabTransformer & 0.928            & 0.925            & 0.915            & 0.905            & 0.884            & 0.879            & 0.871            & 0.852            & 0.877            & 0.802            & 0.779            \\
TabR           & 0.915            & 0.913            & 0.904            & 0.894            & 0.894            & 0.881            & 0.881            & 0.855            & 0.860            & 0.834            & 0.801            \\
ModernNCA      & 0.964            & 0.952            & 0.945            & 0.928            & 0.842            & 0.861            & 0.842            & 0.768            & 0.812            & 0.719            & 0.713            \\
Llama3-8B      & \textbackslash{} & \textbackslash{} & \textbackslash{} & \textbackslash{} & \textbackslash{} & \textbackslash{} & \textbackslash{} & \textbackslash{} & \textbackslash{} & \textbackslash{} & \textbackslash{} \\
TabLLM         & 0.959            & 0.957            & 0.938            & 0.893            & 0.856            & 0.842            & 0.616            & 0.676            & 0.506            & 0.432            & 0.329            \\
UniPredict     & \textbackslash{} & \textbackslash{} & \textbackslash{} & \textbackslash{} & \textbackslash{} & \textbackslash{} & \textbackslash{} & \textbackslash{} & \textbackslash{} & \textbackslash{} & \textbackslash{}\\
\bottomrule
\end{tabular}
\end{small}
\end{center}
\vskip -0.1in
\end{table*}

\begin{table*}[t]
\begin{center}
\begin{small}
\caption{Results on madelon dataset. We don’t evaluate TabPFN(due to the limitation of TabPFN on high-dimension datasets which have more than 100 features), Llama3-8B, TabLLM and Unipredict (due to the limitations of input).}
\vskip 0.1in
\label{high2}
\begin{tabular}{cccccccccccc}
\toprule
ACC            & ID               & 10\%             & 20\%             & 30\%             & 40\%             & 50\%             & 60\%             & 70\%             & 80\%             & 90\%             & 100\%            \\
\midrule
LightGBM                       & 0.819                                & 0.813                                & 0.815                                & 0.819                                & 0.825                                & 0.796                                & 0.737                                & 0.723                                & 0.635                                & 0.574                                & 0.562                                \\
XGBoost                        & 0.828                                & 0.818                                & 0.827                                & 0.804                                & 0.829                                & 0.769                                & 0.772                                & 0.693                                & 0.614                                & 0.560                                & 0.554                                \\
CatBoost                       & 0.865                                & 0.859                                & 0.797                                & 0.862                                & 0.831                                & 0.836                                & 0.770                                & 0.786                                & 0.630                                & 0.547                                & 0.841                                \\
DANets                         & 0.523                                & 0.523                                & 0.524                                & 0.514                                & 0.513                                & 0.510                                & 0.509                                & 0.502                                & 0.503                                & 0.496                                & 0.495                                \\
MLP                            & 0.518                                & 0.521                                & 0.521                                & 0.515                                & 0.528                                & 0.519                                & 0.511                                & 0.506                                & 0.507                                & 0.503                                & 0.498                                \\
NODE                           & 0.507                                & 0.512                                & 0.515                                & 0.500                                & 0.526                                & 0.510                                & 0.505                                & 0.494                                & 0.501                                & 0.499                                & 0.495                                \\
ResNet                         & 0.555                                & 0.553                                & 0.553                                & 0.535                                & 0.537                                & 0.547                                & 0.536                                & 0.516                                & 0.529                                & 0.507                                & 0.501                                \\
SwitchTab                      & 0.483                                & 0.483                                & 0.483                                & 0.483                                & 0.483                                & 0.483                                & 0.483                                & 0.483                                & 0.483                                & 0.483                                & 0.479                                \\
TabCaps                        & 0.516                                & 0.509                                & 0.515                                & 0.508                                & 0.516                                & 0.514                                & 0.490                                & 0.504                                & 0.503                                & 0.502                                & 0.498                                \\
TabNet                         & 0.512                                & 0.504                                & 0.504                                & 0.507                                & 0.505                                & 0.502                                & 0.496                                & 0.500                                & 0.497                                & 0.496                                & 0.490                                \\
TANGOS                         & 0.571                                & 0.560                                & 0.562                                & 0.551                                & 0.566                                & 0.545                                & 0.550                                & 0.523                                & 0.525                                & 0.511                                & 0.504                                \\
AutoInt                        & 0.592                                & 0.558                                & 0.543                                & 0.543                                & 0.532                                & 0.530                                & 0.526                                & 0.519                                & 0.511                                & 0.503                                & 0.501                                \\
DCNv2                          & 0.528                                & 0.522                                & 0.523                                & 0.509                                & 0.519                                & 0.512                                & 0.528                                & 0.512                                & 0.508                                & 0.513                                & 0.507                                \\
FT-Transformer                 & 0.640                                & 0.579                                & 0.553                                & 0.554                                & 0.546                                & 0.541                                & 0.540                                & 0.536                                & 0.522                                & 0.521                                & 0.517                                \\
GrowNet                        & 0.506                                & 0.503                                & 0.494                                & 0.496                                & 0.499                                & 0.508                                & 0.509                                & 0.502                                & 0.517                                & 0.509                                & 0.506                                \\
Saint                          & 0.544                                & 0.543                                & 0.541                                & 0.539                                & 0.536                                & 0.528                                & 0.524                                & 0.523                                & 0.506                                & 0.503                                & 0.501                                \\
SNN                            & 0.508                                & 0.507                                & 0.503                                & 0.498                                & 0.508                                & 0.506                                & 0.513                                & 0.501                                & 0.493                                & 0.505                                & 0.502                                \\
TabTransformer                 & 0.543                                & 0.544                                & 0.534                                & 0.532                                & 0.548                                & 0.540                                & 0.534                                & 0.513                                & 0.523                                & 0.520                                & 0.514                                \\
TabR                           & 0.640                                & 0.620                                & 0.556                                & 0.538                                & 0.534                                & 0.525                                & 0.520                                & 0.511                                & 0.505                                & 0.506                                & 0.503                                \\
ModernNCA                      & 0.593                                & 0.587                                & 0.573                                & 0.549                                & 0.588                                & 0.553                                & 0.539                                & 0.519                                & 0.516                                & 0.505                                & 0.502                                \\
\multicolumn{1}{c}{Llama3-8B}  & \multicolumn{1}{c}{\textbackslash{}} & \multicolumn{1}{c}{\textbackslash{}} & \multicolumn{1}{c}{\textbackslash{}} & \multicolumn{1}{c}{\textbackslash{}} & \multicolumn{1}{c}{\textbackslash{}} & \multicolumn{1}{c}{\textbackslash{}} & \multicolumn{1}{c}{\textbackslash{}} & \multicolumn{1}{c}{\textbackslash{}} & \multicolumn{1}{c}{\textbackslash{}} & \multicolumn{1}{c}{\textbackslash{}} & \multicolumn{1}{c}{\textbackslash{}} \\
\multicolumn{1}{c}{TabLLM}     & \multicolumn{1}{c}{\textbackslash{}} & \multicolumn{1}{c}{\textbackslash{}} & \multicolumn{1}{c}{\textbackslash{}} & \multicolumn{1}{c}{\textbackslash{}} & \multicolumn{1}{c}{\textbackslash{}} & \multicolumn{1}{c}{\textbackslash{}} & \multicolumn{1}{c}{\textbackslash{}} & \multicolumn{1}{c}{\textbackslash{}} & \multicolumn{1}{c}{\textbackslash{}} & \multicolumn{1}{c}{\textbackslash{}} & \multicolumn{1}{c}{\textbackslash{}} \\
\multicolumn{1}{c}{UniPredict} & \multicolumn{1}{c}{\textbackslash{}} & \multicolumn{1}{c}{\textbackslash{}} & \multicolumn{1}{c}{\textbackslash{}} & \multicolumn{1}{c}{\textbackslash{}} & \multicolumn{1}{c}{\textbackslash{}} & \multicolumn{1}{c}{\textbackslash{}} & \multicolumn{1}{c}{\textbackslash{}} & \multicolumn{1}{c}{\textbackslash{}} & \multicolumn{1}{c}{\textbackslash{}} & \multicolumn{1}{c}{\textbackslash{}} & \multicolumn{1}{c}{\textbackslash{}}\\
\bottomrule
\end{tabular}
\end{small}
\end{center}
\vskip -0.1in
\end{table*}

\subsection{Feature-Shift Performance vs. Runtime}
In Figure~\ref{E.3}, we plot feature-shift performance versus runtime for all models, averaged across datasets. LLMs and Tabular LLMs exhibit the longest runtime, whereas tree-based models achieve excellent performance with minimal runtime requirements. However, we observe that models such as NODE and LLMs consume significantly more memory compared to tree-based approaches.

\begin{figure*}
\vskip -0.2in
\label{E.3}
\begin{center}
\centerline{\includegraphics[width=\textwidth]{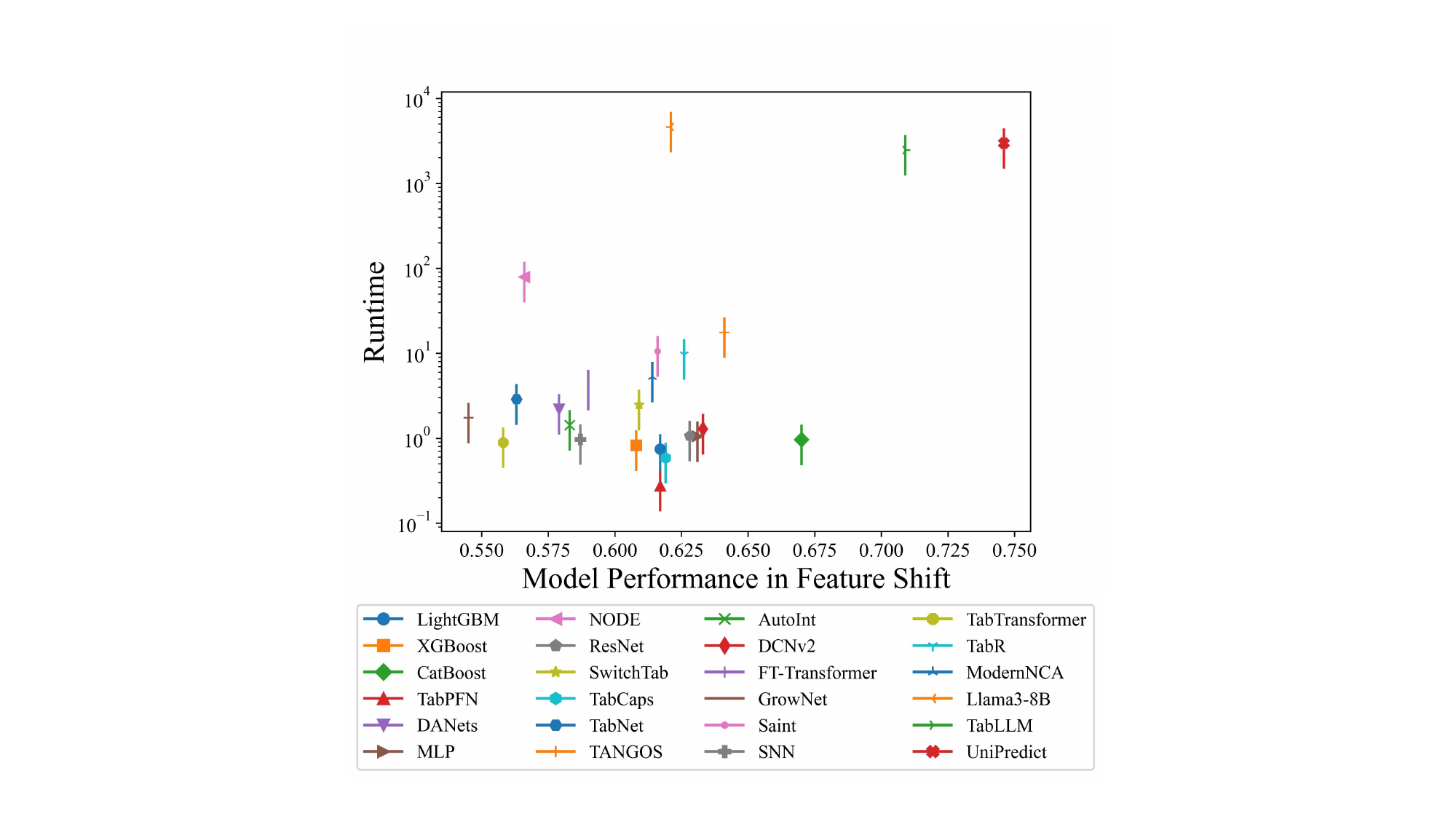}}
\caption{Average feature-shift performance vs. average runtime for each model.}
\label{experiment 3}
\end{center}
\end{figure*}

\subsection{Dataset-Specific Experiment Details}
\label{appendix:E}
This section provides detailed dataset-specific experimental results. For each dataset, we list all models and their corresponding performance metrics. The experiments are categorized into four scenarios: \textbf{SC} (single-column missing), \textbf{MC-D} (multi-column missing in Pearson correlation descending order), \textbf{MC-A} (multi-column missing in Pearson correlation ascending order), and \textbf{MC-R} (multi-column missing randomly). For classification tasks, we report accuracy and ROC-AUC scores, while for regression tasks, we provide RMSE results.

\begin{table}
\caption{Model performance on Credit dataset.}
\subtable[Model performance of SC experiment under accuracy on Credit dataset.]{
\resizebox{\textwidth}{24mm}{
\begin{tabular}{ccccccccccccccccccccccccccc}
\toprule
Column & PCC              & \multicolumn{1}{c}{LGB} & \multicolumn{1}{c}{XGB} & \multicolumn{1}{c}{CATB} & \multicolumn{1}{c}{PFN} & \multicolumn{1}{c}{DAN} & \multicolumn{1}{c}{MLP} & \multicolumn{1}{c}{NODE} & \multicolumn{1}{c}{RES} & \multicolumn{1}{c}{SWI} & \multicolumn{1}{c}{CAP} & \multicolumn{1}{c}{NET} & \multicolumn{1}{c}{GOS} & \multicolumn{1}{c}{INT} & \multicolumn{1}{c}{DCN} & \multicolumn{1}{c}{FTT} & \multicolumn{1}{c}{GRO} & \multicolumn{1}{c}{SAT} & \multicolumn{1}{c}{SNN} & \multicolumn{1}{c}{TTF} & \multicolumn{1}{c}{TABR} & \multicolumn{1}{c}{NCA} & \multicolumn{1}{c}{LMA} & \multicolumn{1}{c}{TLLM} & \multicolumn{1}{c}{UNI} & AVE   \\
\midrule
0   & \textbackslash{} & 0.757    & 0.750   & 0.775    & 0.762  & 0.695  & 0.693 & 0.693 & 0.680  & 0.695     & 0.668   & 0.525  & 0.695  & 0.688   & 0.685 & 0.695          & 0.608   & 0.695 & 0.537 & 0.528          & 0.687 & 0.717     & 0.750     & 0.630  & 0.620      & 0.676   \\
1   & -0.003           & 0.755    & 0.753   & 0.773    & 0.753  & 0.695  & 0.692 & 0.693 & 0.683  & 0.695     & 0.678   & 0.540  & 0.695  & 0.682   & 0.687 & 0.695          & 0.607   & 0.695 & 0.543 & 0.537          & 0.697 & 0.713     & 0.645     & 0.580  & 0.615      & 0.671   \\
2   & 0.003            & 0.755    & 0.750   & 0.772    & 0.755  & 0.695  & 0.695 & 0.692 & 0.673  & 0.695     & 0.660   & 0.528  & 0.695  & 0.688   & 0.678 & 0.695          & 0.613   & 0.695 & 0.540 & 0.545          & 0.685 & 0.715     & 0.650     & 0.605  & 0.620      & 0.671   \\
3   & -0.009           & 0.733    & 0.748   & 0.780    & 0.763  & 0.695  & 0.697 & 0.695 & 0.692  & 0.695     & 0.677   & 0.582  & 0.695  & 0.682   & 0.685 & 0.695          & 0.607   & 0.695 & 0.532 & 0.528          & 0.683 & 0.713     & 0.645     & 0.595  & 0.625      & 0.672   \\
\bottomrule
\end{tabular}
}
}
\subtable[Model performance of SC experiment under ROC-AUC on Credit dataset.]{
\resizebox{\textwidth}{24mm}{
\begin{tabular}{ccccccccccccccccccccccccccc}
\toprule
Column & PCC & \multicolumn{1}{c}{LGB} & \multicolumn{1}{c}{XGB} & \multicolumn{1}{c}{CATB} & \multicolumn{1}{c}{PFN} & \multicolumn{1}{c}{DAN} & \multicolumn{1}{c}{MLP} & \multicolumn{1}{c}{NODE} & \multicolumn{1}{c}{RES} & \multicolumn{1}{c}{SWI} & \multicolumn{1}{c}{CAP} & \multicolumn{1}{c}{NET} & \multicolumn{1}{c}{GOS} & \multicolumn{1}{c}{INT} & \multicolumn{1}{c}{DCN} & \multicolumn{1}{c}{FTT} & \multicolumn{1}{c}{GRO} & \multicolumn{1}{c}{SAT} & \multicolumn{1}{c}{SNN} & \multicolumn{1}{c}{TTF} & \multicolumn{1}{c}{TABR} & \multicolumn{1}{c}{NCA} & \multicolumn{1}{c}{LMA} & \multicolumn{1}{c}{TLLM} & \multicolumn{1}{c}{UNI} & AVE   \\
\midrule
0   & \textbackslash{} & 0.761    & 0.755   & 0.799    & 0.781  & 0.486  & 0.504 & 0.644 & 0.697  & 0.699     & 0.650   & 0.496  & 0.592  & 0.538   & 0.591 & 0.576          & 0.470   & 0.732 & 0.559 & 0.528          & 0.691 & 0.755     & 0.701     & 0.580  & 0.642      & 0.634   \\
1   & -0.003           & 0.769    & 0.762   & 0.804    & 0.782  & 0.483  & 0.504 & 0.641 & 0.696  & 0.698     & 0.653   & 0.490  & 0.595  & 0.545   & 0.594 & 0.576          & 0.470   & 0.736 & 0.563 & 0.530          & 0.697 & 0.750     & 0.615     & 0.508  & 0.631      & 0.629   \\
2   & 0.003            & 0.762    & 0.755   & 0.794    & 0.777  & 0.487  & 0.510 & 0.641 & 0.688  & 0.700     & 0.645   & 0.484  & 0.599  & 0.552   & 0.611 & 0.563          & 0.470   & 0.738 & 0.566 & 0.528          & 0.692 & 0.754     & 0.569     & 0.548  & 0.654      & 0.629   \\
3   & -0.009           & 0.754    & 0.746   & 0.805    & 0.784  & 0.471  & 0.493 & 0.645 & 0.692  & 0.703     & 0.659   & 0.555  & 0.588  & 0.532   & 0.587 & 0.578          & 0.478   & 0.733 & 0.564 & 0.527          & 0.691 & 0.757     & 0.536     & 0.527  & 0.607      & 0.625   \\
\bottomrule
\end{tabular}
}
}
\subtable[Model performance of MC-M experiment under accuracy on Credit dataset.]{
\resizebox{\textwidth}{24mm}{
\begin{tabular}{cccccccccccccccccccccccccc}
\toprule
Degree & \multicolumn{1}{c}{LGB} & \multicolumn{1}{c}{XGB} & \multicolumn{1}{c}{CATB} & \multicolumn{1}{c}{PFN} & \multicolumn{1}{c}{DAN} & \multicolumn{1}{c}{MLP} & \multicolumn{1}{c}{NODE} & \multicolumn{1}{c}{RES} & \multicolumn{1}{c}{SWI} & \multicolumn{1}{c}{CAP} & \multicolumn{1}{c}{NET} & \multicolumn{1}{c}{GOS} & \multicolumn{1}{c}{INT} & \multicolumn{1}{c}{DCN} & \multicolumn{1}{c}{FTT} & \multicolumn{1}{c}{GRO} & \multicolumn{1}{c}{SAT} & \multicolumn{1}{c}{SNN} & \multicolumn{1}{c}{TTF} & \multicolumn{1}{c}{TABR} & \multicolumn{1}{c}{NCA} & \multicolumn{1}{c}{LMA} & \multicolumn{1}{c}{TLLM} & \multicolumn{1}{c}{UNI} & AVE   \\
\midrule
0     & 0.757    & 0.750   & 0.775    & 0.762  & 0.695  & 0.693 & 0.693 & 0.680  & 0.695     & 0.668   & 0.525  & 0.695  & 0.688   & 0.685 & 0.695          & 0.608   & 0.695 & 0.537 & 0.528          & 0.687 & 0.717     & 0.750     & 0.630  & 0.620      & 0.676   \\
5\%   & 0.727    & 0.645   & 0.735    & 0.728  & 0.695  & 0.693 & 0.693 & 0.678  & 0.695     & 0.675   & 0.552  & 0.695  & 0.683   & 0.685 & 0.695          & 0.627   & 0.695 & 0.503 & 0.532          & 0.668 & 0.693     & 0.700     & 0.590  & 0.580      & 0.661   \\
10\%  & 0.712    & 0.600   & 0.700    & 0.723  & 0.695  & 0.697 & 0.692 & 0.675  & 0.695     & 0.653   & 0.562  & 0.695  & 0.685   & 0.685 & 0.695          & 0.645   & 0.695 & 0.507 & 0.543          & 0.677 & 0.702     & 0.645     & 0.585  & 0.600      & 0.657   \\
15\%  & 0.727    & 0.643   & 0.705    & 0.720  & 0.695  & 0.692 & 0.693 & 0.668  & 0.695     & 0.658   & 0.557  & 0.695  & 0.687   & 0.675 & 0.695          & 0.643   & 0.695 & 0.507 & 0.553          & 0.678 & 0.683     & 0.585     & 0.585  & 0.570      & 0.654   \\
\bottomrule
\end{tabular}
}
}
\subtable[Model performance of MC-M experiment under ROC-AUC on Credit dataset.]{
\resizebox{\textwidth}{24mm}{
\begin{tabular}{cccccccccccccccccccccccccc}
\toprule
Degree & \multicolumn{1}{c}{LGB} & \multicolumn{1}{c}{XGB} & \multicolumn{1}{c}{CATB} & \multicolumn{1}{c}{PFN} & \multicolumn{1}{c}{DAN} & \multicolumn{1}{c}{MLP} & \multicolumn{1}{c}{NODE} & \multicolumn{1}{c}{RES} & \multicolumn{1}{c}{SWI} & \multicolumn{1}{c}{CAP} & \multicolumn{1}{c}{NET} & \multicolumn{1}{c}{GOS} & \multicolumn{1}{c}{INT} & \multicolumn{1}{c}{DCN} & \multicolumn{1}{c}{FTT} & \multicolumn{1}{c}{GRO} & \multicolumn{1}{c}{SAT} & \multicolumn{1}{c}{SNN} & \multicolumn{1}{c}{TTF} & \multicolumn{1}{c}{TABR} & \multicolumn{1}{c}{NCA} & \multicolumn{1}{c}{LMA} & \multicolumn{1}{c}{TLLM} & \multicolumn{1}{c}{UNI} & AVE   \\
\midrule
0     & 0.761    & 0.755   & 0.799    & 0.781  & 0.486  & 0.504 & 0.644 & 0.697  & 0.699     & 0.650   & 0.496  & 0.592  & 0.538   & 0.591 & 0.576          & 0.470   & 0.732 & 0.559 & 0.528          & 0.691 & 0.755     & 0.701     & 0.580  & 0.642      & 0.634   \\
5\%   & 0.680    & 0.679   & 0.712    & 0.692  & 0.477  & 0.494 & 0.609 & 0.650  & 0.649     & 0.596   & 0.520  & 0.546  & 0.537   & 0.568 & 0.550          & 0.472   & 0.688 & 0.538 & 0.519          & 0.630 & 0.690     & 0.565     & 0.538  & 0.633      & 0.593   \\
10\%  & 0.665    & 0.658   & 0.657    & 0.672  & 0.478  & 0.475 & 0.581 & 0.641  & 0.631     & 0.586   & 0.532  & 0.525  & 0.528   & 0.571 & 0.535          & 0.477   & 0.686 & 0.529 & 0.524          & 0.618 & 0.684     & 0.561     & 0.514  & 0.685      & 0.584   \\
15\%  & 0.662    & 0.668   & 0.641    & 0.642  & 0.486  & 0.494 & 0.579 & 0.621  & 0.602     & 0.587   & 0.522  & 0.522  & 0.540   & 0.556 & 0.527          & 0.453   & 0.651 & 0.524 & 0.515          & 0.601 & 0.655     & 0.558     & 0.510  & 0.650      & 0.574   \\
\bottomrule
\end{tabular}
}
}
\label{table6}
\end{table}
\newpage
\begin{table}
\subtable[Model performance of MC-L experiment under accuracy on Credit dataset.]{
\resizebox{\textwidth}{24mm}{
\begin{tabular}{cccccccccccccccccccccccccc}
\toprule
Degree & \multicolumn{1}{c}{LGB} & \multicolumn{1}{c}{XGB} & \multicolumn{1}{c}{CATB} & \multicolumn{1}{c}{PFN} & \multicolumn{1}{c}{DAN} & \multicolumn{1}{c}{MLP} & \multicolumn{1}{c}{NODE} & \multicolumn{1}{c}{RES} & \multicolumn{1}{c}{SWI} & \multicolumn{1}{c}{CAP} & \multicolumn{1}{c}{NET} & \multicolumn{1}{c}{GOS} & \multicolumn{1}{c}{INT} & \multicolumn{1}{c}{DCN} & \multicolumn{1}{c}{FTT} & \multicolumn{1}{c}{GRO} & \multicolumn{1}{c}{SAT} & \multicolumn{1}{c}{SNN} & \multicolumn{1}{c}{TTF} & \multicolumn{1}{c}{TABR} & \multicolumn{1}{c}{NCA} & \multicolumn{1}{c}{LMA} & \multicolumn{1}{c}{TLLM} & \multicolumn{1}{c}{UNI} & AVE   \\
\midrule
0     & 0.757    & 0.750   & 0.775    & 0.762  & 0.695  & 0.693 & 0.693 & 0.680  & 0.695     & 0.668   & 0.525  & 0.695  & 0.688   & 0.685 & 0.695          & 0.608   & 0.695 & 0.537 & 0.528          & 0.687 & 0.717     & 0.750     & 0.630  & 0.620      & 0.676   \\
5\%   & 0.755    & 0.753   & 0.773    & 0.753  & 0.695  & 0.692 & 0.693 & 0.683  & 0.695     & 0.678   & 0.540  & 0.695  & 0.682   & 0.687 & 0.695          & 0.607   & 0.695 & 0.543 & 0.537          & 0.697 & 0.713     & 0.645     & 0.580  & 0.615      & 0.671   \\
10\%  & 0.755    & 0.753   & 0.775    & 0.748  & 0.695  & 0.692 & 0.690 & 0.668  & 0.695     & 0.678   & 0.537  & 0.695  & 0.682   & 0.678 & 0.695          & 0.615   & 0.695 & 0.553 & 0.538          & 0.695 & 0.710     & 0.630     & 0.580  & 0.615      & 0.670   \\
15\%  & 0.743    & 0.750   & 0.772    & 0.760  & 0.695  & 0.695 & 0.695 & 0.677  & 0.695     & 0.672   & 0.578  & 0.695  & 0.680   & 0.690 & 0.695          & 0.607   & 0.695 & 0.560 & 0.543          & 0.682 & 0.710     & 0.535     & 0.575  & 0.595      & 0.666   \\
\bottomrule
\end{tabular}
}
}

\subtable[Model performance of MC-L experiment under ROC-AUC on Credit dataset.]{
\resizebox{\textwidth}{24mm}{
\begin{tabular}{cccccccccccccccccccccccccc}
\toprule
Degree & \multicolumn{1}{c}{LGB} & \multicolumn{1}{c}{XGB} & \multicolumn{1}{c}{CATB} & \multicolumn{1}{c}{PFN} & \multicolumn{1}{c}{DAN} & \multicolumn{1}{c}{MLP} & \multicolumn{1}{c}{NODE} & \multicolumn{1}{c}{RES} & \multicolumn{1}{c}{SWI} & \multicolumn{1}{c}{CAP} & \multicolumn{1}{c}{NET} & \multicolumn{1}{c}{GOS} & \multicolumn{1}{c}{INT} & \multicolumn{1}{c}{DCN} & \multicolumn{1}{c}{FTT} & \multicolumn{1}{c}{GRO} & \multicolumn{1}{c}{SAT} & \multicolumn{1}{c}{SNN} & \multicolumn{1}{c}{TTF} & \multicolumn{1}{c}{TABR} & \multicolumn{1}{c}{NCA} & \multicolumn{1}{c}{LMA} & \multicolumn{1}{c}{TLLM} & \multicolumn{1}{c}{UNI} & AVE   \\
\midrule
0     & 0.761    & 0.755   & 0.799    & 0.781  & 0.486  & 0.504 & 0.644 & 0.697  & 0.699     & 0.650   & 0.496  & 0.592  & 0.538   & 0.591 & 0.576          & 0.470   & 0.732 & 0.559 & 0.528          & 0.691 & 0.755     & 0.701     & 0.580  & 0.642      & 0.634   \\
5\%   & 0.769    & 0.762   & 0.804    & 0.782  & 0.483  & 0.504 & 0.641 & 0.696  & 0.698     & 0.653   & 0.490  & 0.595  & 0.545   & 0.594 & 0.576          & 0.470   & 0.736 & 0.563 & 0.530          & 0.697 & 0.750     & 0.615     & 0.508  & 0.631      & 0.629   \\
10\%  & 0.770    & 0.762   & 0.799    & 0.779  & 0.481  & 0.511 & 0.637 & 0.692  & 0.700     & 0.649   & 0.499  & 0.603  & 0.554   & 0.612 & 0.562          & 0.473   & 0.742 & 0.573 & 0.528          & 0.701 & 0.749     & 0.619     & 0.510  & 0.627      & 0.630   \\
15\%  & 0.762    & 0.754   & 0.809    & 0.782  & 0.476  & 0.497 & 0.639 & 0.693  & 0.704     & 0.657   & 0.541  & 0.600  & 0.542   & 0.615 & 0.563          & 0.487   & 0.742 & 0.579 & 0.520          & 0.695 & 0.750     & 0.586     & 0.501  & 0.612      & 0.629   \\
\bottomrule
\end{tabular}
}
}
\subtable[Model performance of RD experiment under accuracy on Credit dataset.]{
\resizebox{\textwidth}{24mm}{
\begin{tabular}{cccccccccccccccccccccccccc}
\toprule
Degree & \multicolumn{1}{c}{LGB} & \multicolumn{1}{c}{XGB} & \multicolumn{1}{c}{CATB} & \multicolumn{1}{c}{PFN} & \multicolumn{1}{c}{DAN} & \multicolumn{1}{c}{MLP} & \multicolumn{1}{c}{NODE} & \multicolumn{1}{c}{RES} & \multicolumn{1}{c}{SWI} & \multicolumn{1}{c}{CAP} & \multicolumn{1}{c}{NET} & \multicolumn{1}{c}{GOS} & \multicolumn{1}{c}{INT} & \multicolumn{1}{c}{DCN} & \multicolumn{1}{c}{FTT} & \multicolumn{1}{c}{GRO} & \multicolumn{1}{c}{SAT} & \multicolumn{1}{c}{SNN} & \multicolumn{1}{c}{TTF} & \multicolumn{1}{c}{TABR} & \multicolumn{1}{c}{NCA} & \multicolumn{1}{c}{LMA} & \multicolumn{1}{c}{TLLM} & \multicolumn{1}{c}{UNI} & AVE   \\
\midrule
0     & 0.757    & 0.750   & 0.775    & 0.762  & 0.695  & 0.693 & 0.693 & 0.680  & 0.695     & 0.668   & 0.525  & 0.695  & 0.688   & 0.685 & 0.695          & 0.608   & 0.695 & 0.537 & 0.528          & 0.687 & 0.717     & 0.750     & 0.630  & 0.620      & 0.676   \\
5\%   & 0.748    & 0.740   & 0.770    & 0.750  & 0.695  & 0.695 & 0.693 & 0.681  & 0.695     & 0.662   & 0.531  & 0.695  & 0.683   & 0.685 & 0.695          & 0.620   & 0.695 & 0.531 & 0.533          & 0.686 & 0.716     & 0.732     & 0.597  & 0.606      & 0.672   \\
10\%  & 0.741    & 0.729   & 0.767    & 0.746  & 0.695  & 0.696 & 0.693 & 0.678  & 0.695     & 0.661   & 0.537  & 0.695  & 0.681   & 0.680 & 0.695          & 0.621   & 0.695 & 0.530 & 0.535          & 0.690 & 0.709     & 0.734     & 0.590  & 0.601      & 0.670   \\
15\%  & 0.736    & 0.719   & 0.764    & 0.743  & 0.695  & 0.697 & 0.695 & 0.676  & 0.695     & 0.657   & 0.561  & 0.695  & 0.680   & 0.684 & 0.695          & 0.629   & 0.695 & 0.534 & 0.535          & 0.683 & 0.711     & 0.731     & 0.586  & 0.599      & 0.671   \\
\bottomrule
\end{tabular}
}
}
\subtable[Model performance of RD experiment under ROC-AUC on Credit dataset.]{
\resizebox{\textwidth}{24mm}{
\begin{tabular}{cccccccccccccccccccccccccc}
\toprule
Degree & \multicolumn{1}{c}{LGB} & \multicolumn{1}{c}{XGB} & \multicolumn{1}{c}{CATB} & \multicolumn{1}{c}{PFN} & \multicolumn{1}{c}{DAN} & \multicolumn{1}{c}{MLP} & \multicolumn{1}{c}{NODE} & \multicolumn{1}{c}{RES} & \multicolumn{1}{c}{SWI} & \multicolumn{1}{c}{CAP} & \multicolumn{1}{c}{NET} & \multicolumn{1}{c}{GOS} & \multicolumn{1}{c}{INT} & \multicolumn{1}{c}{DCN} & \multicolumn{1}{c}{FTT} & \multicolumn{1}{c}{GRO} & \multicolumn{1}{c}{SAT} & \multicolumn{1}{c}{SNN} & \multicolumn{1}{c}{TTF} & \multicolumn{1}{c}{TABR} & \multicolumn{1}{c}{NCA} & \multicolumn{1}{c}{LMA} & \multicolumn{1}{c}{TLLM} & \multicolumn{1}{c}{UNI} & AVE   \\
\midrule
0     & 0.761    & 0.755   & 0.799    & 0.781  & 0.486  & 0.504 & 0.644 & 0.697  & 0.699     & 0.650   & 0.496  & 0.592  & 0.538   & 0.591 & 0.576          & 0.470   & 0.732 & 0.559 & 0.528          & 0.691 & 0.755     & 0.701     & 0.580  & 0.642      & 0.634   \\
5\%   & 0.754    & 0.748   & 0.795    & 0.771  & 0.483  & 0.502 & 0.641 & 0.687  & 0.690     & 0.642   & 0.499  & 0.585  & 0.540   & 0.586 & 0.571          & 0.471   & 0.726 & 0.555 & 0.530          & 0.684 & 0.745     & 0.650     & 0.538  & 0.623      & 0.626   \\
10\%  & 0.748    & 0.742   & 0.785    & 0.763  & 0.485  & 0.498 & 0.640 & 0.688  & 0.689     & 0.643   & 0.495  & 0.589  & 0.547   & 0.585 & 0.564          & 0.474   & 0.722 & 0.558 & 0.530          & 0.687 & 0.740     & 0.643     & 0.525  & 0.615      & 0.623   \\
15\%  & 0.741    & 0.735   & 0.779    & 0.755  & 0.479  & 0.506 & 0.631 & 0.674  & 0.676     & 0.634   & 0.524  & 0.576  & 0.528   & 0.577 & 0.574          & 0.475   & 0.710 & 0.550 & 0.520          & 0.678 & 0.727     & 0.637     & 0.518  & 0.655      & 0.619   \\
\bottomrule
\end{tabular}
}
}
\end{table}
\begin{table}
\subtable[Model performance gap $\Delta$ of RD experiment under accuracy on Credit dataset. Partial model performance degradation of 0.000 as the table retains three decimal places.]{
\resizebox{\textwidth}{24mm}{
\begin{tabular}{cccccccccccccccccccccccccc}
\toprule
Degree & \multicolumn{1}{c}{LGB} & \multicolumn{1}{c}{XGB} & \multicolumn{1}{c}{CATB} & \multicolumn{1}{c}{PFN} & \multicolumn{1}{c}{DAN} & \multicolumn{1}{c}{MLP} & \multicolumn{1}{c}{NODE} & \multicolumn{1}{c}{RES} & \multicolumn{1}{c}{SWI} & \multicolumn{1}{c}{CAP} & \multicolumn{1}{c}{NET} & \multicolumn{1}{c}{GOS} & \multicolumn{1}{c}{INT} & \multicolumn{1}{c}{DCN} & \multicolumn{1}{c}{FTT} & \multicolumn{1}{c}{GRO} & \multicolumn{1}{c}{SAT} & \multicolumn{1}{c}{SNN} & \multicolumn{1}{c}{TTF} & \multicolumn{1}{c}{TABR} & \multicolumn{1}{c}{NCA} & \multicolumn{1}{c}{LMA} & \multicolumn{1}{c}{TLLM} & \multicolumn{1}{c}{UNI} & AVE   \\
\midrule
5\%   & -0.012   & -0.014  & -0.006   & -0.015 & 0.000  & 0.002  & 0.000  & 0.001  & 0.000     & -0.010  & 0.011  & 0.000  & -0.008  & 0.000  & 0.000          & 0.019   & 0.000 & -0.011 & 0.009          & 0.000  & -0.002    & -0.024    & -0.053 & -0.023     & -0.006  \\
10\%  & -0.020   & -0.028  & -0.011   & -0.020 & 0.000  & 0.004  & 0.000  & -0.003 & 0.000     & -0.011  & 0.022  & 0.000  & -0.011  & -0.007 & 0.000          & 0.021   & 0.000 & -0.013 & 0.012          & 0.005  & -0.010    & -0.021    & -0.064 & -0.031     & -0.008  \\
15\%  & -0.027   & -0.041  & -0.015   & -0.024 & 0.000  & 0.005  & 0.002  & -0.006 & 0.000     & -0.017  & 0.068  & 0.000  & -0.012  & -0.001 & 0.000          & 0.034   & 0.000 & -0.005 & 0.013          & -0.005 & -0.008    & -0.025    & -0.070 & -0.034     & -0.008  \\
20\%  & -0.046   & -0.099  & -0.020   & -0.027 & 0.000  & 0.003  & -0.001 & -0.014 & 0.000     & -0.025  & 0.062  & 0.000  & -0.011  & -0.004 & 0.000          & 0.030   & 0.000 & -0.004 & 0.011          & -0.006 & -0.015    & -0.030    & -0.059 & -0.036     & -0.014  \\
\bottomrule
\end{tabular}
}
}
\subtable[Model performance gap $\Delta$ of RD experiment under ROC-AUC on Credit dataset.]{
\resizebox{\textwidth}{24mm}{
\begin{tabular}{cccccccccccccccccccccccccc}
\toprule
Degree & \multicolumn{1}{c}{LGB} & \multicolumn{1}{c}{XGB} & \multicolumn{1}{c}{CATB} & \multicolumn{1}{c}{PFN} & \multicolumn{1}{c}{DAN} & \multicolumn{1}{c}{MLP} & \multicolumn{1}{c}{NODE} & \multicolumn{1}{c}{RES} & \multicolumn{1}{c}{SWI} & \multicolumn{1}{c}{CAP} & \multicolumn{1}{c}{NET} & \multicolumn{1}{c}{GOS} & \multicolumn{1}{c}{INT} & \multicolumn{1}{c}{DCN} & \multicolumn{1}{c}{FTT} & \multicolumn{1}{c}{GRO} & \multicolumn{1}{c}{SAT} & \multicolumn{1}{c}{SNN} & \multicolumn{1}{c}{TTF} & \multicolumn{1}{c}{TABR} & \multicolumn{1}{c}{NCA} & \multicolumn{1}{c}{LMA} & \multicolumn{1}{c}{TLLM} & \multicolumn{1}{c}{UNI} & AVE   \\
\midrule
5\%   & -0.008   & -0.009  & -0.005   & -0.012 & -0.007 & -0.004 & -0.006 & -0.015 & -0.012    & -0.012  & 0.006  & -0.011 & 0.003   & -0.008 & -0.009         & 0.004   & -0.008 & -0.007 & 0.004          & -0.011 & -0.014    & -0.073    & -0.073 & -0.029     & -0.014  \\
10\%  & -0.017   & -0.017  & -0.018   & -0.022 & -0.002 & -0.011 & -0.007 & -0.013 & -0.013    & -0.011  & -0.002 & -0.005 & 0.016   & -0.009 & -0.020         & 0.010   & -0.013 & -0.002 & 0.003          & -0.007 & -0.020    & -0.083    & -0.096 & -0.042     & -0.018  \\
15\%  & -0.026   & -0.026  & -0.025   & -0.032 & -0.015 & 0.005  & -0.021 & -0.033 & -0.032    & -0.024  & 0.056  & -0.027 & -0.019  & -0.022 & -0.002         & 0.011   & -0.030 & -0.016 & -0.015         & -0.019 & -0.038    & -0.092    & -0.108 & 0.020      & -0.024  \\
20\%  & -0.050   & -0.049  & -0.035   & -0.061 & -0.007 & 0.009  & -0.032 & -0.043 & -0.035    & -0.042  & 0.033  & -0.030 & 0.027   & -0.043 & -0.030         & 0.002   & -0.036 & -0.007 & -0.017         & -0.034 & -0.042    & -0.127    & -0.095 & -0.030     & -0.035  \\
\bottomrule
\end{tabular}
}
}
\end{table}

\begin{table}
\caption{Model performance on Electricity dataset.}
\subtable[Model performance of SC experiment under accuracy on Electricity dataset.]{
\resizebox{\textwidth}{14mm}{
\begin{tabular}{ccccccccccccccccccccccccccc}
\toprule
Column & PCC              & \multicolumn{1}{c}{LGB} & \multicolumn{1}{c}{XGB} & \multicolumn{1}{c}{CATB} & \multicolumn{1}{c}{PFN} & \multicolumn{1}{c}{DAN} & \multicolumn{1}{c}{MLP} & \multicolumn{1}{c}{NODE} & \multicolumn{1}{c}{RES} & \multicolumn{1}{c}{SWI} & \multicolumn{1}{c}{CAP} & \multicolumn{1}{c}{NET} & \multicolumn{1}{c}{GOS} & \multicolumn{1}{c}{INT} & \multicolumn{1}{c}{DCN} & \multicolumn{1}{c}{FTT} & \multicolumn{1}{c}{GRO} & \multicolumn{1}{c}{SAT} & \multicolumn{1}{c}{SNN} & \multicolumn{1}{c}{TTF} & \multicolumn{1}{c}{TABR} & \multicolumn{1}{c}{NCA} & \multicolumn{1}{c}{LMA} & \multicolumn{1}{c}{TLLM} & \multicolumn{1}{c}{UNI} & AVE   \\
\midrule
0   & \textbackslash{} & 0.913    & 0.913   & 0.953    & 0.861  & 0.847  & 0.863 & 0.729 & 0.873  & 0.829     & 0.843   & 0.857  & 0.871  & 0.861   & 0.862 & 0.863          & 0.794   & 0.878 & 0.834 & 0.720          & 0.940 & 0.985     & 0.804     & 0.921  & 0.858      & 0.861   \\
1   & -0.006           & 0.817    & 0.818   & 0.840    & 0.842  & 0.832  & 0.844 & 0.731 & 0.847  & 0.819     & 0.830   & 0.834  & 0.847  & 0.843   & 0.843 & 0.838          & 0.772   & 0.845 & 0.821 & 0.716          & 0.839 & 0.764     & 0.500     & 0.887  & 0.778      & 0.806   \\
2   & -0.048           & 0.905    & 0.906   & 0.925    & 0.851  & 0.839  & 0.852 & 0.727 & 0.859  & 0.828     & 0.835   & 0.847  & 0.857  & 0.850   & 0.850 & 0.853          & 0.798   & 0.857 & 0.828 & 0.751          & 0.852 & 0.881     & 0.501     & 0.888  & 0.864      & 0.834   \\
3   & 0.065            & 0.906    & 0.905   & 0.945    & 0.849  & 0.848  & 0.863 & 0.728 & 0.873  & 0.836     & 0.843   & 0.856  & 0.871  & 0.860   & 0.862 & 0.863          & 0.794   & 0.876 & 0.834 & 0.722          & 0.937 & 0.976     & 0.501     & 0.920  & 0.858      & 0.847   \\
\bottomrule
\end{tabular}
}
}
\subtable[Model performance of SC experiment under ROC-AUC on Electricity dataset.]{
\resizebox{\textwidth}{14mm}{
\begin{tabular}{ccccccccccccccccccccccccccc}
\toprule
Column & PCC & \multicolumn{1}{c}{LGB} & \multicolumn{1}{c}{XGB} & \multicolumn{1}{c}{CATB} & \multicolumn{1}{c}{PFN} & \multicolumn{1}{c}{DAN} & \multicolumn{1}{c}{MLP} & \multicolumn{1}{c}{NODE} & \multicolumn{1}{c}{RES} & \multicolumn{1}{c}{SWI} & \multicolumn{1}{c}{CAP} & \multicolumn{1}{c}{NET} & \multicolumn{1}{c}{GOS} & \multicolumn{1}{c}{INT} & \multicolumn{1}{c}{DCN} & \multicolumn{1}{c}{FTT} & \multicolumn{1}{c}{GRO} & \multicolumn{1}{c}{SAT} & \multicolumn{1}{c}{SNN} & \multicolumn{1}{c}{TTF} & \multicolumn{1}{c}{TABR} & \multicolumn{1}{c}{NCA} & \multicolumn{1}{c}{LMA} & \multicolumn{1}{c}{TLLM} & \multicolumn{1}{c}{UNI} & AVE   \\
\midrule
0   & \textbackslash{} & 0.913    & 0.913   & 0.953    & 0.861  & 0.847  & 0.863 & 0.729 & 0.873  & 0.829     & 0.843   & 0.857  & 0.871  & 0.861   & 0.862 & 0.863          & 0.794   & 0.878 & 0.834 & 0.720          & 0.940 & 0.985     & 0.804     & 0.921  & 0.858      & 0.861   \\
1   & -0.006           & 0.817    & 0.818   & 0.840    & 0.842  & 0.832  & 0.844 & 0.731 & 0.847  & 0.819     & 0.830   & 0.834  & 0.847  & 0.843   & 0.843 & 0.838          & 0.772   & 0.845 & 0.821 & 0.716          & 0.839 & 0.764     & 0.500     & 0.887  & 0.778      & 0.806   \\
2   & -0.048           & 0.905    & 0.906   & 0.925    & 0.851  & 0.839  & 0.852 & 0.727 & 0.859  & 0.828     & 0.835   & 0.847  & 0.857  & 0.850   & 0.850 & 0.853          & 0.798   & 0.857 & 0.828 & 0.751          & 0.852 & 0.881     & 0.501     & 0.888  & 0.864      & 0.834   \\
3   & 0.065            & 0.906    & 0.905   & 0.945    & 0.849  & 0.848  & 0.863 & 0.728 & 0.873  & 0.836     & 0.843   & 0.856  & 0.871  & 0.860   & 0.862 & 0.863          & 0.794   & 0.876 & 0.834 & 0.722          & 0.937 & 0.976     & 0.501     & 0.920  & 0.858      & 0.847   \\
\bottomrule
\end{tabular}
}
}
\subtable[Model performance of MC-M experiment under accuracy on Electricity dataset.]{
\resizebox{\textwidth}{14mm}{
\begin{tabular}{cccccccccccccccccccccccccc}
\toprule
Degree & \multicolumn{1}{c}{LGB} & \multicolumn{1}{c}{XGB} & \multicolumn{1}{c}{CATB} & \multicolumn{1}{c}{PFN} & \multicolumn{1}{c}{DAN} & \multicolumn{1}{c}{MLP} & \multicolumn{1}{c}{NODE} & \multicolumn{1}{c}{RES} & \multicolumn{1}{c}{SWI} & \multicolumn{1}{c}{CAP} & \multicolumn{1}{c}{NET} & \multicolumn{1}{c}{GOS} & \multicolumn{1}{c}{INT} & \multicolumn{1}{c}{DCN} & \multicolumn{1}{c}{FTT} & \multicolumn{1}{c}{GRO} & \multicolumn{1}{c}{SAT} & \multicolumn{1}{c}{SNN} & \multicolumn{1}{c}{TTF} & \multicolumn{1}{c}{TABR} & \multicolumn{1}{c}{NCA} & \multicolumn{1}{c}{LMA} & \multicolumn{1}{c}{TLLM} & \multicolumn{1}{c}{UNI} & AVE   \\
\midrule
0      & 0.838    & 0.839   & 0.880    & 0.784  & 0.772  & 0.787 & 0.633 & 0.798  & 0.763     & 0.771   & 0.780  & 0.795  & 0.785   & 0.786 & 0.784          & 0.729   & 0.799 & 0.763 & 0.660          & 0.860 & 0.941     & 0.833     & 0.925  & 0.740      & 0.794   \\
12.5\% & 0.480    & 0.477   & 0.532    & 0.580  & 0.637  & 0.630 & 0.606 & 0.634  & 0.614     & 0.644   & 0.626  & 0.639  & 0.632   & 0.622 & 0.625          & 0.625   & 0.623 & 0.623 & 0.578          & 0.554 & 0.506     & 0.427     & 0.836  & 0.669      & 0.601   \\
25\%   & 0.482    & 0.480   & 0.518    & 0.492  & 0.552  & 0.514 & 0.580 & 0.510  & 0.587     & 0.621   & 0.571  & 0.542  & 0.562   & 0.517 & 0.597          & 0.581   & 0.583 & 0.522 & 0.555          & 0.527 & 0.503     & 0.427     & 0.835  & 0.667      & 0.555   \\
37.5\% & 0.474    & 0.473   & 0.518    & 0.491  & 0.541  & 0.530 & 0.572 & 0.531  & 0.567     & 0.604   & 0.559  & 0.559  & 0.570   & 0.525 & 0.576          & 0.549   & 0.576 & 0.519 & 0.545          & 0.505 & 0.499     & 0.427     & 0.834  & 0.668      & 0.551   \\
\bottomrule
\end{tabular}
}
}
\subtable[Model performance of MC-M experiment under ROC-AUC on Electricity dataset.]{
\resizebox{\textwidth}{14mm}{
\begin{tabular}{cccccccccccccccccccccccccc}
\toprule
Degree & \multicolumn{1}{c}{LGB} & \multicolumn{1}{c}{XGB} & \multicolumn{1}{c}{CATB} & \multicolumn{1}{c}{PFN} & \multicolumn{1}{c}{DAN} & \multicolumn{1}{c}{MLP} & \multicolumn{1}{c}{NODE} & \multicolumn{1}{c}{RES} & \multicolumn{1}{c}{SWI} & \multicolumn{1}{c}{CAP} & \multicolumn{1}{c}{NET} & \multicolumn{1}{c}{GOS} & \multicolumn{1}{c}{INT} & \multicolumn{1}{c}{DCN} & \multicolumn{1}{c}{FTT} & \multicolumn{1}{c}{GRO} & \multicolumn{1}{c}{SAT} & \multicolumn{1}{c}{SNN} & \multicolumn{1}{c}{TTF} & \multicolumn{1}{c}{TABR} & \multicolumn{1}{c}{NCA} & \multicolumn{1}{c}{LMA} & \multicolumn{1}{c}{TLLM} & \multicolumn{1}{c}{UNI} & AVE   \\
\midrule
0      & 0.913    & 0.913   & 0.953    & 0.861  & 0.847  & 0.863 & 0.729 & 0.873  & 0.829     & 0.843   & 0.857  & 0.871  & 0.861   & 0.862 & 0.863          & 0.794   & 0.878 & 0.834 & 0.720          & 0.940 & 0.985     & 0.804     & 0.921  & 0.858      & 0.861   \\
12.5\% & 0.485    & 0.480   & 0.571    & 0.613  & 0.697  & 0.665 & 0.687 & 0.679  & 0.675     & 0.700   & 0.666  & 0.691  & 0.669   & 0.657 & 0.658          & 0.665   & 0.647 & 0.675 & 0.587          & 0.552 & 0.504     & 0.500     & 0.833  & 0.788      & 0.639   \\
25\%   & 0.487    & 0.484   & 0.532    & 0.486  & 0.563  & 0.527 & 0.654 & 0.535  & 0.583     & 0.605   & 0.531  & 0.558  & 0.548   & 0.523 & 0.573          & 0.608   & 0.546 & 0.508 & 0.558          & 0.501 & 0.499     & 0.500     & 0.832  & 0.782      & 0.564   \\
37.5\% & 0.467    & 0.464   & 0.523    & 0.495  & 0.547  & 0.536 & 0.622 & 0.543  & 0.535     & 0.581   & 0.516  & 0.557  & 0.540   & 0.521 & 0.545          & 0.581   & 0.545 & 0.508 & 0.539          & 0.471 & 0.494     & 0.500     & 0.832  & 0.781      & 0.552   \\
\bottomrule
\end{tabular}
}
}
\subtable[Model performance of MC-L experiment under accuracy on Electricity dataset.]{
\resizebox{\textwidth}{14mm}{
\begin{tabular}{cccccccccccccccccccccccccc}
\toprule
Degree & \multicolumn{1}{c}{LGB} & \multicolumn{1}{c}{XGB} & \multicolumn{1}{c}{CATB} & \multicolumn{1}{c}{PFN} & \multicolumn{1}{c}{DAN} & \multicolumn{1}{c}{MLP} & \multicolumn{1}{c}{NODE} & \multicolumn{1}{c}{RES} & \multicolumn{1}{c}{SWI} & \multicolumn{1}{c}{CAP} & \multicolumn{1}{c}{NET} & \multicolumn{1}{c}{GOS} & \multicolumn{1}{c}{INT} & \multicolumn{1}{c}{DCN} & \multicolumn{1}{c}{FTT} & \multicolumn{1}{c}{GRO} & \multicolumn{1}{c}{SAT} & \multicolumn{1}{c}{SNN} & \multicolumn{1}{c}{TTF} & \multicolumn{1}{c}{TABR} & \multicolumn{1}{c}{NCA} & \multicolumn{1}{c}{LMA} & \multicolumn{1}{c}{TLLM} & \multicolumn{1}{c}{UNI} & AVE   \\
\midrule
0      & 0.838    & 0.839   & 0.880    & 0.784  & 0.772  & 0.787 & 0.633 & 0.798  & 0.763     & 0.771   & 0.780  & 0.795  & 0.785   & 0.786 & 0.784          & 0.729   & 0.799 & 0.763 & 0.660          & 0.860 & 0.941     & 0.833     & 0.925  & 0.740      & 0.794   \\
12.5\% & 0.633    & 0.636   & 0.655    & 0.770  & 0.762  & 0.766 & 0.633 & 0.768  & 0.754     & 0.760   & 0.762  & 0.769  & 0.767   & 0.762 & 0.716          & 0.705   & 0.759 & 0.759 & 0.651          & 0.659 & 0.705     & 0.427     & 0.889  & 0.678      & 0.714   \\
25\%   & 0.647    & 0.653   & 0.687    & 0.763  & 0.762  & 0.768 & 0.631 & 0.768  & 0.758     & 0.757   & 0.762  & 0.769  & 0.768   & 0.767 & 0.739          & 0.714   & 0.750 & 0.756 & 0.675          & 0.646 & 0.710     & 0.431     & 0.886  & 0.706      & 0.720   \\
37.5\% & 0.653    & 0.654   & 0.690    & 0.760  & 0.762  & 0.767 & 0.630 & 0.768  & 0.762     & 0.757   & 0.762  & 0.768  & 0.768   & 0.766 & 0.738          & 0.714   & 0.751 & 0.756 & 0.677          & 0.662 & 0.710     & 0.433     & 0.885  & 0.703      & 0.721   \\
\bottomrule
\end{tabular}
}
}

\subtable[Model performance of MC-L experiment under ROC-AUC on Electricity dataset.]{
\resizebox{\textwidth}{14mm}{
\begin{tabular}{cccccccccccccccccccccccccc}
\toprule
Degree & \multicolumn{1}{c}{LGB} & \multicolumn{1}{c}{XGB} & \multicolumn{1}{c}{CATB} & \multicolumn{1}{c}{PFN} & \multicolumn{1}{c}{DAN} & \multicolumn{1}{c}{MLP} & \multicolumn{1}{c}{NODE} & \multicolumn{1}{c}{RES} & \multicolumn{1}{c}{SWI} & \multicolumn{1}{c}{CAP} & \multicolumn{1}{c}{NET} & \multicolumn{1}{c}{GOS} & \multicolumn{1}{c}{INT} & \multicolumn{1}{c}{DCN} & \multicolumn{1}{c}{FTT} & \multicolumn{1}{c}{GRO} & \multicolumn{1}{c}{SAT} & \multicolumn{1}{c}{SNN} & \multicolumn{1}{c}{TTF} & \multicolumn{1}{c}{TABR} & \multicolumn{1}{c}{NCA} & \multicolumn{1}{c}{LMA} & \multicolumn{1}{c}{TLLM} & \multicolumn{1}{c}{UNI} & AVE   \\
\midrule
0      & 0.913    & 0.913   & 0.953    & 0.861  & 0.847  & 0.863 & 0.729 & 0.873  & 0.829     & 0.843   & 0.857  & 0.871  & 0.861   & 0.862 & 0.863          & 0.794   & 0.878 & 0.834 & 0.720          & 0.940 & 0.985     & 0.804     & 0.921  & 0.858      & 0.861   \\
12.5\% & 0.817    & 0.818   & 0.840    & 0.842  & 0.832  & 0.844 & 0.731 & 0.847  & 0.819     & 0.830   & 0.834  & 0.847  & 0.843   & 0.843 & 0.838          & 0.772   & 0.845 & 0.821 & 0.716          & 0.839 & 0.764     & 0.500     & 0.887  & 0.778      & 0.806   \\
25\%   & 0.811    & 0.812   & 0.831    & 0.834  & 0.823  & 0.834 & 0.728 & 0.836  & 0.820     & 0.826   & 0.823  & 0.835  & 0.835   & 0.832 & 0.826          & 0.775   & 0.828 & 0.816 & 0.763          & 0.787 & 0.773     & 0.500     & 0.877  & 0.817      & 0.802   \\
37.5\% & 0.801    & 0.801   & 0.817    & 0.828  & 0.824  & 0.834 & 0.728 & 0.836  & 0.824     & 0.826   & 0.823  & 0.834  & 0.835   & 0.831 & 0.826          & 0.775   & 0.828 & 0.816 & 0.767          & 0.788 & 0.771     & 0.503     & 0.875  & 0.817      & 0.800   \\
\bottomrule
\end{tabular}
}
}
\end{table}
\newpage
\begin{table}
\subtable[Model performance of RD experiment under accuracy on Electricity dataset.]{
\resizebox{\textwidth}{14mm}{
\begin{tabular}{cccccccccccccccccccccccccc}
\toprule
Degree & \multicolumn{1}{c}{LGB} & \multicolumn{1}{c}{XGB} & \multicolumn{1}{c}{CATB} & \multicolumn{1}{c}{PFN} & \multicolumn{1}{c}{DAN} & \multicolumn{1}{c}{MLP} & \multicolumn{1}{c}{NODE} & \multicolumn{1}{c}{RES} & \multicolumn{1}{c}{SWI} & \multicolumn{1}{c}{CAP} & \multicolumn{1}{c}{NET} & \multicolumn{1}{c}{GOS} & \multicolumn{1}{c}{INT} & \multicolumn{1}{c}{DCN} & \multicolumn{1}{c}{FTT} & \multicolumn{1}{c}{GRO} & \multicolumn{1}{c}{SAT} & \multicolumn{1}{c}{SNN} & \multicolumn{1}{c}{TTF} & \multicolumn{1}{c}{TABR} & \multicolumn{1}{c}{NCA} & \multicolumn{1}{c}{LMA} & \multicolumn{1}{c}{TLLM} & \multicolumn{1}{c}{UNI} & AVE   \\
\midrule
0      & 0.838    & 0.839   & 0.880    & 0.784  & 0.772  & 0.787 & 0.633 & 0.798  & 0.763     & 0.771   & 0.780  & 0.795  & 0.785   & 0.786 & 0.784          & 0.729   & 0.799 & 0.763 & 0.660          & 0.860 & 0.941     & 0.833     & 0.925  & 0.740      & 0.794   \\
12.5\% & 0.761    & 0.761   & 0.788    & 0.747  & 0.750  & 0.757 & 0.621 & 0.761  & 0.742     & 0.751   & 0.752  & 0.762  & 0.756   & 0.755 & 0.750          & 0.707   & 0.762 & 0.743 & 0.650          & 0.755 & 0.757     & 0.817     & 0.905  & 0.728      & 0.752   \\
25\%   & 0.693    & 0.693   & 0.716    & 0.714  & 0.727  & 0.726 & 0.610 & 0.727  & 0.721     & 0.729   & 0.724  & 0.731  & 0.727   & 0.724 & 0.715          & 0.682   & 0.729 & 0.721 & 0.638          & 0.663 & 0.664     & 0.764     & 0.877  & 0.704      & 0.713   \\
37.5\% & 0.633    & 0.633   & 0.651    & 0.682  & 0.702  & 0.694 & 0.599 & 0.694  & 0.697     & 0.706   & 0.697  & 0.701  & 0.696   & 0.692 & 0.676          & 0.657   & 0.698 & 0.697 & 0.625          & 0.611 & 0.564     & 0.649     & 0.881  & 0.703      & 0.677   \\
\bottomrule
\end{tabular}
}
}
\subtable[Model performance of RD experiment under ROC-AUC on Electricity dataset.]{
\resizebox{\textwidth}{14mm}{
\begin{tabular}{cccccccccccccccccccccccccc}
\toprule
Degree & \multicolumn{1}{c}{LGB} & \multicolumn{1}{c}{XGB} & \multicolumn{1}{c}{CATB} & \multicolumn{1}{c}{PFN} & \multicolumn{1}{c}{DAN} & \multicolumn{1}{c}{MLP} & \multicolumn{1}{c}{NODE} & \multicolumn{1}{c}{RES} & \multicolumn{1}{c}{SWI} & \multicolumn{1}{c}{CAP} & \multicolumn{1}{c}{NET} & \multicolumn{1}{c}{GOS} & \multicolumn{1}{c}{INT} & \multicolumn{1}{c}{DCN} & \multicolumn{1}{c}{FTT} & \multicolumn{1}{c}{GRO} & \multicolumn{1}{c}{SAT} & \multicolumn{1}{c}{SNN} & \multicolumn{1}{c}{TTF} & \multicolumn{1}{c}{TABR} & \multicolumn{1}{c}{NCA} & \multicolumn{1}{c}{LMA} & \multicolumn{1}{c}{TLLM} & \multicolumn{1}{c}{UNI} & AVE   \\
\midrule
0      & 0.913    & 0.913   & 0.953    & 0.861  & 0.847  & 0.863 & 0.729 & 0.873  & 0.829     & 0.843   & 0.857  & 0.871  & 0.861   & 0.862 & 0.863          & 0.794   & 0.878 & 0.834 & 0.720          & 0.940 & 0.985     & 0.804     & 0.921  & 0.858      & 0.861   \\
12.5\% & 0.844    & 0.844   & 0.876    & 0.816  & 0.821  & 0.828 & 0.726 & 0.835  & 0.807     & 0.818   & 0.821  & 0.836  & 0.828   & 0.827 & 0.831          & 0.768   & 0.835 & 0.807 & 0.705          & 0.835 & 0.807     & 0.814     & 0.901  & 0.843      & 0.820   \\
25\%   & 0.780    & 0.779   & 0.814    & 0.774  & 0.791  & 0.791 & 0.721 & 0.795  & 0.781     & 0.790   & 0.782  & 0.799  & 0.791   & 0.789 & 0.796          & 0.739   & 0.793 & 0.777 & 0.688          & 0.720 & 0.711     & 0.761     & 0.873  & 0.827      & 0.777   \\
37.5\% & 0.722    & 0.720   & 0.755    & 0.733  & 0.758  & 0.752 & 0.714 & 0.754  & 0.749     & 0.759   & 0.742  & 0.761  & 0.751   & 0.749 & 0.761          & 0.710   & 0.752 & 0.743 & 0.669          & 0.665 & 0.658     & 0.657     & 0.876  & 0.817      & 0.739   \\
\bottomrule
\end{tabular}
}
}
\subtable[Model performance gap $\Delta$ of RD experiment under accuracy on Electricity dataset.]{
\resizebox{\textwidth}{14mm}{
\begin{tabular}{cccccccccccccccccccccccccc}
\toprule
Degree & \multicolumn{1}{c}{LGB} & \multicolumn{1}{c}{XGB} & \multicolumn{1}{c}{CATB} & \multicolumn{1}{c}{PFN} & \multicolumn{1}{c}{DAN} & \multicolumn{1}{c}{MLP} & \multicolumn{1}{c}{NODE} & \multicolumn{1}{c}{RES} & \multicolumn{1}{c}{SWI} & \multicolumn{1}{c}{CAP} & \multicolumn{1}{c}{NET} & \multicolumn{1}{c}{GOS} & \multicolumn{1}{c}{INT} & \multicolumn{1}{c}{DCN} & \multicolumn{1}{c}{FTT} & \multicolumn{1}{c}{GRO} & \multicolumn{1}{c}{SAT} & \multicolumn{1}{c}{SNN} & \multicolumn{1}{c}{TTF} & \multicolumn{1}{c}{TABR} & \multicolumn{1}{c}{NCA} & \multicolumn{1}{c}{LMA} & \multicolumn{1}{c}{TLLM} & \multicolumn{1}{c}{UNI} & AVE   \\
\midrule
12.5\% & -0.092   & -0.093  & -0.105   & -0.047 & -0.028 & -0.038 & -0.019 & -0.046 & -0.027    & -0.026  & -0.036 & -0.042 & -0.036  & -0.039 & -0.043         & -0.031  & -0.046 & -0.026 & -0.015         & -0.123 & -0.195    & -0.020    & -0.021 & -0.017     & -0.053  \\
25\%   & -0.173   & -0.174  & -0.187   & -0.089 & -0.058 & -0.077 & -0.037 & -0.089 & -0.056    & -0.054  & -0.073 & -0.080 & -0.073  & -0.079 & -0.088         & -0.065  & -0.088 & -0.056 & -0.034         & -0.229 & -0.294    & -0.083    & -0.052 & -0.049     & -0.101  \\
37.5\% & -0.245   & -0.246  & -0.261   & -0.130 & -0.090 & -0.117 & -0.053 & -0.131 & -0.086    & -0.084  & -0.108 & -0.118 & -0.113  & -0.120 & -0.137         & -0.100  & -0.127 & -0.087 & -0.053         & -0.290 & -0.401    & -0.221    & -0.047 & -0.050     & -0.147  \\
50\%   & -0.309   & -0.309  & -0.327   & -0.172 & -0.085 & -0.114 & -0.054 & -0.131 & -0.096    & -0.093  & -0.110 & -0.116 & -0.119  & -0.124 & -0.175         & -0.124  & -0.141 & -0.093 & -0.058         & -0.326 & -0.428    & -0.232    & -0.062 & -0.074     & -0.167  \\
\bottomrule
\end{tabular}
}
}
\subtable[Model performance gap $\Delta$ of RD experiment under ROC-AUC on Electricity dataset.]{
\resizebox{\textwidth}{14mm}{
\begin{tabular}{cccccccccccccccccccccccccc}
\toprule
Degree & \multicolumn{1}{c}{LGB} & \multicolumn{1}{c}{XGB} & \multicolumn{1}{c}{CATB} & \multicolumn{1}{c}{PFN} & \multicolumn{1}{c}{DAN} & \multicolumn{1}{c}{MLP} & \multicolumn{1}{c}{NODE} & \multicolumn{1}{c}{RES} & \multicolumn{1}{c}{SWI} & \multicolumn{1}{c}{CAP} & \multicolumn{1}{c}{NET} & \multicolumn{1}{c}{GOS} & \multicolumn{1}{c}{INT} & \multicolumn{1}{c}{DCN} & \multicolumn{1}{c}{FTT} & \multicolumn{1}{c}{GRO} & \multicolumn{1}{c}{SAT} & \multicolumn{1}{c}{SNN} & \multicolumn{1}{c}{TTF} & \multicolumn{1}{c}{TABR} & \multicolumn{1}{c}{NCA} & \multicolumn{1}{c}{LMA} & \multicolumn{1}{c}{TLLM} & \multicolumn{1}{c}{UNI} & AVE   \\
\midrule
12.5\% & -0.075   & -0.076  & -0.080   & -0.052 & -0.031 & -0.041 & -0.005 & -0.043 & -0.026    & -0.029  & -0.042 & -0.040 & -0.038  & -0.041 & -0.037         & -0.034  & -0.049 & -0.032 & -0.021         & -0.111 & -0.181    & 0.013     & -0.022 & -0.018     & -0.048  \\
25\%   & -0.145   & -0.147  & -0.146   & -0.101 & -0.066 & -0.084 & -0.011 & -0.089 & -0.058    & -0.063  & -0.088 & -0.082 & -0.082  & -0.085 & -0.078         & -0.070  & -0.097 & -0.069 & -0.044         & -0.234 & -0.278    & -0.053    & -0.053 & -0.037     & -0.097  \\
37.5\% & -0.209   & -0.211  & -0.208   & -0.148 & -0.105 & -0.129 & -0.021 & -0.136 & -0.097    & -0.100  & -0.134 & -0.126 & -0.128  & -0.131 & -0.119         & -0.107  & -0.143 & -0.109 & -0.071         & -0.292 & -0.332    & -0.182    & -0.048 & -0.048     & -0.142  \\
50\%   & -0.266   & -0.269  & -0.266   & -0.196 & -0.110 & -0.133 & -0.023 & -0.148 & -0.112    & -0.108  & -0.136 & -0.136 & -0.136  & -0.141 & -0.117         & -0.121  & -0.145 & -0.121 & -0.073         & -0.305 & -0.354    & -0.194    & -0.068 & -0.052     & -0.160  \\
\bottomrule
\end{tabular}
}
}
\end{table}

\begin{table}
\caption{Model performance on Heart dataset.}
\subtable[Model performance of SC experiment under accuracy on Heart dataset.]{
\resizebox{\textwidth}{16mm}{
\begin{tabular}{ccccccccccccccccccccccccccc}
\toprule
Column & PCC              & \multicolumn{1}{c}{LGB} & \multicolumn{1}{c}{XGB} & \multicolumn{1}{c}{CATB} & \multicolumn{1}{c}{PFN} & \multicolumn{1}{c}{DAN} & \multicolumn{1}{c}{MLP} & \multicolumn{1}{c}{NODE} & \multicolumn{1}{c}{RES} & \multicolumn{1}{c}{SWI} & \multicolumn{1}{c}{CAP} & \multicolumn{1}{c}{NET} & \multicolumn{1}{c}{GOS} & \multicolumn{1}{c}{INT} & \multicolumn{1}{c}{DCN} & \multicolumn{1}{c}{FTT} & \multicolumn{1}{c}{GRO} & \multicolumn{1}{c}{SAT} & \multicolumn{1}{c}{SNN} & \multicolumn{1}{c}{TTF} & \multicolumn{1}{c}{TABR} & \multicolumn{1}{c}{NCA} & \multicolumn{1}{c}{LMA} & \multicolumn{1}{c}{TLLM} & \multicolumn{1}{c}{UNI} & AVE   \\
\midrule
0   & \textbackslash{} & 0.833    & 0.855   & 0.879    & 0.862  & 0.674  & 0.848 & 0.663 & 0.857  & 0.871     & 0.775   & 0.681  & 0.859  & 0.683   & 0.868 & 0.774          & 0.589   & 0.875 & 0.788 & 0.522          & 0.893 & 0.880     & 0.848     & 0.783  & 0.853      & 0.792   \\
1   & 0.057            & 0.837    & 0.859   & 0.870    & 0.866  & 0.683  & 0.857 & 0.667 & 0.861  & 0.871     & 0.788   & 0.687  & 0.864  & 0.685   & 0.870 & 0.777          & 0.594   & 0.880 & 0.788 & 0.522          & 0.889 & 0.882     & 0.853     & 0.788  & 0.870      & 0.796   \\
2   & 0.108            & 0.839    & 0.855   & 0.871    & 0.862  & 0.659  & 0.855 & 0.645 & 0.866  & 0.879     & 0.779   & 0.705  & 0.866  & 0.701   & 0.868 & 0.784          & 0.603   & 0.880 & 0.784 & 0.529          & 0.895 & 0.879     & 0.851     & 0.804  & 0.853      & 0.796   \\
3   & -0.233           & 0.817    & 0.824   & 0.864    & 0.859  & 0.674  & 0.839 & 0.649 & 0.842  & 0.862     & 0.781   & 0.685  & 0.859  & 0.654   & 0.851 & 0.772          & 0.592   & 0.861 & 0.797 & 0.516          & 0.879 & 0.877     & 0.843     & 0.804  & 0.837      & 0.785   \\
\bottomrule
\end{tabular}
}
}
\subtable[Model performance of SC experiment under ROC-AUC on Heart dataset.]{
\resizebox{\textwidth}{16mm}{
\begin{tabular}{ccccccccccccccccccccccccccc}
\toprule
Column & PCC & \multicolumn{1}{c}{LGB} & \multicolumn{1}{c}{XGB} & \multicolumn{1}{c}{CATB} & \multicolumn{1}{c}{PFN} & \multicolumn{1}{c}{DAN} & \multicolumn{1}{c}{MLP} & \multicolumn{1}{c}{NODE} & \multicolumn{1}{c}{RES} & \multicolumn{1}{c}{SWI} & \multicolumn{1}{c}{CAP} & \multicolumn{1}{c}{NET} & \multicolumn{1}{c}{GOS} & \multicolumn{1}{c}{INT} & \multicolumn{1}{c}{DCN} & \multicolumn{1}{c}{FTT} & \multicolumn{1}{c}{GRO} & \multicolumn{1}{c}{SAT} & \multicolumn{1}{c}{SNN} & \multicolumn{1}{c}{TTF} & \multicolumn{1}{c}{TABR} & \multicolumn{1}{c}{NCA} & \multicolumn{1}{c}{LMA} & \multicolumn{1}{c}{TLLM} & \multicolumn{1}{c}{UNI} & AVE   \\
\midrule
0   & \textbackslash{} & 0.902    & 0.909   & 0.938    & 0.939  & 0.727  & 0.931 & 0.881 & 0.937  & 0.949     & 0.867   & 0.706  & 0.947  & 0.784   & 0.938 & 0.865          & 0.626   & 0.955 & 0.865 & 0.448          & 0.961 & 0.960     & 0.653     & 0.773  & 0.911      & 0.849   \\
1   & 0.057            & 0.903    & 0.910   & 0.932    & 0.934  & 0.732  & 0.934 & 0.880 & 0.936  & 0.951     & 0.874   & 0.732  & 0.945  & 0.793   & 0.942 & 0.878          & 0.623   & 0.954 & 0.861 & 0.465          & 0.958 & 0.961     & 0.629     & 0.786  & 0.896      & 0.850   \\
2   & 0.108            & 0.905    & 0.910   & 0.936    & 0.936  & 0.734  & 0.940 & 0.883 & 0.940  & 0.950     & 0.878   & 0.745  & 0.949  & 0.797   & 0.938 & 0.872          & 0.628   & 0.957 & 0.871 & 0.447          & 0.959 & 0.958     & 0.727     & 0.797  & 0.894      & 0.856   \\
3   & -0.233           & 0.889    & 0.900   & 0.932    & 0.937  & 0.734  & 0.926 & 0.871 & 0.926  & 0.944     & 0.872   & 0.721  & 0.940  & 0.749   & 0.925 & 0.847          & 0.623   & 0.951 & 0.869 & 0.457          & 0.955 & 0.950     & 0.739     & 0.804  & 0.906      & 0.849   \\
\bottomrule
\end{tabular}
}
}
\subtable[Model performance of MC-M experiment under accuracy on Heart dataset.]{
\resizebox{\textwidth}{16mm}{
\begin{tabular}{cccccccccccccccccccccccccc}
\toprule
Degree & \multicolumn{1}{c}{LGB} & \multicolumn{1}{c}{XGB} & \multicolumn{1}{c}{CATB} & \multicolumn{1}{c}{PFN} & \multicolumn{1}{c}{DAN} & \multicolumn{1}{c}{MLP} & \multicolumn{1}{c}{NODE} & \multicolumn{1}{c}{RES} & \multicolumn{1}{c}{SWI} & \multicolumn{1}{c}{CAP} & \multicolumn{1}{c}{NET} & \multicolumn{1}{c}{GOS} & \multicolumn{1}{c}{INT} & \multicolumn{1}{c}{DCN} & \multicolumn{1}{c}{FTT} & \multicolumn{1}{c}{GRO} & \multicolumn{1}{c}{SAT} & \multicolumn{1}{c}{SNN} & \multicolumn{1}{c}{TTF} & \multicolumn{1}{c}{TABR} & \multicolumn{1}{c}{NCA} & \multicolumn{1}{c}{LMA} & \multicolumn{1}{c}{TLLM} & \multicolumn{1}{c}{UNI} & AVE   \\
\midrule
0     & 0.833    & 0.855   & 0.879    & 0.862  & 0.674  & 0.848 & 0.663 & 0.857  & 0.871     & 0.775   & 0.681  & 0.859  & 0.683   & 0.868 & 0.774          & 0.589   & 0.875 & 0.788 & 0.522          & 0.893 & 0.880     & 0.848     & 0.783  & 0.853      & 0.792   \\
9\%   & 0.736    & 0.750   & 0.663    & 0.817  & 0.582  & 0.808 & 0.639 & 0.824  & 0.835     & 0.721   & 0.656  & 0.835  & 0.659   & 0.832 & 0.694          & 0.596   & 0.832 & 0.770 & 0.487          & 0.853 & 0.788     & 0.674     & 0.777  & 0.842      & 0.736   \\
18\%  & 0.665    & 0.737   & 0.663    & 0.797  & 0.571  & 0.783 & 0.616 & 0.781  & 0.821     & 0.665   & 0.618  & 0.793  & 0.665   & 0.797 & 0.681          & 0.596   & 0.792 & 0.732 & 0.509          & 0.842 & 0.739     & 0.685     & 0.783  & 0.859      & 0.716   \\
27\%  & 0.634    & 0.688   & 0.730    & 0.779  & 0.585  & 0.750 & 0.603 & 0.766  & 0.790     & 0.645   & 0.623  & 0.763  & 0.629   & 0.763 & 0.678          & 0.596   & 0.755 & 0.692 & 0.529          & 0.817 & 0.714     & 0.707     & 0.793  & 0.810      & 0.702   \\
\bottomrule
\end{tabular}
}
}
\subtable[Model performance of MC-M experiment under ROC-AUC on Heart dataset.]{
\resizebox{\textwidth}{16mm}{
\begin{tabular}{cccccccccccccccccccccccccc}
\toprule
Degree & \multicolumn{1}{c}{LGB} & \multicolumn{1}{c}{XGB} & \multicolumn{1}{c}{CATB} & \multicolumn{1}{c}{PFN} & \multicolumn{1}{c}{DAN} & \multicolumn{1}{c}{MLP} & \multicolumn{1}{c}{NODE} & \multicolumn{1}{c}{RES} & \multicolumn{1}{c}{SWI} & \multicolumn{1}{c}{CAP} & \multicolumn{1}{c}{NET} & \multicolumn{1}{c}{GOS} & \multicolumn{1}{c}{INT} & \multicolumn{1}{c}{DCN} & \multicolumn{1}{c}{FTT} & \multicolumn{1}{c}{GRO} & \multicolumn{1}{c}{SAT} & \multicolumn{1}{c}{SNN} & \multicolumn{1}{c}{TTF} & \multicolumn{1}{c}{TABR} & \multicolumn{1}{c}{NCA} & \multicolumn{1}{c}{LMA} & \multicolumn{1}{c}{TLLM} & \multicolumn{1}{c}{UNI} & AVE   \\
\midrule
0     & 0.902    & 0.909   & 0.938    & 0.939  & 0.727  & 0.931 & 0.881 & 0.937  & 0.949     & 0.867   & 0.706  & 0.947  & 0.784   & 0.938 & 0.865          & 0.626   & 0.955 & 0.865 & 0.448          & 0.961 & 0.960     & 0.653     & 0.773  & 0.911      & 0.849   \\
9\%   & 0.848    & 0.861   & 0.856    & 0.908  & 0.651  & 0.910 & 0.843 & 0.918  & 0.930     & 0.826   & 0.691  & 0.935  & 0.720   & 0.920 & 0.794          & 0.646   & 0.936 & 0.851 & 0.457          & 0.942 & 0.919     & 0.630     & 0.762  & 0.888      & 0.818   \\
18\%  & 0.817    & 0.832   & 0.811    & 0.897  & 0.613  & 0.887 & 0.819 & 0.892  & 0.911     & 0.737   & 0.651  & 0.917  & 0.738   & 0.887 & 0.783          & 0.639   & 0.923 & 0.805 & 0.442          & 0.928 & 0.904     & 0.649     & 0.771  & 0.894      & 0.798   \\
27\%  & 0.789    & 0.810   & 0.834    & 0.862  & 0.592  & 0.858 & 0.781 & 0.861  & 0.871     & 0.685   & 0.640  & 0.889  & 0.715   & 0.845 & 0.805          & 0.641   & 0.893 & 0.757 & 0.496          & 0.893 & 0.861     & 0.680     & 0.782  & 0.896      & 0.781   \\
\bottomrule
\end{tabular}
}
}
\subtable[Model performance of MC-L experiment under accuracy on Heart dataset.]{
\resizebox{\textwidth}{16mm}{
\begin{tabular}{cccccccccccccccccccccccccc}
\toprule
Degree & \multicolumn{1}{c}{LGB} & \multicolumn{1}{c}{XGB} & \multicolumn{1}{c}{CATB} & \multicolumn{1}{c}{PFN} & \multicolumn{1}{c}{DAN} & \multicolumn{1}{c}{MLP} & \multicolumn{1}{c}{NODE} & \multicolumn{1}{c}{RES} & \multicolumn{1}{c}{SWI} & \multicolumn{1}{c}{CAP} & \multicolumn{1}{c}{NET} & \multicolumn{1}{c}{GOS} & \multicolumn{1}{c}{INT} & \multicolumn{1}{c}{DCN} & \multicolumn{1}{c}{FTT} & \multicolumn{1}{c}{GRO} & \multicolumn{1}{c}{SAT} & \multicolumn{1}{c}{SNN} & \multicolumn{1}{c}{TTF} & \multicolumn{1}{c}{TABR} & \multicolumn{1}{c}{NCA} & \multicolumn{1}{c}{LMA} & \multicolumn{1}{c}{TLLM} & \multicolumn{1}{c}{UNI} & AVE   \\
\midrule
0     & 0.833    & 0.855   & 0.879    & 0.862  & 0.674  & 0.848 & 0.663 & 0.857  & 0.871     & 0.775   & 0.681  & 0.859  & 0.683   & 0.868 & 0.774          & 0.589   & 0.875 & 0.788 & 0.522          & 0.893 & 0.880     & 0.848     & 0.783  & 0.853      & 0.792   \\
9\%   & 0.837    & 0.859   & 0.870    & 0.866  & 0.683  & 0.857 & 0.667 & 0.861  & 0.871     & 0.788   & 0.687  & 0.864  & 0.685   & 0.870 & 0.777          & 0.594   & 0.880 & 0.788 & 0.522          & 0.889 & 0.882     & 0.853     & 0.788  & 0.870      & 0.796   \\
18\%  & 0.842    & 0.857   & 0.871    & 0.861  & 0.668  & 0.862 & 0.652 & 0.864  & 0.875     & 0.795   & 0.685  & 0.871  & 0.712   & 0.871 & 0.793          & 0.600   & 0.879 & 0.783 & 0.516          & 0.889 & 0.879     & 0.804     & 0.815  & 0.837      & 0.795   \\
27\%  & 0.817    & 0.830   & 0.844    & 0.855  & 0.690  & 0.850 & 0.647 & 0.850  & 0.866     & 0.810   & 0.690  & 0.862  & 0.685   & 0.861 & 0.781          & 0.585   & 0.861 & 0.793 & 0.516          & 0.875 & 0.875     & 0.793     & 0.788  & 0.853      & 0.787   \\
\bottomrule
\end{tabular}
}
}
\end{table}
\newpage
\begin{table}
\subtable[Model performance of MC-L experiment under ROC-AUC on Heart dataset.]{
\resizebox{\textwidth}{16mm}{
\begin{tabular}{cccccccccccccccccccccccccc}
\toprule
Degree & \multicolumn{1}{c}{LGB} & \multicolumn{1}{c}{XGB} & \multicolumn{1}{c}{CATB} & \multicolumn{1}{c}{PFN} & \multicolumn{1}{c}{DAN} & \multicolumn{1}{c}{MLP} & \multicolumn{1}{c}{NODE} & \multicolumn{1}{c}{RES} & \multicolumn{1}{c}{SWI} & \multicolumn{1}{c}{CAP} & \multicolumn{1}{c}{NET} & \multicolumn{1}{c}{GOS} & \multicolumn{1}{c}{INT} & \multicolumn{1}{c}{DCN} & \multicolumn{1}{c}{FTT} & \multicolumn{1}{c}{GRO} & \multicolumn{1}{c}{SAT} & \multicolumn{1}{c}{SNN} & \multicolumn{1}{c}{TTF} & \multicolumn{1}{c}{TABR} & \multicolumn{1}{c}{NCA} & \multicolumn{1}{c}{LMA} & \multicolumn{1}{c}{TLLM} & \multicolumn{1}{c}{UNI} & AVE   \\
\midrule
0     & 0.902    & 0.909   & 0.938    & 0.939  & 0.727  & 0.931 & 0.881 & 0.937  & 0.949     & 0.867   & 0.706  & 0.947  & 0.784   & 0.938 & 0.865          & 0.626   & 0.955 & 0.865 & 0.448          & 0.961 & 0.960     & 0.653     & 0.773  & 0.911      & 0.849   \\
9\%   & 0.903    & 0.910   & 0.932    & 0.934  & 0.732  & 0.934 & 0.880 & 0.936  & 0.951     & 0.874   & 0.732  & 0.945  & 0.793   & 0.942 & 0.878          & 0.623   & 0.954 & 0.861 & 0.465          & 0.958 & 0.961     & 0.629     & 0.786  & 0.896      & 0.850   \\
18\%  & 0.907    & 0.910   & 0.929    & 0.931  & 0.732  & 0.944 & 0.883 & 0.939  & 0.951     & 0.884   & 0.748  & 0.948  & 0.814   & 0.944 & 0.886          & 0.616   & 0.956 & 0.869 & 0.462          & 0.957 & 0.960     & 0.812     & 0.817  & 0.898      & 0.862   \\
27\%  & 0.899    & 0.905   & 0.924    & 0.930  & 0.742  & 0.937 & 0.876 & 0.928  & 0.946     & 0.889   & 0.757  & 0.940  & 0.777   & 0.938 & 0.872          & 0.613   & 0.952 & 0.873 & 0.461          & 0.951 & 0.954     & 0.804     & 0.794  & 0.914      & 0.857   \\
\bottomrule
\end{tabular}
}
}
\subtable[Model performance of RD experiment under accuracy on Heart dataset.]{
\resizebox{\textwidth}{16mm}{
\begin{tabular}{cccccccccccccccccccccccccc}
\toprule
Degree & \multicolumn{1}{c}{LGB} & \multicolumn{1}{c}{XGB} & \multicolumn{1}{c}{CATB} & \multicolumn{1}{c}{PFN} & \multicolumn{1}{c}{DAN} & \multicolumn{1}{c}{MLP} & \multicolumn{1}{c}{NODE} & \multicolumn{1}{c}{RES} & \multicolumn{1}{c}{SWI} & \multicolumn{1}{c}{CAP} & \multicolumn{1}{c}{NET} & \multicolumn{1}{c}{GOS} & \multicolumn{1}{c}{INT} & \multicolumn{1}{c}{DCN} & \multicolumn{1}{c}{FTT} & \multicolumn{1}{c}{GRO} & \multicolumn{1}{c}{SAT} & \multicolumn{1}{c}{SNN} & \multicolumn{1}{c}{TTF} & \multicolumn{1}{c}{TABR} & \multicolumn{1}{c}{NCA} & \multicolumn{1}{c}{LMA} & \multicolumn{1}{c}{TLLM} & \multicolumn{1}{c}{UNI} & AVE   \\
\midrule
0     & 0.833    & 0.855   & 0.879    & 0.862  & 0.674  & 0.848 & 0.663 & 0.857  & 0.871     & 0.775   & 0.681  & 0.859  & 0.683   & 0.868 & 0.774          & 0.589   & 0.875 & 0.788 & 0.522          & 0.893 & 0.880     & 0.848     & 0.783  & 0.853      & 0.792   \\
9\%   & 0.816    & 0.825   & 0.851    & 0.859  & 0.661  & 0.843 & 0.640 & 0.856  & 0.864     & 0.781   & 0.667  & 0.859  & 0.674   & 0.857 & 0.760          & 0.593   & 0.864 & 0.779 & 0.513          & 0.889 & 0.872     & 0.772     & 0.793  & 0.847      & 0.781   \\
18\%  & 0.794    & 0.798   & 0.826    & 0.848  & 0.649  & 0.833 & 0.622 & 0.849  & 0.858     & 0.772   & 0.661  & 0.853  & 0.656   & 0.846 & 0.742          & 0.599   & 0.849 & 0.767 & 0.516          & 0.886 & 0.850     & 0.752     & 0.793  & 0.835      & 0.769   \\
27\%  & 0.772    & 0.773   & 0.801    & 0.835  & 0.633  & 0.821 & 0.603 & 0.836  & 0.844     & 0.767   & 0.631  & 0.838  & 0.656   & 0.827 & 0.718          & 0.585   & 0.832 & 0.758 & 0.502          & 0.878 & 0.833     & 0.721     & 0.780  & 0.831      & 0.753   \\
\bottomrule
\end{tabular}
}
}
\subtable[Model performance of RD experiment under ROC-AUC on Heart dataset.]{
\resizebox{\textwidth}{16mm}{
%
}
}
\subtable[Model performance gap $\Delta$ of RD experiment under accuracy on Heart dataset.]{
\resizebox{\textwidth}{16mm}{
%
}
}
\subtable[Model performance gap $\Delta$ of RD experiment under ROC-AUC on Heart dataset.]{
\resizebox{\textwidth}{16mm}{
%
}
}
\end{table}

\begin{table}
\caption{Model performance on MiniBooNE dataset.}
\subtable[Model performance of SC experiment under accuracy on MiniBooNE dataset.]{
\resizebox{\textwidth}{53mm}{
%
}
}
\subtable[Model performance of SC experiment under ROC-AUC on MiniBooNE dataset.]{
\resizebox{\textwidth}{53mm}{
%
}
}
\end{table}
\newpage
\begin{table}
\subtable[Model performance of MC-M experiment under accuracy on MiniBooNE dataset.]{
\resizebox{\textwidth}{53mm}{
%
}
}
\subtable[Model performance of MC-M experiment under ROC-AUC on MiniBooNE dataset.]{
\resizebox{\textwidth}{53mm}{
%
}
}
\end{table}
\newpage
\begin{table}
\subtable[Model performance of MC-L experiment under accuracy on MiniBooNE dataset.]{
\resizebox{\textwidth}{53mm}{
%
}
}

\subtable[Model performance of MC-L experiment under ROC-AUC on MiniBooNE dataset.]{
\resizebox{\textwidth}{53mm}{
%
}
}
\end{table}
\newpage
\begin{table}
\subtable[Model performance of RD experiment under accuracy on MiniBooNE dataset.]{
\resizebox{\textwidth}{53mm}{
%
}
}
\subtable[Model performance of RD experiment under ROC-AUC on MiniBooNE dataset.]{
\resizebox{\textwidth}{53mm}{
%
}
}
\end{table}
\newpage
\begin{table}
\subtable[Model performance gap $\Delta$ of RD experiment under accuracy on MiniBooNE dataset.]{
\resizebox{\textwidth}{53mm}{
%
}
}
\subtable[Model performance gap $\Delta$ of RD experiment under ROC-AUC on MiniBooNE dataset.]{
\resizebox{\textwidth}{53mm}{
%
}
}
\end{table}

\begin{table}
\caption{Model performance on Eyemovements dataset.}
\subtable[Model performance of SC experiment under accuracy on Eyemovements dataset.]{
\resizebox{\textwidth}{33mm}{
%
}
}
\subtable[Model performance of SC experiment under ROC-AUC on Eyemovements dataset.]{
\resizebox{\textwidth}{33mm}{
%
}
}
\subtable[Model performance of MC-M experiment under accuracy on Eyemovements dataset.]{
\resizebox{\textwidth}{33mm}{
%
}
}
\end{table}
\newpage
\begin{table}
\subtable[Model performance of MC-M experiment under ROC-AUC on Eyemovements dataset.]{
\resizebox{\textwidth}{33mm}{
%
}
}
\subtable[Model performance of MC-L experiment under accuracy on Eyemovements dataset.]{
\resizebox{\textwidth}{33mm}{
%
}
}
\subtable[Model performance of MC-L experiment under ROC-AUC on Eyemovements dataset.]{
\resizebox{\textwidth}{33mm}{
%
}
}
\end{table}
\newpage
\begin{table}
\subtable[Model performance of RD experiment under accuracy on Eyemovements dataset.]{
\resizebox{\textwidth}{33mm}{
%
}
}
\subtable[Model performance of RD experiment under ROC-AUC on Eyemovements dataset.]{
\resizebox{\textwidth}{33mm}{
%
}
}
\subtable[Model performance gap $\Delta$ of RD experiment under accuracy on Eyemovements dataset.]{
\resizebox{\textwidth}{33mm}{
%
}
}
\end{table}
\newpage
\begin{table}
\subtable[Model performance gap $\Delta$ of RD experiment under ROC-AUC on Eyemovements dataset.]{
\resizebox{\textwidth}{33mm}{
%
}
}
\end{table}

\begin{table}
\caption{Model performance on Iris dataset.}
\subtable[Model performance of SC experiment under accuracy on Iris dataset.]{
\resizebox{\textwidth}{8mm}{
%
}
}
\subtable[Model performance of SC experiment under ROC-AUC on Iris dataset.]{
\resizebox{\textwidth}{8mm}{
%
}
}
\subtable[Model performance of MC-M experiment under accuracy on Iris dataset.]{
\resizebox{\textwidth}{8mm}{
%
}
}
\subtable[Model performance of MC-M experiment under ROC-AUC on Iris dataset.]{
\resizebox{\textwidth}{8mm}{
%
}
}
\subtable[Model performance of MC-L experiment under accuracy on Iris dataset.]{
\resizebox{\textwidth}{8mm}{
%
}
}
\subtable[Model performance of MC-L experiment under ROC-AUC on Iris dataset.]{
\resizebox{\textwidth}{7mm}{
%
}
}
\subtable[Model performance of RD experiment under accuracy on Iris dataset.]{
\resizebox{\textwidth}{7mm}{
%
}
}
\subtable[Model performance of RD experiment under ROC-AUC on Iris dataset.]{
\resizebox{\textwidth}{8mm}{
%
}
}
\subtable[Model performance gap $\Delta$ of RD experiment under accuracy on Iris dataset.]{
\resizebox{\textwidth}{8mm}{
%
}
}
\subtable[Model performance gap $\Delta$ of RD experiment under ROC-AUC on Iris dataset.]{
\resizebox{\textwidth}{8mm}{
%
}
}
\end{table}

\begin{table}
\caption{Model performance on Jannis dataset.}
\subtable[Model performance of SC experiment under accuracy on Jannis dataset.]{
\resizebox{\textwidth}{65mm}{
%
}
}
\end{table}
\newpage
\begin{table}
\subtable[Model performance of SC experiment under ROC-AUC on Jannis dataset.]{
\resizebox{\textwidth}{65mm}{
%
}
}
\end{table}
\newpage
\begin{table}
\subtable[Model performance of MC-M experiment under accuracy on Jannis dataset.]{
\resizebox{\textwidth}{65mm}{
%
}
}
\end{table}
\newpage
\begin{table}
\subtable[Model performance of MC-M experiment under ROC-AUC on Jannis dataset.]{
\resizebox{\textwidth}{65mm}{
%
}
}
\end{table}
\newpage
\begin{table}
\subtable[Model performance of MC-L experiment under accuracy on Jannis dataset.]{
\resizebox{\textwidth}{65mm}{
%
}
}
\end{table}
\newpage
\begin{table}
\subtable[Model performance of MC-L experiment under ROC-AUC on Jannis dataset.]{
\resizebox{\textwidth}{65mm}{
%
}
}
\end{table}
\newpage
\begin{table}
\subtable[Model performance of RD experiment under accuracy on Jannis dataset.]{
\resizebox{\textwidth}{65mm}{
%
}
}
\end{table}
\newpage
\begin{table}
\subtable[Model performance of RD experiment under ROC-AUC on Jannis dataset.]{
\resizebox{\textwidth}{65mm}{
%
}
}
\end{table}
\newpage
\begin{table}
\subtable[Model performance gap $\Delta$ of RD experiment under accuracy on Jannis dataset.]{
\resizebox{\textwidth}{65mm}{
%
}
}
\end{table}
\newpage
\begin{table}
\subtable[Model performance gap $\Delta$ of RD experiment under ROC-AUC on Jannis dataset.]{
\resizebox{\textwidth}{65mm}{
%
}
}
\end{table}

\begin{table}
\caption{Model performance on Penguins dataset.}
\subtable[Model performance of SC experiment under accuracy on Penguins dataset.]{
\resizebox{\textwidth}{10mm}{
%
}
}
\subtable[Model performance of SC experiment under ROC-AUC on Penguins dataset.]{
\resizebox{\textwidth}{10mm}{
%
}
}
\subtable[Model performance of MC-M experiment under accuracy on Penguins dataset.]{
\resizebox{\textwidth}{10mm}{
%
}
}
\subtable[Model performance of MC-M experiment under ROC-AUC on Penguins dataset.]{
\resizebox{\textwidth}{10mm}{
%
}
}
\subtable[Model performance of MC-L experiment under accuracy on Penguins dataset.]{
\resizebox{\textwidth}{10mm}{
%
}
}

\subtable[Model performance of MC-L experiment under ROC-AUC on Penguins dataset.]{
\resizebox{\textwidth}{10mm}{
%
}
}
\subtable[Model performance of RD experiment under accuracy on Penguins dataset.]{
\resizebox{\textwidth}{10mm}{
%
}
}
\subtable[Model performance of RD experiment under ROC-AUC on Penguins dataset.]{
\resizebox{\textwidth}{10mm}{
%
}
}
\end{table}
\newpage
\begin{table}
\subtable[Model performance gap $\Delta$ of RD experiment under accuracy on Penguins dataset.]{
\resizebox{\textwidth}{10mm}{
%
}
}
\subtable[Model performance gap $\Delta$ of RD experiment under ROC-AUC on Penguins dataset.]{
\resizebox{\textwidth}{10mm}{
%
}
}
\end{table}

\begin{table}
\caption{Model performance on Abalone dataset.}
\subtable[Model performance of SC experiment under RMSE on Abalone dataset.]{
\resizebox{\textwidth}{15mm}{
%
}
}
\subtable[Model performance of MC-M experiment under RMSE on Abalone dataset.]{
\resizebox{\textwidth}{15mm}{
%
}
}
\subtable[Model performance of MC-L experiment under RMSE on Abalone dataset.]{
\resizebox{\textwidth}{15mm}{
%
}
}
\subtable[Model performance of RD experiment under RMSE on Abalone dataset.]{
\resizebox{\textwidth}{15mm}{
%
}
}
\subtable[Model performance gap $\Delta$ of RD experiment under RMSE on Abalone dataset.]{
\resizebox{\textwidth}{14mm}{
%
}
}
\end{table}

\begin{table}
\caption{Model performance on Bike dataset.}
\subtable[Model performance of SC experiment under RMSE on Bike dataset.]{
\resizebox{\textwidth}{14mm}{
%
}
}
\subtable[Model performance of MC-M experiment under RMSE on Bike dataset.]{
\resizebox{\textwidth}{14mm}{
%
}
}
\subtable[Model performance of MC-L experiment under RMSE on Bike dataset.]{
\resizebox{\textwidth}{14mm}{
%
}
}
\subtable[Model performance of RD experiment under RMSE on Bike dataset.]{
\resizebox{\textwidth}{14mm}{
%
}
}
\subtable[Model performance gap $\Delta$ of RD experiment under RMSE on Bike dataset.]{
\resizebox{\textwidth}{14mm}{
%
}
}
\end{table}

\begin{table}
\caption{Model performance on Concrete dataset.}
\subtable[Model performance of SC experiment under RMSE on Concrete dataset.]{
\resizebox{\textwidth}{14mm}{
%
}
}
\subtable[Model performance of MC-M experiment under RMSE on Concrete dataset.]{
\resizebox{\textwidth}{14mm}{
%
}
}
\subtable[Model performance of MC-L experiment under RMSE on Concrete dataset.]{
\resizebox{\textwidth}{14mm}{
%
}
}
\subtable[Model performance of RD experiment under RMSE on Concrete dataset.]{
\resizebox{\textwidth}{14mm}{
%
}
}
\subtable[Model performance gap $\Delta$ of RD experiment under RMSE on Concrete dataset.]{
\resizebox{\textwidth}{14mm}{
%
}
}
\end{table}

\begin{table}
\caption{Model performance on Laptop dataset.}
\label{table17}
\subtable[Model performance of SC experiment under RMSE on Laptop dataset.]{
\resizebox{\textwidth}{26mm}{
%
}
}
\subtable[Model performance of MC-M experiment under RMSE on Laptop dataset.]{
\resizebox{\textwidth}{26mm}{
%
}
}
\subtable[Model performance of MC-L experiment under RMSE on Laptop dataset.]{
\resizebox{\textwidth}{26mm}{
%
}
}
\end{table}
\newpage
\begin{table}
\subtable[Model performance of RD experiment under RMSE on Laptop dataset.]{
\resizebox{\textwidth}{26mm}{
%
}
}
\subtable[Model performance gap $\Delta$ of RD experiment under RMSE on Laptop dataset.]{
\resizebox{\textwidth}{26mm}{
%
}
}
\end{table}

\end{document}